\RequirePackage{fix-cm}
\documentclass[smallcondensed]{svjour3}     
\smartqed  
\usepackage{graphicx}
\usepackage{epstopdf}
\usepackage{graphicx}
\usepackage{hyperref} 
\usepackage{algorithm,algorithmic}
\usepackage{colortbl}
\usepackage{color}
\usepackage[numbers]{natbib}
\usepackage{chngcntr}
\usepackage{longtable}
\usepackage{multirow}
\usepackage{rotating}
\usepackage{subfig}

\usepackage{amsmath,amsfonts,amssymb,amsthm}

\theoremstyle{plain}
\newtheorem{assumption}{Assumption}
\newtheorem{theorem1}{Theorem}
\begin{document}
	
	\title{Faster Learning by Reduction of Data Access Time \thanks{(a) This is a post-peer-review, pre-copyedit version of an article published in Applied Intelligence. The final authenticated version is available online at: \url{https://doi.org/10.1007/s10489-018-1235-x} \cite{Chauhan2018SS_AI} (b) The first  and second preprints of the paper were published with a different name \cite{Chauhan2018SS}.}
	}
	
	
	\author{Vinod~Kumar~Chauhan$\color{red}^*$   \thanks{$\color{red}^*$corresponding author}         \and
		Anuj~Sharma \and
		Kalpana~Dahiya
	}
	
	
	\institute{Vinod Kumar Chauhan \at
		Computer Science \& Applications\\
		Panjab University Chandigarh, INDIA\\
		\email{vkumar@pu.ac.in,jmdvinodjmd@gmail.com}           
		\and
		Anuj Sharma \at
		Computer Science \& Applications\\
		Panjab University Chandigarh, INDIA\\
		\email{anujs@pu.ac.in}\\
		Homepage: \url{https://sites.google.com/view/anujsharma/}
		\and
		Kalpana Dahiya \at
		University Institute of Engineering and Technology\\
		Panjab University Chandigarh, INDIA\\
		\email{kalpanas@pu.ac.in}		
	}
	
	\date{Received: date / Accepted: date}

	\maketitle

\begin{abstract}
Nowadays, the major challenge in machine learning is the `Big~Data' challenge. The big data problems due to large number of data points or large number of features in each data point, or both, the training of models have become very slow. The training time has two major components: Time to access the data and time to process (learn from) the data. So far, the research has focused only on the second part, i.e., learning from the data. In this paper, we have proposed one possible solution to handle the big data problems in machine learning. The idea is to reduce the training time through reducing data access time by proposing systematic sampling and cyclic/sequential sampling to select mini-batches from the dataset. To prove the effectiveness of proposed sampling techniques, we have used empirical risk minimization, which is commonly used machine learning problem, for strongly convex and smooth case. The problem has been solved using SAG, SAGA, SVRG, SAAG-II and MBSGD (Mini-batched SGD), each using two step determination techniques, namely, constant step size and backtracking line search method. Theoretical results prove similar convergence for systematic and cyclic sampling as the widely used random sampling technique, in expectation. Experimental results with bench marked datasets prove the efficacy of the proposed sampling techniques and show up to six times faster training.
\keywords{Systematic Sampling \and Random Sampling \and Cyclic Sampling\and Big data \and Large-scale Learning \and stochastic learning \and Empirical Risk Minimization}
\end{abstract}

\section{Introduction}
\label{sec_intro}
The technological developments and the availability of a number of data sources have evolved the data into `Big Data', where the meaning of `Big' in `Big Data' is continuously changing because of increasing of size of datasets. Big data has multiple aspects, like, volume, velocity, variety and veracity etc., and one aspect, namely, volume, i.e., size of datasets have posed a challenge to the machine learners to train the models over these large datasets. These problems which can be called as large-scale or big data problems, have large number of data points or large number of features in each data point, or both, which lead to slow training of models. So nowadays, the major challenge is to develop efficient and scalable learning algorithms for dealing with big data problems \cite{Zhou2017,Chauhan2017Saag,Chauhan2018Review}. The training time of models have two major components \cite{Yu2010} as:
\begin{equation}
	\label{eq_time}
	\textit{\text{training time} = \text{time to access data} + \text{time to process data}}.
\end{equation}

As it is known that for processing any data or running any program, it should be first brought into memory, more precisely into RAM (Random Access Memory), from hard disk. The time taken in brining the data or program from hard disk into memory is called access time. Access time has, further, three components, namely, seek time (it is the time taken by the reading head to move from current position up to the track containing the data), rotational latency time (it is the time taken by the reading head from the current position to reach up to the sector/block containing the data) and transfer time (it is the time taken to transfer data from sector/block over the disk to memory). Moreover, data is not read content wise rather block-wise, where each block can have multiple sectors. Now, it is interesting to note that for data stored on the contiguous memory or in the close proximity, has lesser data access time as compared with the data dispersed far away from each other. This is due to lesser seek, latency and transfer times. In case of SSD (Solid State Disk) and RAM (Random Access Memory), there are no seek and latency times because they do not have moving parts and they are based on direct access mechanism, but transfer time still plays its role. Since data are read/written block-wise so for contiguous data access, there would be one or two transfer times but dispersed data access would require more number of transfer times. Moreover, cache memory strategies also favor the contiguous memory access and make it faster as compared to dispersed data access. Thus contiguous data access time is faster than dispersed data access, in all the cases whether data is stored on RAM, SSD or HDD. But the difference in access time would be more prominent for HDD.\\
The second component of training time is processing (learning) time which is the time taken by the CPU (Central Processing Unit) to process the data to solve for model parameters. Due to mini-batching and iterative learning algorithms, the accessing and processing of data is intermixed and occurs quite frequently. The data access time is dependent on the sampling technique and processing time is dependent on the method used to solve the problem. Till this date, to improve the training time, most of the focus is given only to improve the processing time by using different methods.  The training time is dependent on both data accessing time and data processing time, and this is to be noted that, generally, data access time is very high as compared to the data processing time. Thus, efforts should be equally done to improve the data accessing time. This paper has tried to reduce the training time for big data problems by reducing the data access time using systematic sampling technique \cite{William1949,William1944} and cyclic sampling to select mini-batches of data points from the dataset.

\subsection{Optimization Problem}
\label{subsec_opt_prob}
Empirical risk minimization (ERM) problem \cite{Reddi2016} is a commonly used problem in machine learning and statistics, which typically consists of a loss function and a regularization term. In this paper, $l2$-regularized ERM problem with the assumptions of strong convexity and smoothness, is used for demonstrating the effectiveness of proposed sampling techniques. For the training data $\lbrace \left(x_1, y_1\right),\left(x_2, y_2\right),...,\left(x_l, y_l\right) \rbrace$, where $x_i \in \mathbb{R}^n, \; y_i \in \mathbb{R}$, and for loss functions $f_1, f_2,...,f_l: \; \mathbb{R}^n \rightarrow \mathbb{R}$, model parameter vector $w \in \mathbb{R}^n$ and regularization coefficient $C >0$, $l2$-regularized ERM problem is given by
\begin{equation}
\label{eq_erm}
\begin{split}
\underset{w}{\min} \;\; f(w) = \dfrac{1}{l}\sum_{i=1}^{l} f_i(w) + \dfrac{C}{2} \|w\|^2,
\end{split}
\end{equation}
where first term is the data fitting term and second term is used for avoiding the over-fitting of data. Commonly used loss functions are the square loss $\left( w^Tx_i-y_i\right) ^2$, the logistic loss $\log\left(1 + exp\left(-y_iw^Tx_i\right)\right)$ and the hinge loss\\ $\max\left(0, 1-y_iw^Tx_i\right)$. 
In this paper, experimentation uses logistic loss function. When problem represented by eq.(\ref{eq_erm}) is a large-scale or big data problem then per iteration complexity of learning algorithms with traditional methods like Gradient Descent is $O(nl)$ which is very high. This is because, such problems have large number of data points ($l$) or large number of features ($n$), or both, and each iteration updates $n$ variables over $l$ data points. Because of this high computational complexity per iteration it would be very expensive or even infeasible for single machine to process a single iteration of learning algorithm. Stochastic approximation approach is widely used for handling such cases (e.g., \cite{Zhang2004,Nemirovski2009,Shalev-Shwartz2011,Johnson2013,Konecny2013,Martin2013,Li2014,Defazio2014,Xu2015,Byrd2016,Schmidt2016,Chauhan2017Saag,Chauhan2018SAAGs34}), which uses one data point or mini-batch of data points during each iteration. The reduced subproblem with mini-batch $B_j$ of data points for $j^{th}$ inner iteration is given by
\begin{equation}
\label{eq_BBOF}
\begin{split}
\underset{w}{\min} \;\; \dfrac{1}{|B_j|} \sum_{i \in B_j} f_i(w) + \dfrac{C}{2} \|w\|^2, \quad j=1,2,...,m.
\end{split}
\end{equation}
The iteration complexity for solving this reduced subproblem is $O(n)$ (for one data point) or $O(n|B_j|)$ (for mini-batch of data points) where $|B_j|$ is the size of mini-batch, which is very low and independent of $l$. Since it is easier to solve the problem~(\ref{eq_BBOF}) and this is widely used approach to handle large-scale problems so the paper uses this reduced subproblem approach.

\subsection{Literature Review}
\label{subsec_literature}
Big data challenge is one of the major challenge in machine learning \cite{Zhou2017,Chauhan2017Saag,Chauhan2018Review}. For dealing with big data problems, stochastic approximation (and coordinate descent) approaches are widely used (e.g., \cite{Chang2008,Johnson2013,Konecny2013,Defazio2014,Wright2015,Xu2015,Schmidt2016,Chauhan2017Saag}), which take one or more data points (one or block of features in coordinate descent) to form a reduced subproblem during each iteration. This helps in reducing the computational complexity per iteration and thereby helps in solving large-scale problems. But due to stochastic noise, like in SGD (Stochastic Gradient Descent) method, stochastic approximation affects the solution of the problem. This stochastic noise can be reduced using different techniques, like, mini-batching \cite{Martin2013}, decreasing step sizes \cite{Shalev-Shwartz2011}, importance sampling \cite{Csiba2016} and variance reduction techniques \cite{Johnson2013}, as discussed in \cite{Csiba2016}. For selecting one data point or mini-batch of data points from the dataset in stochastic approximation approach, random sampling (e.g., \cite{Byrd2016,Defazio2014,Johnson2013,Konecny2013,Li2014,Nemirovski2009,Schmidt2016,Shalev-Shwartz2011,Martin2013,Wang2017,Xu2015,Zhang2004}) is widely used, and other sampling techniques are importance sampling (e.g., \cite{Csiba2016,Csiba2016b,Deanna2014,Qu2015,Zhang2016,Zhao2015}), stratified sampling \cite{Zhao2014b} and adaptive sampling \cite{Gopal2016}. Importance sampling is a non-uniform sampling technique, which uses properties of data for finding probabilities of selection of data points in the iterative process, leading to the acceleration of training process. Importance sampling involves some overhead in calculating the probabilities and the average computational complexity per iteration of importance sampling could be more than the computational complexity of the iteration in some cases, like, when it is implemented dynamically. Stratified sampling technique divides the dataset into clusters of similar data points and then mini-batch of data points are selected from the clusters. In adaptive sampling \cite{Gopal2016}, information about the classes, i.e., data-labels is used for selecting the data points; this technique gives good results for problems with large number of classes. Random sampling is widely used in mini-batching for large-scale learning problems (e.g., \cite{Byrd2016,Li2014,Shalev-Shwartz2011,Martin2013}), as only \cite{Csiba2016} is known importance sampling technique in mini-batching. In this paper, we have focused on simple sampling techniques which do not involve any extra overhead and can be effective for dealing with large-scale learning problems. Two simple sampling techniques, namely, systematic sampling and cyclic/sequential sampling techniques have been proposed for selecting mini-batches. To the best of our knowledge, systematic sampling is not used in machine learning for selecting data points. We are the first to introduce systematic sampling in machine learning for selection of mini-batches of data points from the dataset. Before this, cyclic sampling was used in coordinate descent and block coordinate descent methods (e.g.,\cite{Wright2015}) for selecting one coordinate or block of coordinates, respectively, and to the best of our knowledge, cyclic sampling is not used for selecting mini-batches. Both sampling techniques are simple, effective and easy to implement. The proposed sampling techniques try to reduce the training time of models by reducing the data access time because these are based on contiguous access of data. Before this, for reducing the training time of models, the focus is mainly given on reducing the processing time, but this paper has focused on reducing the data access time by changing the sampling techniques.

\subsection{Contributions}
\label{subsec_contributions}
The contributions of the article are summarized below:
\begin{itemize}
	\item[(a)] Novel systematic sampling and sequential/cyclic sampling techniques have been proposed for selecting mini-batches of data points from the dataset for solving large-scale learning problems. The proposed techniques focus on reducing the training time of learning algorithms by reducing the data access time, and are based on simple observations that data stored on the contiguous memory locations are faster to access as compared with data stored on dispersed memory locations. To the best of our knowledge, this paper is the first to focus on reducing data access time to reduce the overall training time for machine learning algorithms.
	\item[(b)] Proposed ideas are independent of problem and method used to solve problem as it focuses on data access only. So it can be extended to other machine learning problems.
	\item[(c)] Experimental results prove the efficacy of systematic sampling and cyclic sampling in reducing the training time, and show up to six times faster training. The results have been provided using five different methods (SAG, SAGA, SVRG, SAAG-II and MBSGD) each using two step determination techniques (constant step size and backtracking line search methods) over eight bench marked datasets.
	\item[(d)] Theoretical results prove similar convergence for learning algorithms using cyclic and systematic sampling, as for widely used random sampling technique, in expectation.
\end{itemize}

\section{Systematic Sampling}
\label{sec_ss}
Sampling is the way of selecting one mini-batch of data points (or one data point) from the whole dataset. Iterative approximation methods used for solving the problem use sampling again and again during each iteration/epoch. The convergence of learning algorithm depends upon the type of sampling used since sampling controls two things: data access time and diversity of data. In general, when consecutive data points are used they reduce the data access time which reduces the training time but selected data points might not be diverse which affects the convergence of learning algorithm. On the other hand, when data is used from different locations then the data access time is more which increases the training time but selected data points might be diverse which can improve the convergence. Thus, sampling has a significant role in the learning algorithms. Three sampling techniques, namely, random sampling, cyclic sampling and systematic sampling are discussed below and used in learning algorithm as they are simple, easy to implement and do not involve any extra overhead and thus effective in handling large-scale problems.

\subsection{Definitions}
\label{subsec_definition}
Suppose a mini-batch $B$ of $m$ data points is to be selected from a training dataset $\lbrace \left(x_1, y_1\right),\left(x_2, y_2\right),...,\left(x_l, y_l\right) \rbrace$ of $l$ data points.
\begin{itemize}
	\item[(a)] \textit{Random Sampling:} Random sampling (RS) can be of two types, RS with replacement and RS without replacement. RS with replacement first selects one data point randomly from the whole dataset where each data point has equal probability of selection, then, similarly, second data point is selected randomly from the whole dataset where previously selected point has equal probability of selection, and so on to select $m$ data points. RS without replacement first selects one data point randomly from the whole dataset where each data point has equal probability of selection, then, similarly, second data point is selected randomly from the remaining $l-1$ points without considering the previously selected point, and so on to select $m$ data points.
	
	\item[(b)] \textit{Cyclic/Sequential Sampling:} First mini-batch is selected by taking the first $1$ to $m$ points. Second mini-batch is selected by taking next $m+1$ to $2m$ points and so on until all data points are covered. Then again start with the first data point.
	
	\item[(c)] \textit{Systematic Sampling:} \cite{William1949,William1944} It randomly selects the first point and then selects the remaining points according to a fixed pattern, e.g., it randomly selects a data point, say $i$, and then selects data points as $i, i+k,...,i+(m-1)k$ as mini-batch where $k$ is some positive integer. For simplicity, $k=1$ can be taken.
\end{itemize}
\textit{Example:} Suppose the training dataset is given by $S=\{1,2,3,....,20\}$ and size of mini-batch to be selected is $m=5$, then four mini-batches can be selected/drawn from $S$ using different sampling techniques as follows: mini-batches selected using random sampling with replacement are - $B_1=\{15,2,20,2,1\}$, $B_2=\{3,10,20,6,1\}$, $B_3=\{5,9,19,2,7\}$ and $B_4=\{1,11,18,3,16\}$; mini-batches selected using random sampling without replacement are - $B_1=\{15,2,20,11,6\}$, $B_2=\{3,10,8,14,1\}$, $B_3=\{16,4,17,7,19\}$ and $B_4=\{9,5,12,18,13\}$; mini-batches selected using cyclic sampling are - $B_1=\{1,2,3,4,5\}$, $B_2=\{6,7,8,9,10\}$, $B_3=\{11,12,13,14,15\}$ and $ B_4=\{16,17,18,19,20\}$; and mini-batches selected using systematic sampling are - $B_1=\{16,17,18,19,20\}$, $B_2=\{1,2,3,4,5\}$, $B_3=\{6,7,8,9,10\}$ and $B_4=\{11,12,13,14,15\}$. As it is clear from the above examples, in random sampling with replacement, points are selected randomly with repetition inside the mini-batch or within mini-batches. Cyclic sampling is the simplest and non-probabilistic sampling, and selects the points in a sequential manner. For systematic sampling, first point is selected randomly then the remaining points are selected back to back, here idea of replacement and without replacement can be applied between mini-batches but only sampling without replacement within mini-batches is demonstrated.\\

It is interesting to note that for RS, data points of the mini-batch are dispersed over the different sectors of the disk, so every data point needs its own seek time and latency time. Since the data is read block-wise and not content-wise so it is possible that each data point is present in a different block and thus needs its own transfer time also. For CS only one seek time is needed for one mini-batch because it starts with first data point and then moves till end and for SS one seek time is needed per mini-batch because only first element is determined randomly but rest points of the mini-batch are stored on contiguous memory locations. So, seek time is the least for CS and the most for RS. The transfer time is almost equal for CS and SS but less as compared with RS because in RS, generally, each data point needs a separate transfer time but for other case it needs as many transfer times as the number of blocks required to fit the mini-batch of data points. Thus, the overall access time to access one mini-batch of data points is minimum for CS and is maximum for RS. It is observed that RS gives the best solution as compared with the CS for a given number of epochs but the access time of RS is the most. On the other hand for CS the access time is the least for a given number of epoch but the convergence is the slowest. So there is a trade-off between reducing the data access time and convergence of learning algorithm. SS balances this trade-off since SS has the best of both techniques, like CS the data points are stored on the contiguous memory locations and like RS it has some randomness as it draws the first point randomly.
Overall, methods using CS and SS converges faster as compared with methods using RS, as discussed in Sec.~\ref{sec_experiments}.

\subsection{Learning using Systematic Sampling}
\label{subsec_methodSS}
A general learning algorithm with systematic sampling to solve large-scale problems is given by Algorithm~\ref{algo_SS}. Similar learning algorithms can be obtained for cyclic and random sampling techniques by using the corresponding sampling technique for selecting the mini-batch at step 5 of Algorithm~\ref{algo_SS}.
\begin{algorithm}
	\caption{A General Learning Algorithm with Systematic Sampling}
	\label{algo_SS}
	\begin{algorithmic}[1]
		\STATE \textbf{Inputs:} $m=\#$mini-batches and $p=$max $\# epochs$.
		\STATE \textbf{Initialize:} Take initial solution $w^0$.
		\FOR{$k=1,2,...,p$}
		\FOR{$j=1,2,...,m$}
		\STATE Select one mini-batch $B_j$ using systematic sampling without replacement.
		\STATE Formulate a subproblem using mini-batch $B_j$ as given below:\\
		$\underset{w}{\min} \;\; \dfrac{1}{|B_j|} \sum_{i \in B_j} f_i(w) + \dfrac{C}{2} \|w\|^2$
		\STATE Solve the subproblem and update the solution using appropriate method.
		\ENDFOR
		\ENDFOR
	\end{algorithmic}
\end{algorithm}
The algorithm starts with the initial solution. It divides the dataset into $m$ mini-batches using systematic sampling for selecting mini-batches. Inside the inner loop, it takes one mini-batch $B_j$, formulates a subproblem over $B_j$ and solves the sub-problem thus formed. This process is repeated until all the sub-problems over all $m$ mini-batches are solved. Then this solution process is repeated for the given number of epochs or other stopping criteria can be used in the algorithm. At step 7 of Algorithm~\ref{algo_SS}, different solvers can be used to update the solution, like, SAG (Stochastic Average Gradient) \cite{Schmidt2016}, SAGA \cite{Defazio2014}, SVRG (Stochastic Variance Reduced Gradient) \cite{Johnson2013}, SAAG-II (Stochastic Average Adjusted Gradient) \cite{Chauhan2017Saag} and MBSGD (Mini-Batch Stochastic Gradient Descent) \cite{Chauhan2017Saag,Cotter2011,Shalev-Shwartz2011}.

\section{Theoretical Analysis}
\label{sec_analysis}
The proposed sampling techniques are simple but effective for solving large-scale learning problems with different solvers, as discussed in Section \ref{sec_experiments}. The convergence proof of learning algorithms have been provided using the simplest solver MBSGD (mini-batched Stochastic Gradient Descent) method with constant step size, for the simplicity of proofs as the focus of study is only on sampling techniques and not on solvers. $l2$-regularized ERM problem has been solved under the following assumptions to demonstrate the efficacy of proposed sampling techniques. It is assumed that the regularization term is hidden inside the loss function term for notational convenience otherwise it needs to write separate gradients for regularization term.

\begin{assumption}[LIPSCHITZ CONTINUOUS GRADIENT]
	\label{assump_lipschitz}
	Suppose function $f: \mathbb{R}^n \rightarrow \mathbb{R}$ is convex and differentiable on $S$, and that gradient $\nabla f$ is $L$-Lipschitz-continuous, where $L>0$ is Lipschitz constant, then, we have,
	\begin{equation}
	\label{eq_lipschitz1}
	\| \nabla f(y) - \nabla f(x)\| \le L \|y-x\|,
	\end{equation}
	\begin{equation}
	\label{eq_lipschitz2}
	\begin{array}{ll}
	\text{and},\quad f(y) \le f(x) +\nabla f(x)^{T}(y-x)+ \dfrac{L}{2} \|y-x\|^2.
	\end{array}
	\end{equation}
\end{assumption}

\begin{assumption}[STRONG CONVEXITY]
	\label{assump_sconvexity}
	Suppose function $f: \mathbb{R}^n \rightarrow \mathbb{R}$ is $\mu$-strongly convex function for $\mu>0$ on $S$ and $p^{*}$ is the optimal value of $f$, then, we have,
	\begin{equation}
	\label{convexivity_eq1}
	f(y) \ge f(x) +\nabla f(x)^T(y-x) + \dfrac{\mu}{2} \|y-x\|^2,
	\end{equation}
	\begin{equation}
	\label{eq_convex1}
	\begin{array}{ll}
	\text{and},\quad & f(x) - p^{*} \le \dfrac{1}{2\mu}\|\nabla f(x)||^2
	\end{array}
	\end{equation}
\end{assumption}

\begin{theorem1}
	\label{theorem}
	Suppose for function given by eq.~(\ref{eq_erm}), under Assumptions~\ref{assump_lipschitz}, \ref{assump_sconvexity} and constant step size $\alpha$, and taking solver MBSGD, Algorithm~\ref{algo_SS}, converges linearly in expectation for cyclic, systematic and sequential sampling techniques.
\end{theorem1}
\begin{proof}
	By definition of MBSGD, we have,
	\begin{equation}
	\label{eq_mbsgd}
	w^{k+1} = w^k - \dfrac{\alpha}{|B_j|} \sum\limits_{i\in B_j}\nabla f_i (w^k)
	\end{equation}
	By $L$-Lipschitz continuity of gradients,
	\begin{equation*}
	\begin{array}{ll}
	f(w^{k+1}) &\le  f(w^k) + \nabla f (w^k)^T \left( w^{k+1}-w^k \right) +\dfrac{L}{2} \left\|w^{k+1}-w^k\right\|^2\\
	&= f(w^k) - \alpha \nabla f(w^k)^T\left[ \dfrac{1}{|B_j|} \sum_{i\in B_j} \nabla f_i (w^k) \right]
	+ \dfrac{L\alpha^2}{2} \left\|  \dfrac{1}{|B_j|} \sum_{i \in B_j}\nabla f_i(w^k) \right\|^2,
	\end{array}
	\end{equation*}
	equality follows from the definition of MBSGD.\\
	
	\textit{Case-I: Mini-batch $B_j$ is selected using random sampling (RS) without replacement or Systematic Sampling (SS) without replacement}\\
	
	Taking expectation on both sides over mini-batches $B_j$ and subtracting optimal value ($p^{*}$), we have,
	\begin{equation*}
	\begin{array}{ll}
	E_{B_j} \left[ f(w^{k+1}) - p^{*} \right] &\le f(w^k) - p^{*} -\alpha \nabla f(w^k)^T E_{B_j} \left[ \dfrac{1}{|B_j|} \sum_{i \in B_j} \nabla f_i(w^k) \ \right] \\ &+ \dfrac{L\alpha^2}{2}  E_{B_j} \left\|\dfrac{1}{|B_j|} \sum_{i \in B_j} \nabla f_i(w^k) \right\|^2\\
	&\le f(w^k) - p^{*} -\alpha \nabla f(w^k)^T \nabla f(w^k) + \dfrac{L\alpha^2}{2} \left\| R_0 \right\|^2,
	\end{array}
	\end{equation*}
	inequality follows using $E_{B_j} \left[ \dfrac{1}{|B_j|} \sum_{i \in B_j} \nabla f_i(w^k) \ \right] = \nabla f(w^k)$ and taking $\left\|\dfrac{1}{|B_j|} \sum_{i \in B_j} \nabla f_i(w^k) \right\|^2 \le R_0, \; \forall j, w $.
	\begin{equation*}
	\begin{array}{ll}
	E_{B_j} \left[ f(w^{k+1}) - p^{*} \right] &\le f(w^k) - p^{*} -\alpha \left\|\nabla f(w^k)\right\|^2 +\dfrac{L\alpha^2R_0 ^2}{2},\\
	&\le f(w^k) - p^{*} -\alpha.2\mu \left(f(w^k) - p^{*}\right) + \dfrac{L\alpha^2R_0 ^2}{2},
	\end{array}
	\end{equation*}
	inequality follows from strong convexity results.
	\begin{equation*}
	E_{B_j} \left[ f(w^{k+1}) - p^{*} \right] \le \left( 1 - 2\alpha\mu \right) \left( f(w^k) - p^{*}\right) + \dfrac{L\alpha^2R_0 ^2}{2}
	\end{equation*}
	Applying inequality recursively, we have,
	\begin{equation*}
	\begin{array}{ll}
	E_{B_j} \left[ f(w^{k+1}) - p^{*} \right] &\le \left( 1 - 2\alpha\mu \right)^{k+1} \left( f(w^0) - p^{*}\right) + \dfrac{L\alpha^2 R_0 ^2}{2} \sum_{i=0}^{k} \left( 1 - 2\alpha\mu \right)^{i}\\
	&\le \left( 1 - 2\alpha\mu \right)^{k+1} \left( f(w^0) - p^{*}\right) + \dfrac{L\alpha^2 R_0 ^2}{2}.\dfrac{1}{2\alpha\mu},
	\end{array}
	\end{equation*}
	inequality follows since $\sum_{i=0}^{k} r^i \le \sum_{i=0}^{\infty} r^i = \dfrac{1}{1-r}, \quad \|r\|<1$.
	\begin{equation*}
	E_{B_j} \left[ f(w^{k+1}) - p^{*} \right] \le \left( 1 - 2\alpha\mu \right)^{k+1} \left( f(w^0) - p^{*}\right) + \dfrac{L\alpha R_0 ^2}{4\mu}.
	\end{equation*}
	Thus, algorithm converges linearly with initial error proportional to $\alpha$.\\
	
	\textit{Case-II: Mini-batch $B_j$ is selected using Cyclic/Sequential Sampling (SS)}\\
	
	Taking summation over number of mini-batches and dividing by number of mini-batches ($m$), and subtracting $p^{*}$, we have,
	\begin{equation*}
	\begin{array}{l}
	\dfrac{1}{m} \sum_{j=1}^{m}\left[ f(w^{k+1}) - p^{*} \right] \\ \le \dfrac{1}{m} \sum_{j=1}^{m}\left[ f(w^k) -p^{*}\right] 
	- \alpha \nabla f(w^k)^T \dfrac{1}{m} \sum_{j=1}^{m} \left[ \dfrac{1}{\|B_j|} \sum_{i\in B_j} \nabla f_i (w^k) \right]\\ + \dfrac{L\alpha^2}{2} \dfrac{1}{m} \sum_{j=1}^{m} \left\|  \dfrac{1}{|B_j|} \sum_{i \in B_j}\nabla f_i(w^k) \right\|^2,\\
	\le  f(w^k) -p^{*}- \alpha \nabla f(w^k)^T \nabla f(w^k)+ \dfrac{L\alpha^2}{2}   R_0^2,
	\end{array}
	\end{equation*}
	inequality follows since $\dfrac{1}{m} \sum_{j=1}^{m} \left[ \dfrac{1}{\|B_j|} \sum_{i\in B_j} \nabla f_i (w^k) \right] = \nabla f(w^k)$\\ and $\left\|  \dfrac{1}{|B_j|} \sum_{i \in B_j}\nabla f_i(w^k) \right\| \le R_0$.
	\begin{equation*}
	\dfrac{1}{m} \sum_{j=1}^{m}\left[ f(w^{k+1}) - p^{*} \right] \le \left( 1 - 2\alpha\mu \right)^{k+1} \left( f(w^0) - p^{*}\right) + \dfrac{L\alpha R_0 ^2}{4\mu},
	\end{equation*}
	inequality follows from the Case-I derivation. Thus, algorithm converges linearly with initial error proportional to $\alpha$. Hence, theorem is proved.
	
\end{proof}

\section{Experiments}
\label{sec_experiments}

\subsection{Experimental Setup}
\label{subsec_setup}
Experiments have been performed using five methods, namely, SAG, SAGA, SAAG-II, SVRG and MBSGD with eight bench marked datasets\footnote{datasets used in the experiments are available at: \url{https://www.csie.ntu.edu.tw/\~cjlin/libsvmtools/datasets/}} as presented in the Table~\ref{tab_datasets}. Each method has been run with two mini-batches of size $500$ and $1000$ data points, and two techniques to find step size, namely, constant step size method and backtracking line search (LS) method for a predefined number of epochs. Constant step size method uses Lipschitz constant $L$ and takes step size as $1/L$ for all methods. Backtracking line search is performed approximately only using the selected mini-batch of data points because performing backtracking line search on whole dataset could hurt the convergence of learning algorithm for large-scale problems by taking huge time. For one dataset, one method runs for $12$ times $(3(\text{sampling\;techniques})*2 (\text{mini-batches})*2(\text{step\;size\;finding\;techniques})) $, and for one dataset, three sampling techniques are compared on $20$ different settings $(5(\text{methods})*2 (\text{mini-batches})*2 (\text{step\;size\;finding\;techniques}))$. Thus $160 (20*8(\text{datasets}))$ settings have been used to compare the results for three sampling techniques. As training time depends on the configuration of machine over which experiments are performed so this is to be noted that all the experiments have been conducted on MacBook Air (8GB 1600 MHz DDR3, 256GB SSD, 1.6 GHz Intel Core i5).
\begin{table}[h]
	\centering
	\caption{Datasets used in experimentation}
	\label{tab_datasets}
	\begin{tabular}{|l|r|r|r|}
		\hline
		Dataset & \#classes & \#features & \#datapoints\\
		\hline
		HIGGS & 2 &  28 & 11,000,000\\
		SUSY & 2 & 18 & 5,000,000\\
				SensIT Vehicle & 3 & 100 & 78,823\\
				(combined) & & &\\
				mnist & 10 & 780 & 60,000\\
		protein & 3 & 357 & 17,766\\
		rcv1.binary & 2 & 47,236  & 20,242\\
				covtype.binary & 2 & 54 & 581,012\\
		ijcnn1 & 2 & 22 & 49,990\\
		\hline
	\end{tabular}
\end{table}

\subsection{Implementation Details}
\label{subsec_implementationDetails}
For each sampling technique, i.e., Random Sampling (RS), Cyclic Sampling (CS) and Systematic Sampling (SS), same algorithmic structure is used with difference only in selecting the mini-batch for each sampling technique. And for each sampling technique, the dataset is divided into predefined number of mini-batches, as per the Algorithm \ref{algo_SS}. For simplicity, dataset has been divided into equal sized mini-batches except the last mini-batch which might has data points less than or equal to other mini-batches.
For RS, during each epoch, an array of size equal to the number of data points in the dataset is taken and this array contains the randomized indexes of data points. To select the mini-batches, array contents equal to mini-batch size or till the end of array, are selected sequentially.
For CS, during each epoch, an array of size equal to the number of data points in the dataset is taken, containing indexes of data points in sorted order. To select the mini-batches, array contents equal to mini-batch size or till the end of array, are selected sequentially.
For SS, during each epoch, an array of size equal to the number of mini-batches is taken and this array contains the randomized indexes of mini-batches. To select a mini-batch, an array element is selected in the sequence. This array element gives us the first index of data point in the selected mini-batch. The other data points are selected sequentially from the starting index of the mini-batch equal to the size of a mini-batch or till the last data point in the dataset.

\subsection{Results}
\label{subsec_results}
\begin{figure}[htb]
	\subfloat{\includegraphics[width=.25\linewidth]{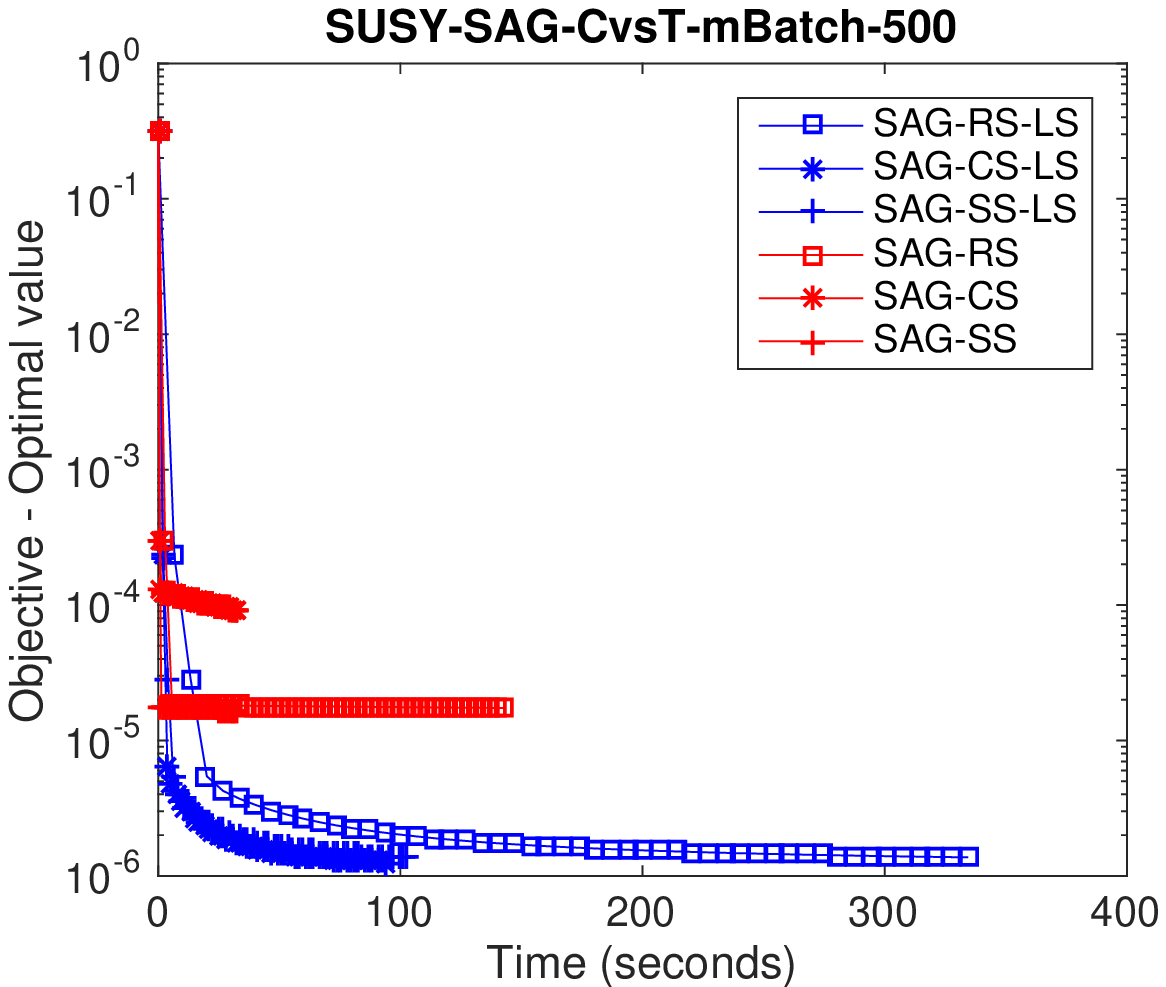}}
	\subfloat{\includegraphics[width=.25\linewidth]{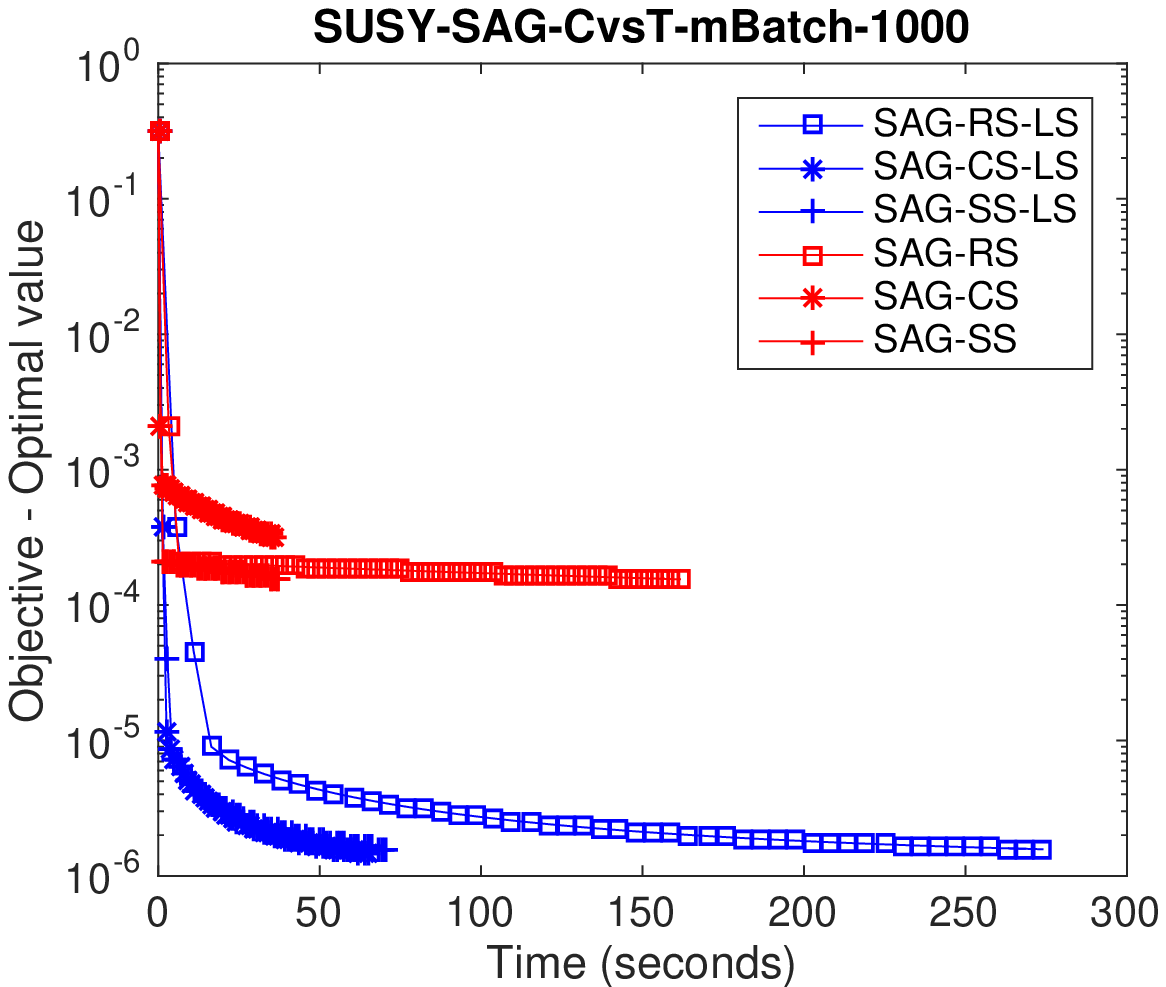}}
	\subfloat{\includegraphics[width=.25\linewidth]{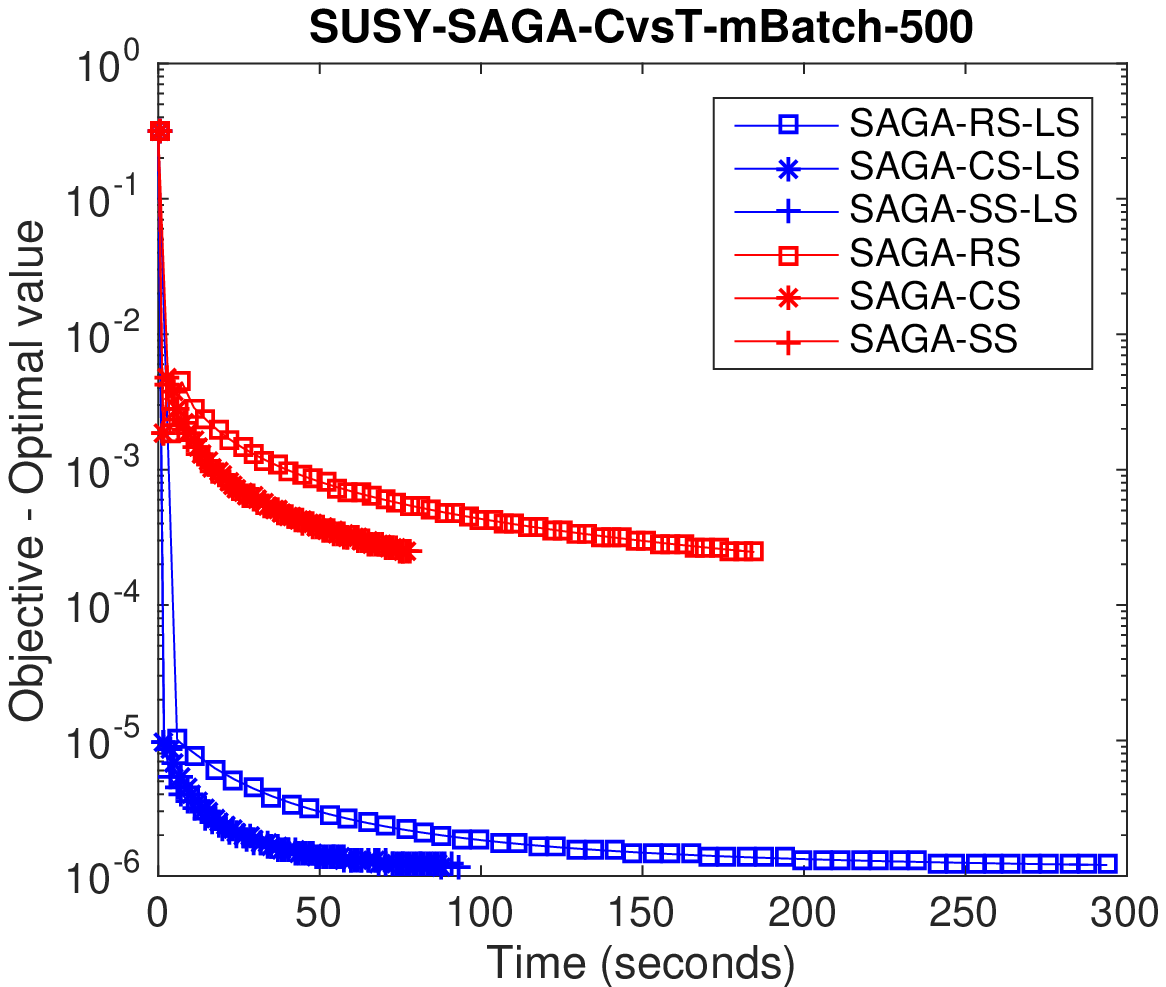}}
	\subfloat{\includegraphics[width=.25\linewidth]{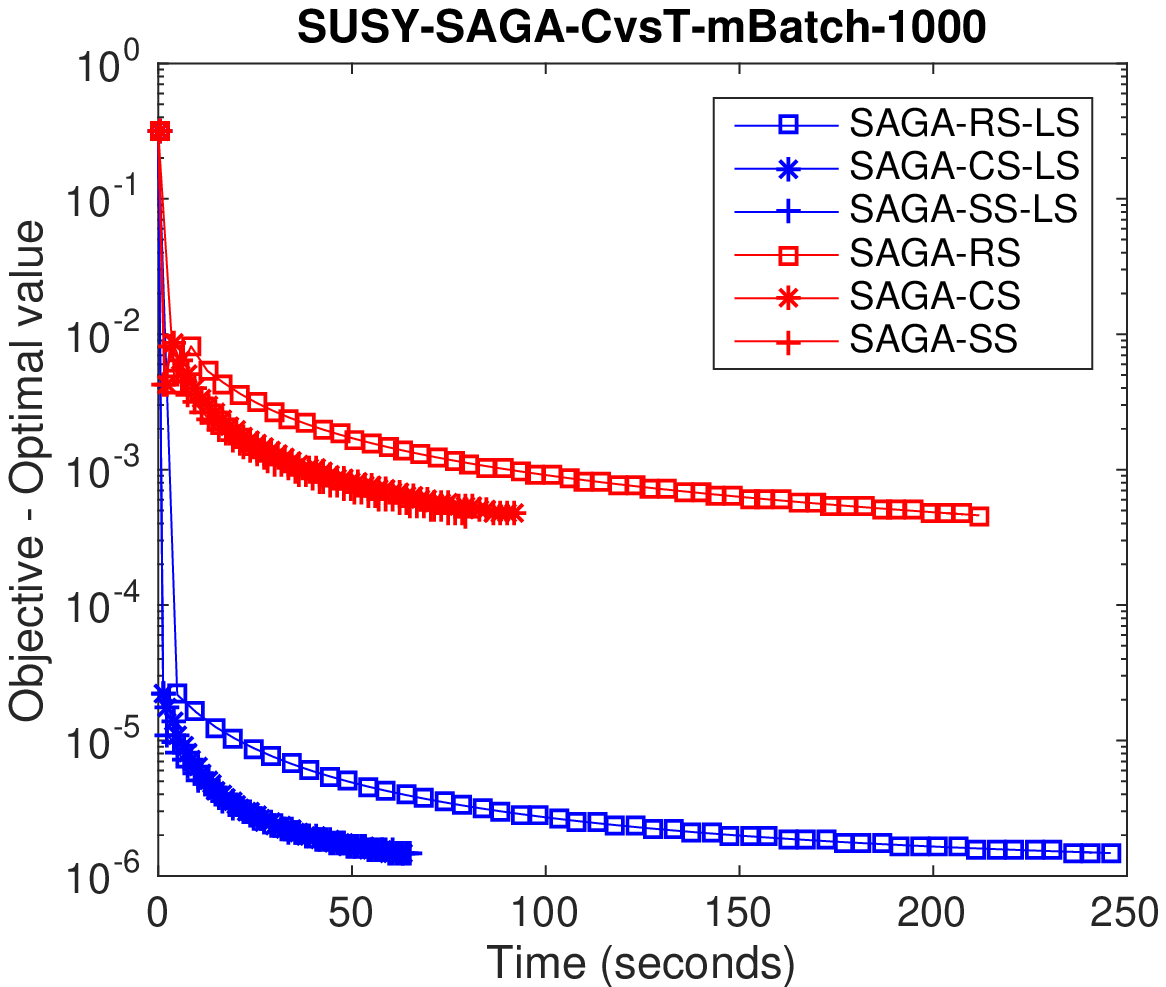}}
	
	\subfloat{\includegraphics[width=.25\linewidth]{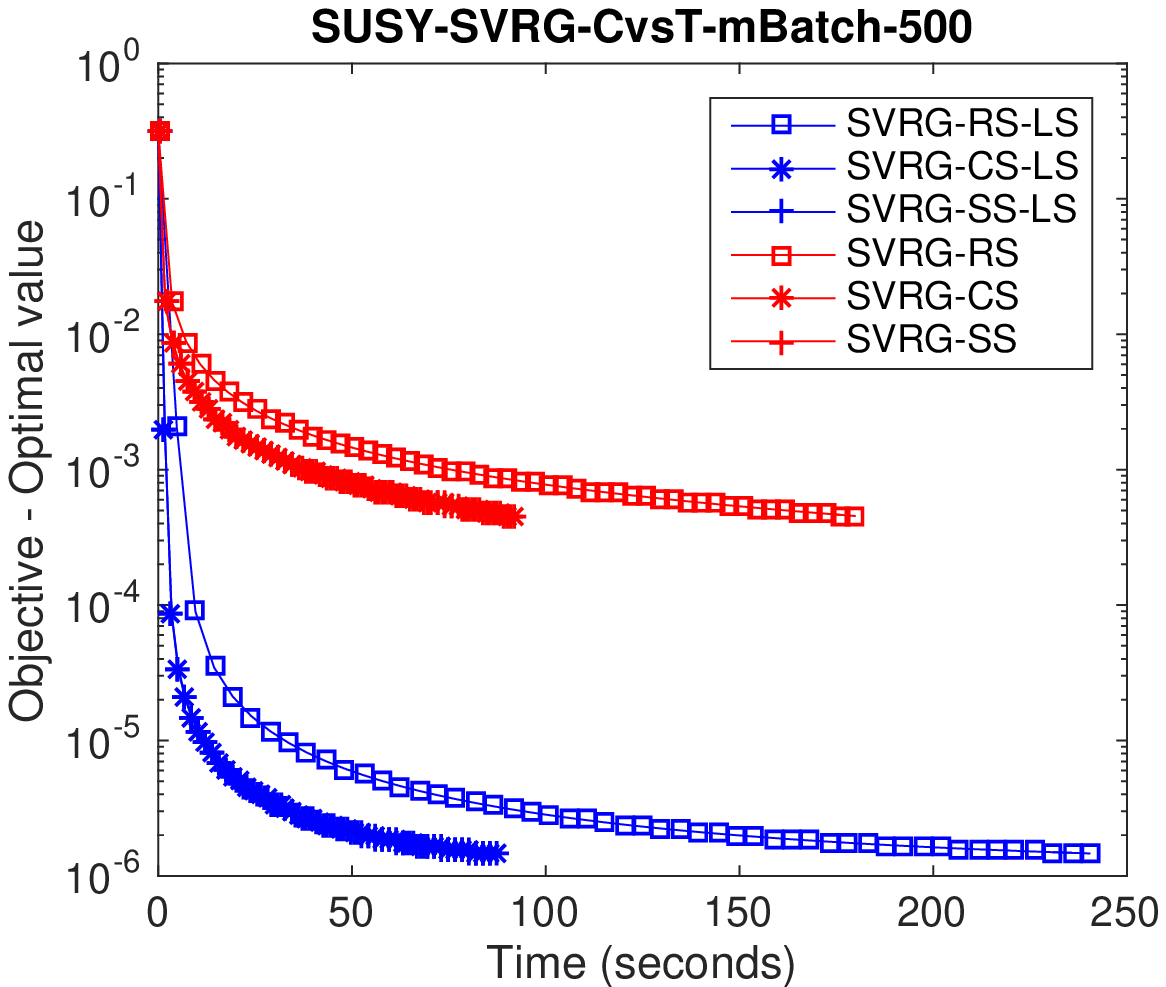}}
	\subfloat{\includegraphics[width=.25\linewidth]{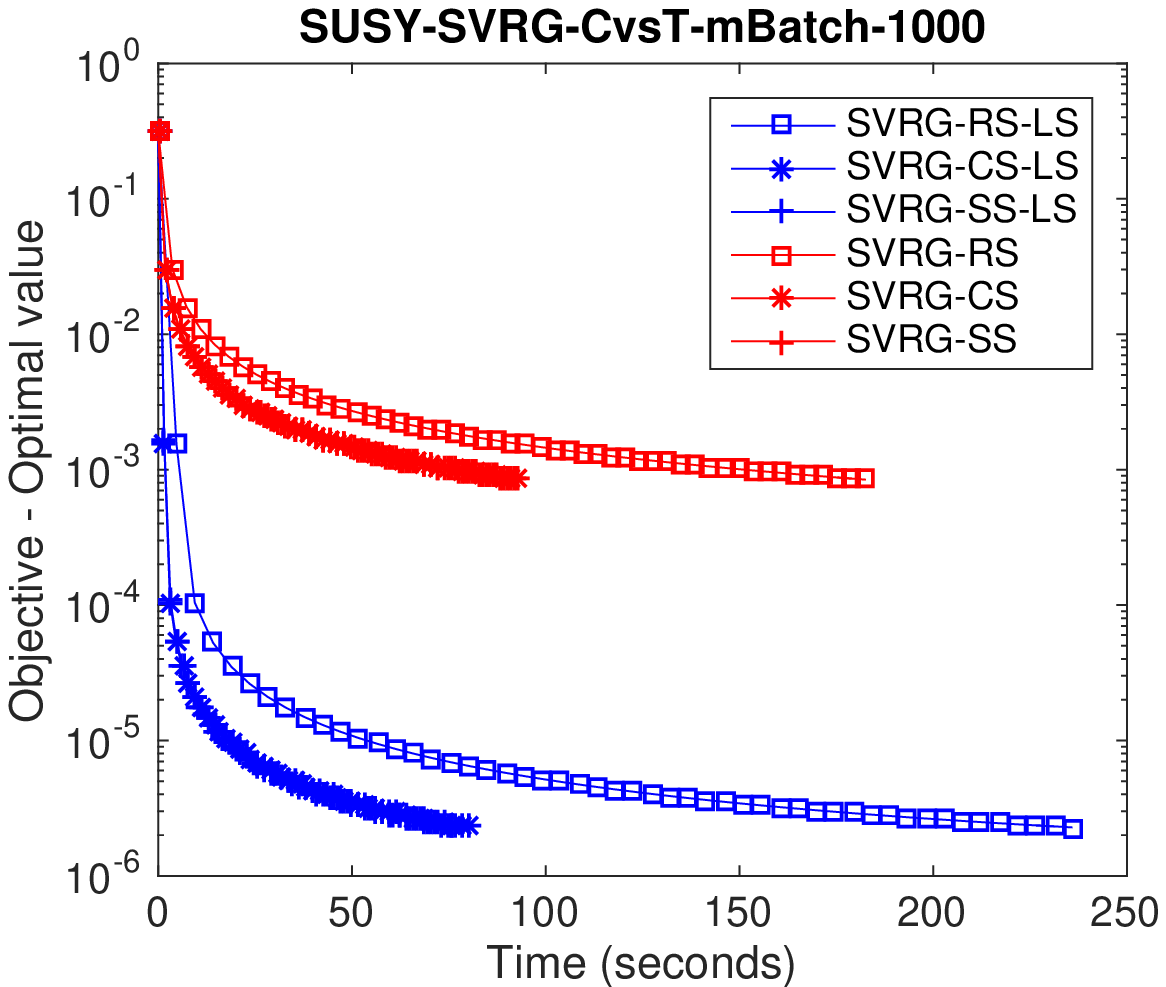}}
	\subfloat{\includegraphics[width=.25\linewidth]{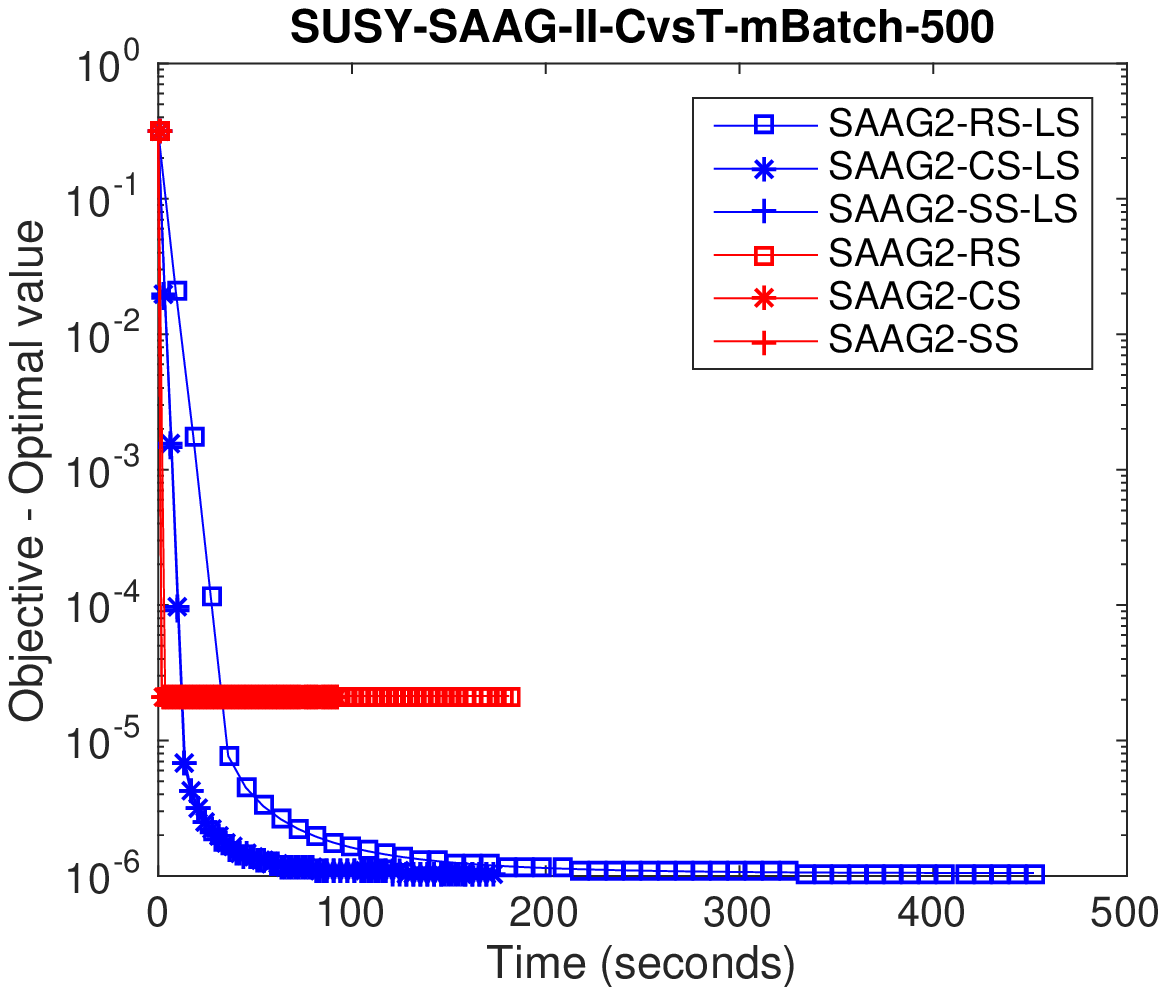}}
	\subfloat{\includegraphics[width=.25\linewidth]{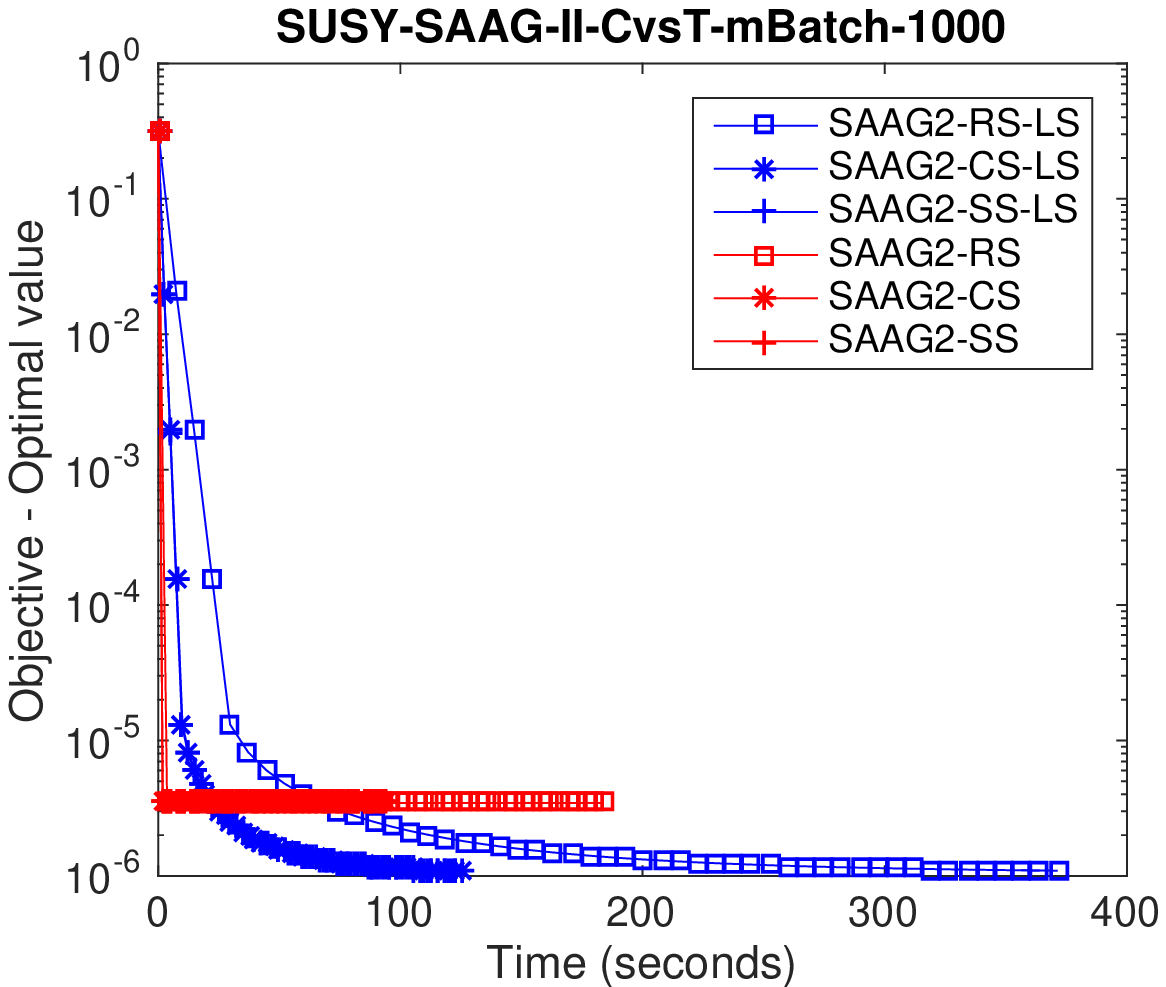}}
	
	\subfloat{\includegraphics[width=.25\linewidth]{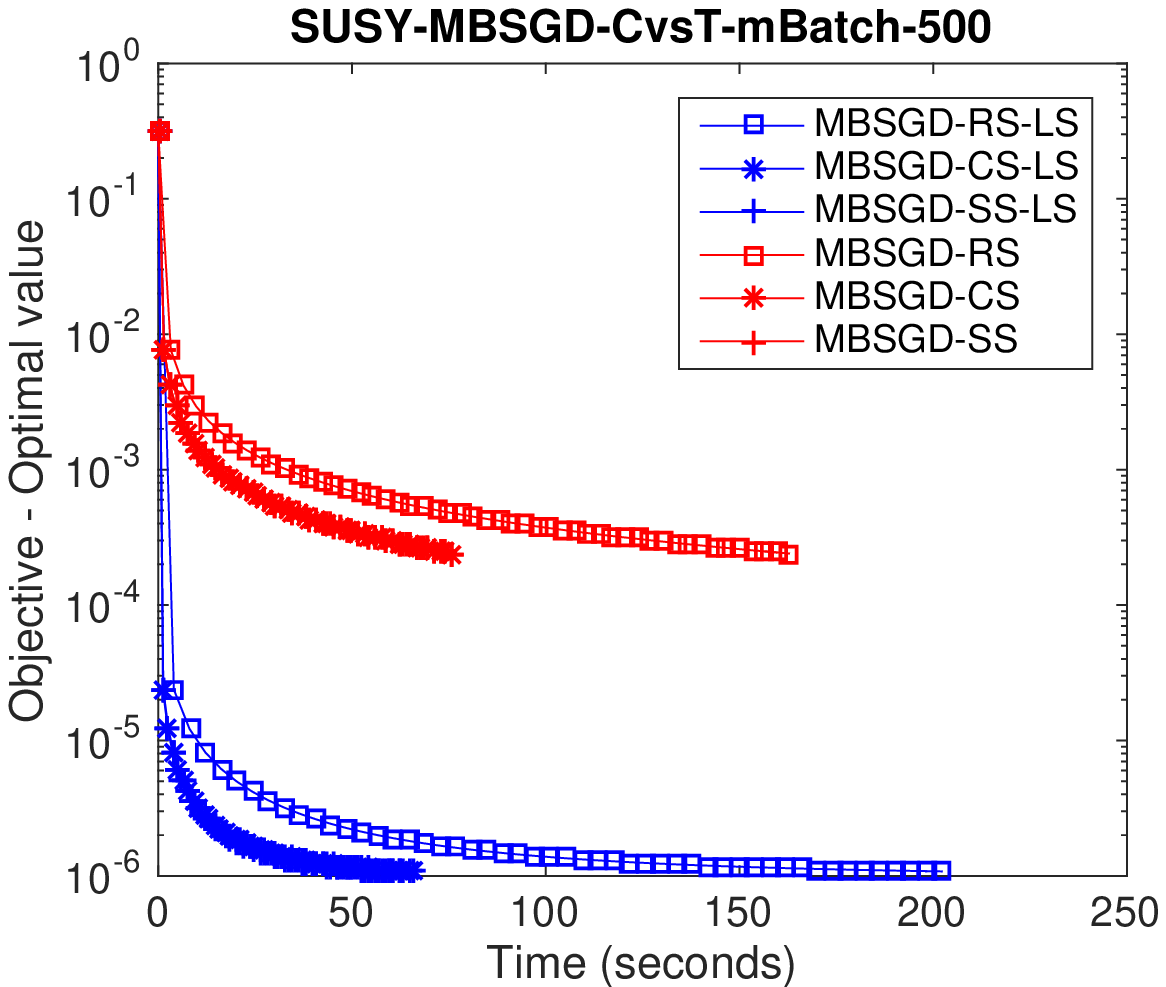}}
	\subfloat{\includegraphics[width=.25\linewidth]{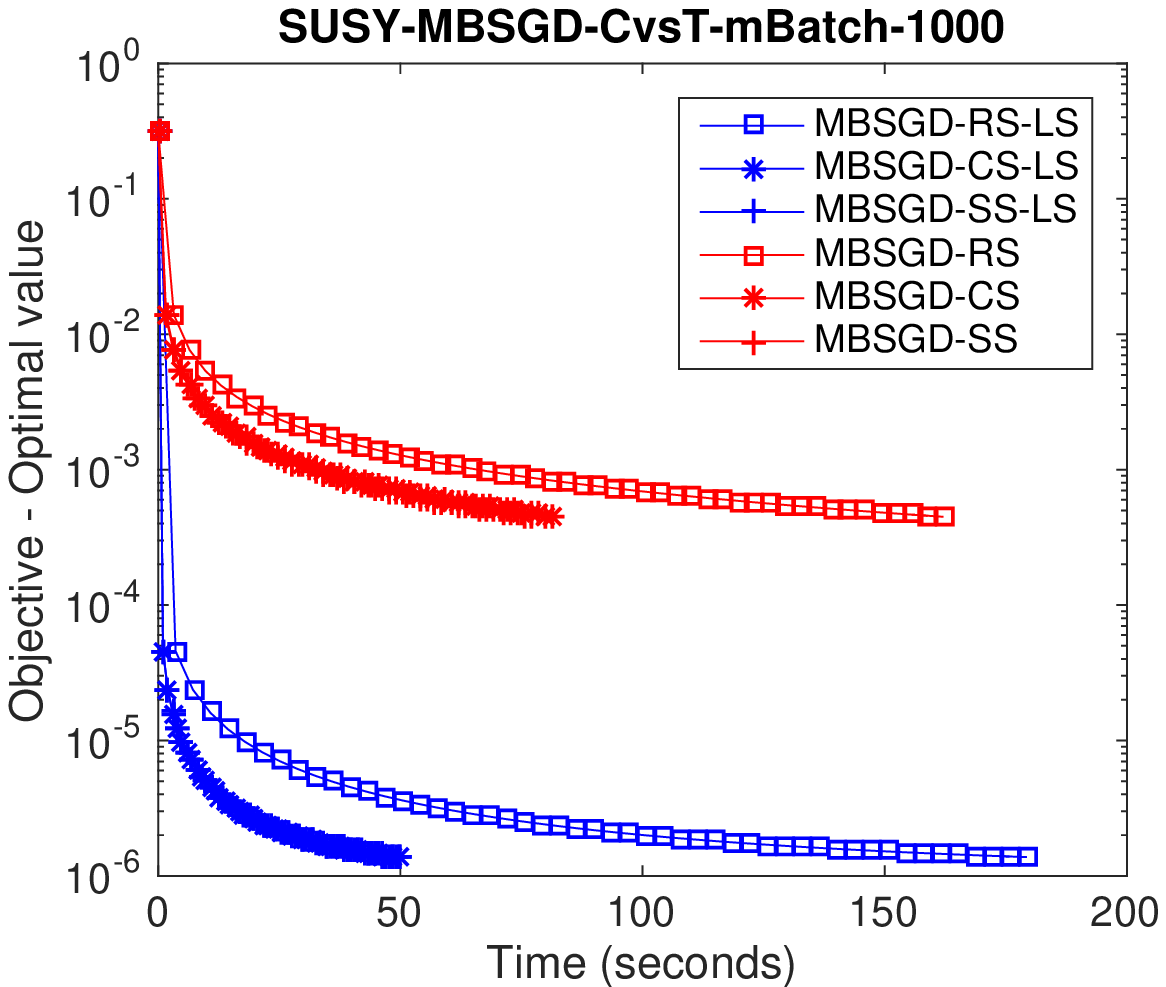}}
	\subfloat{\includegraphics[width=.25\linewidth]{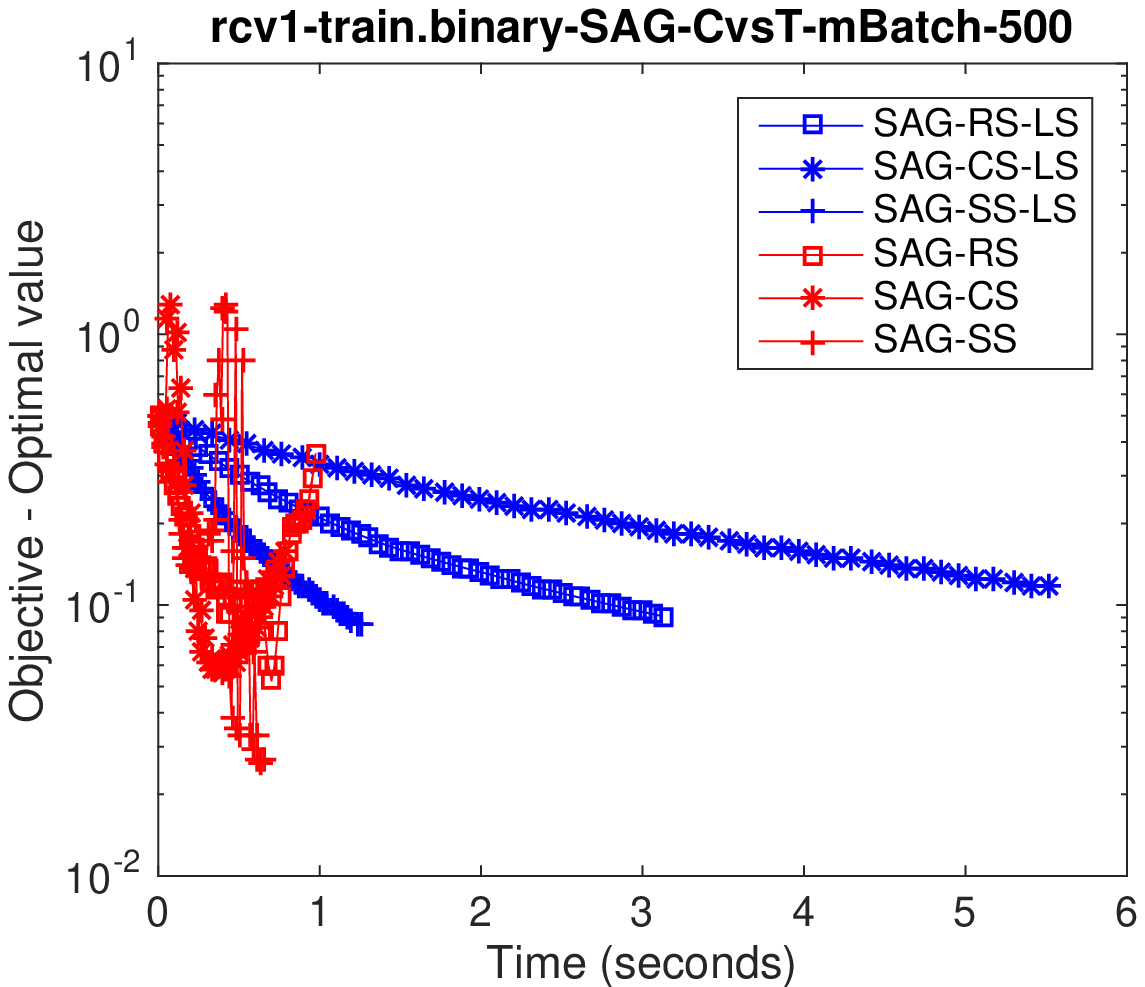}}
	\subfloat{\includegraphics[width=.25\linewidth]{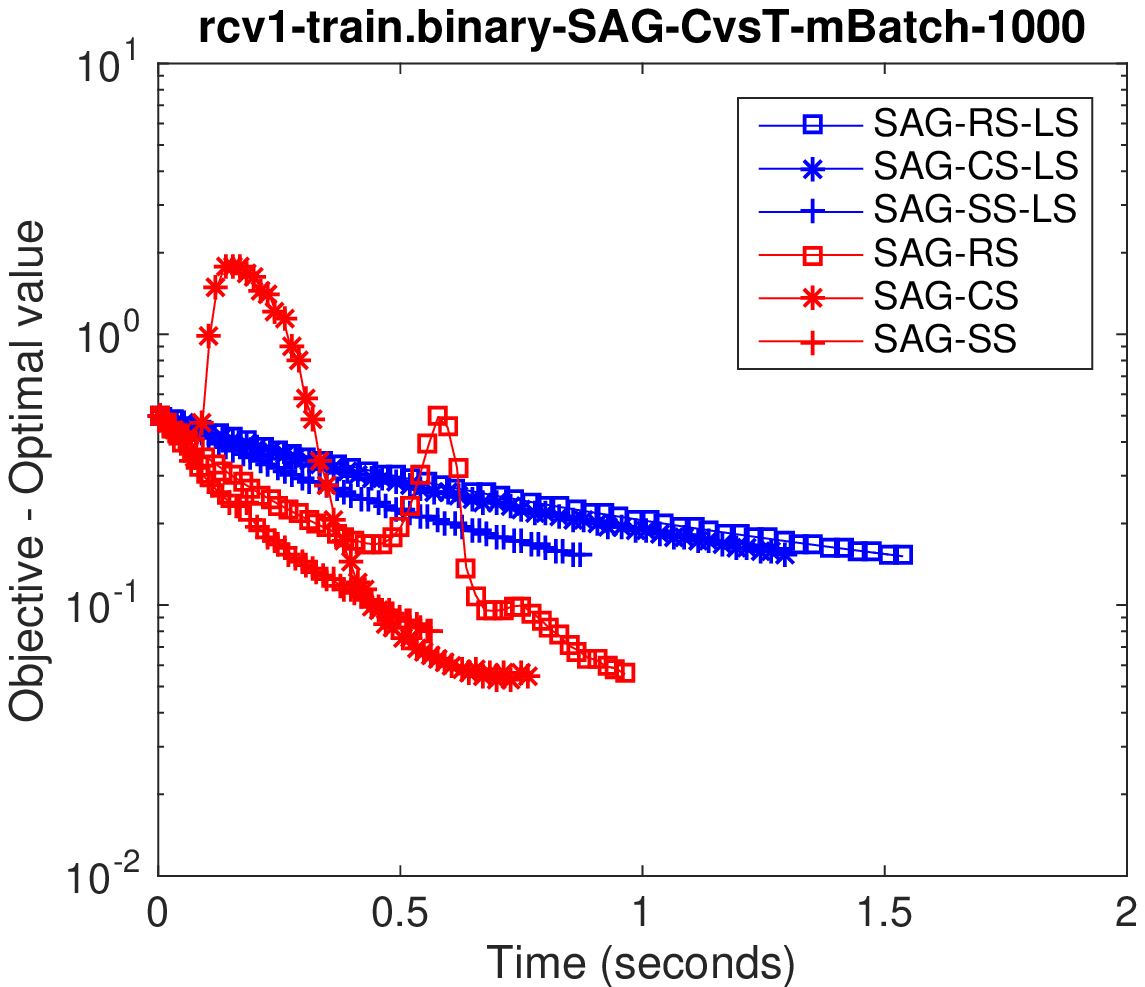}}
	
	\subfloat{\includegraphics[width=.25\linewidth]{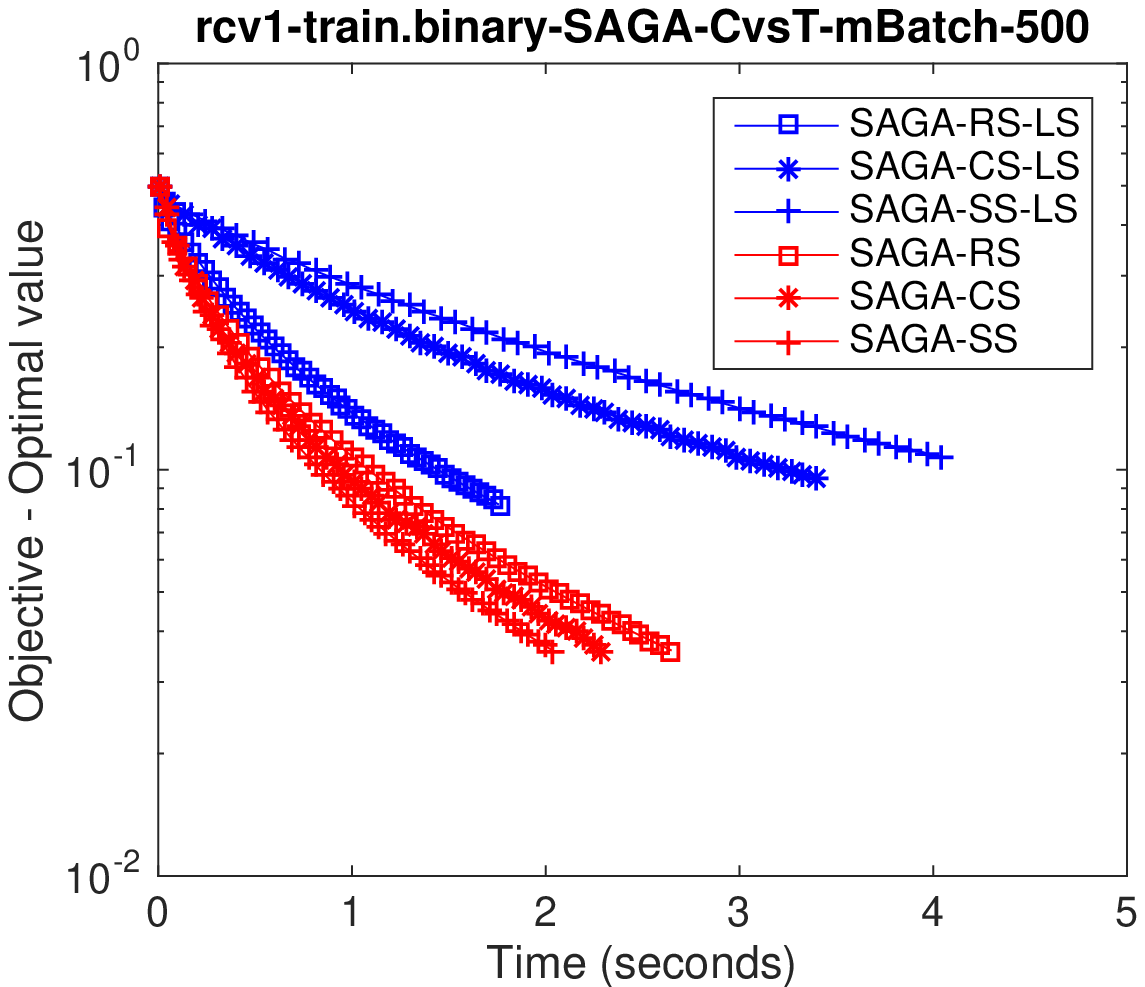}}
	\subfloat{\includegraphics[width=.25\linewidth]{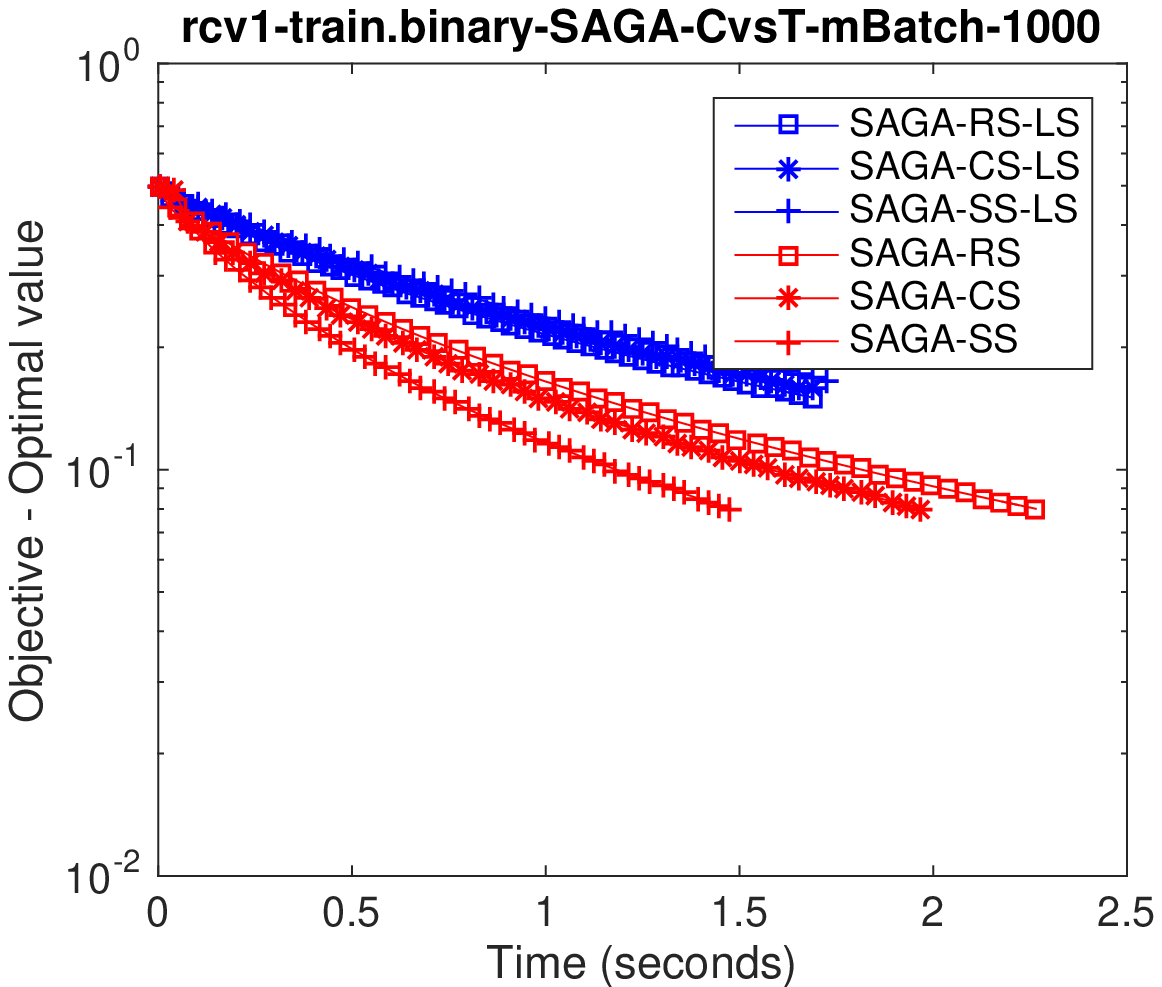}}
	\subfloat{\includegraphics[width=.25\linewidth]{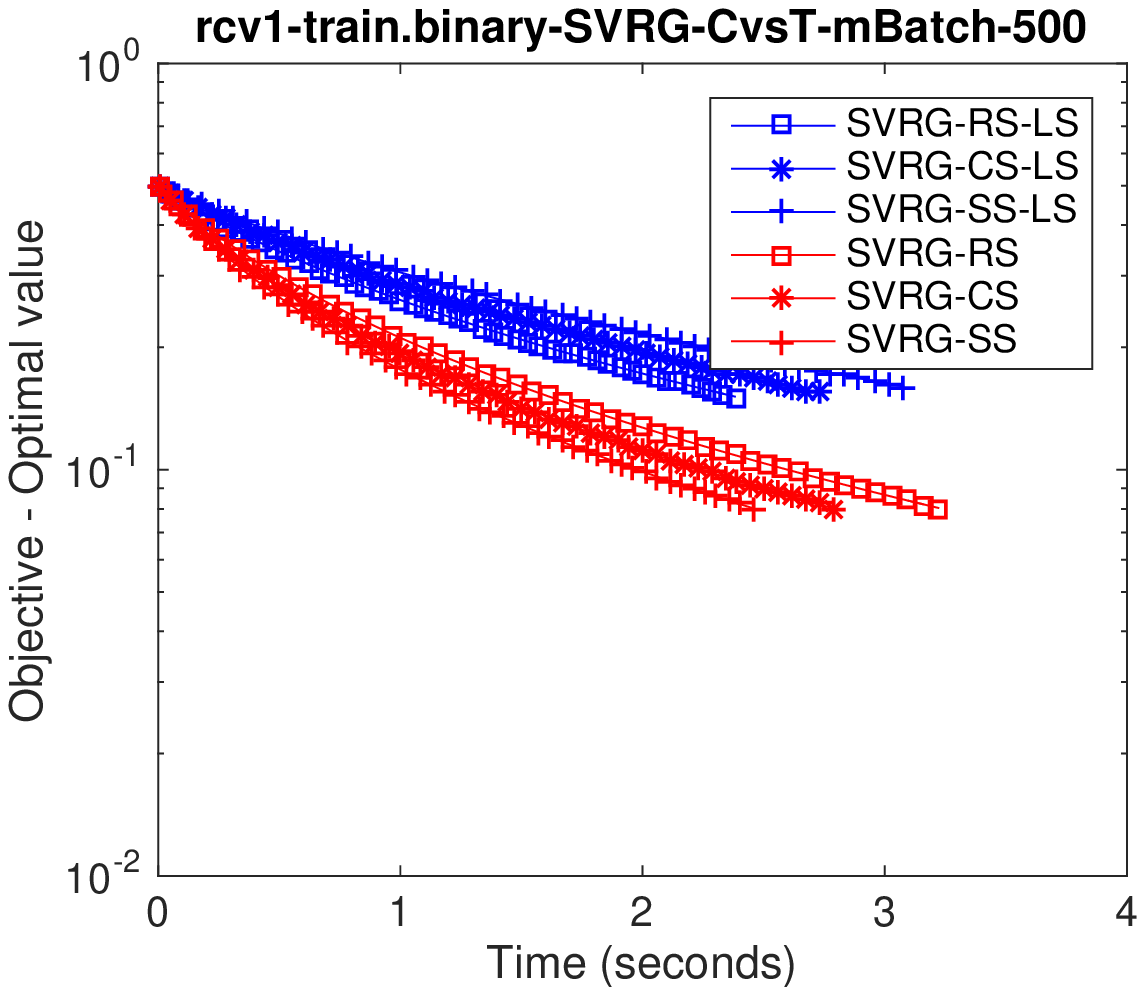}}
	\subfloat{\includegraphics[width=.25\linewidth]{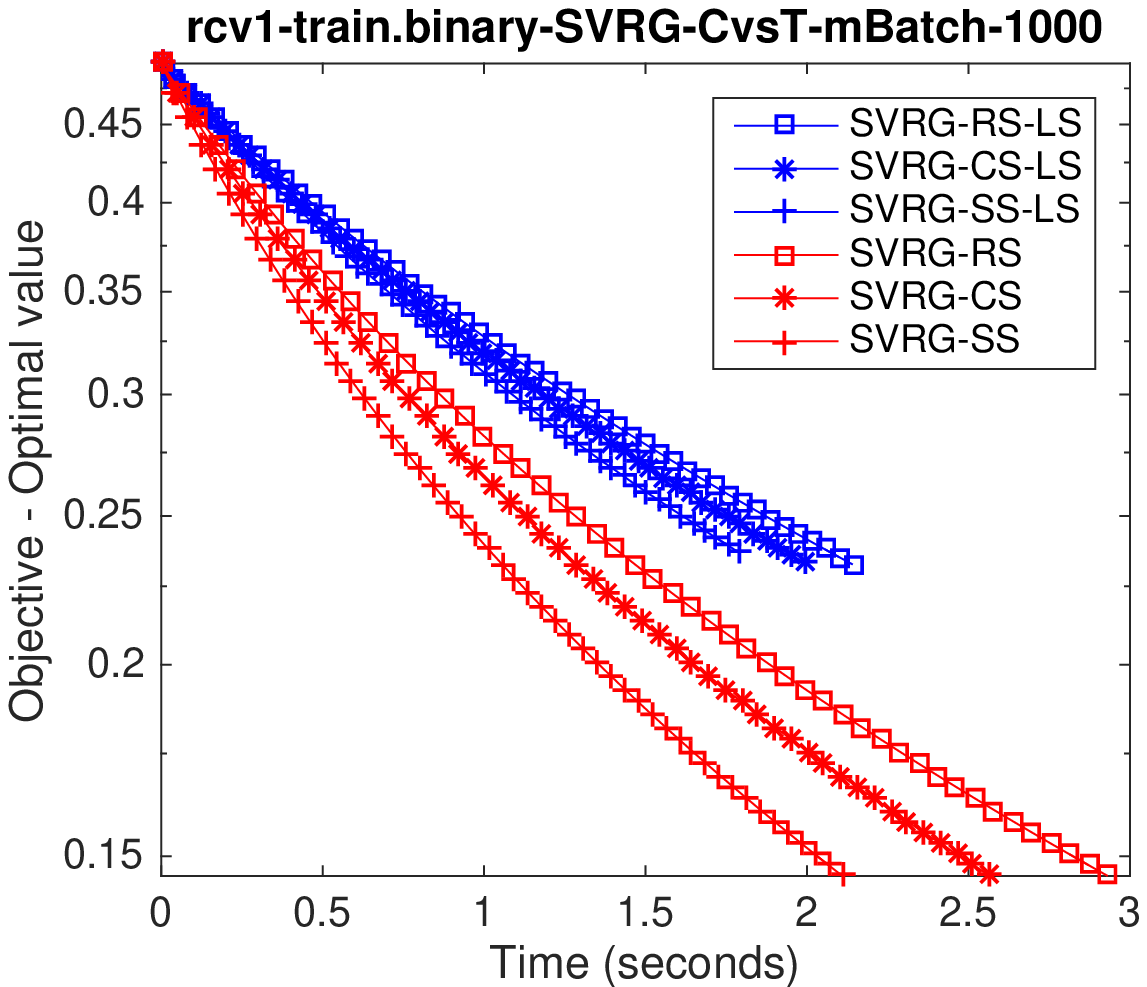}}
	
	\subfloat{\includegraphics[width=.25\linewidth]{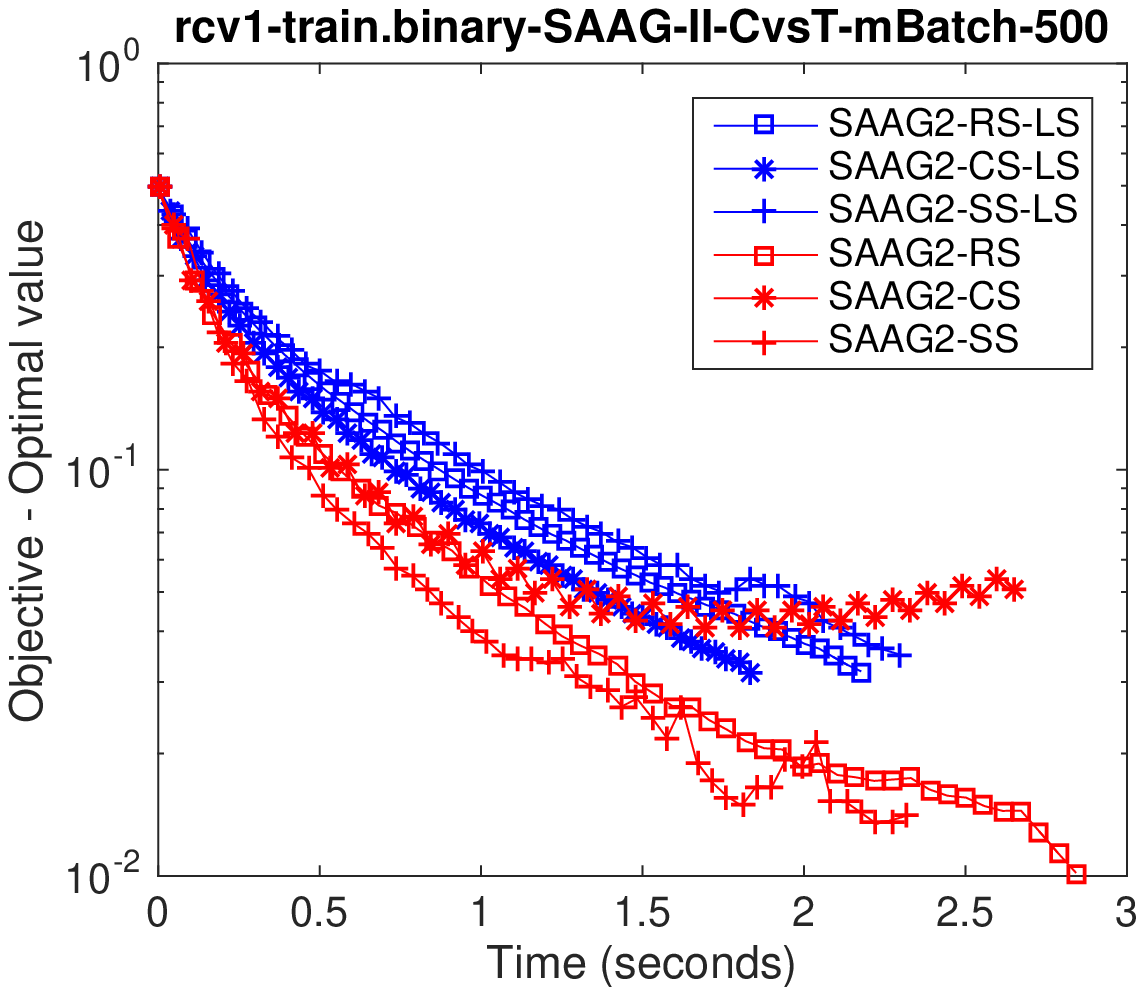}}
	\subfloat{\includegraphics[width=.25\linewidth]{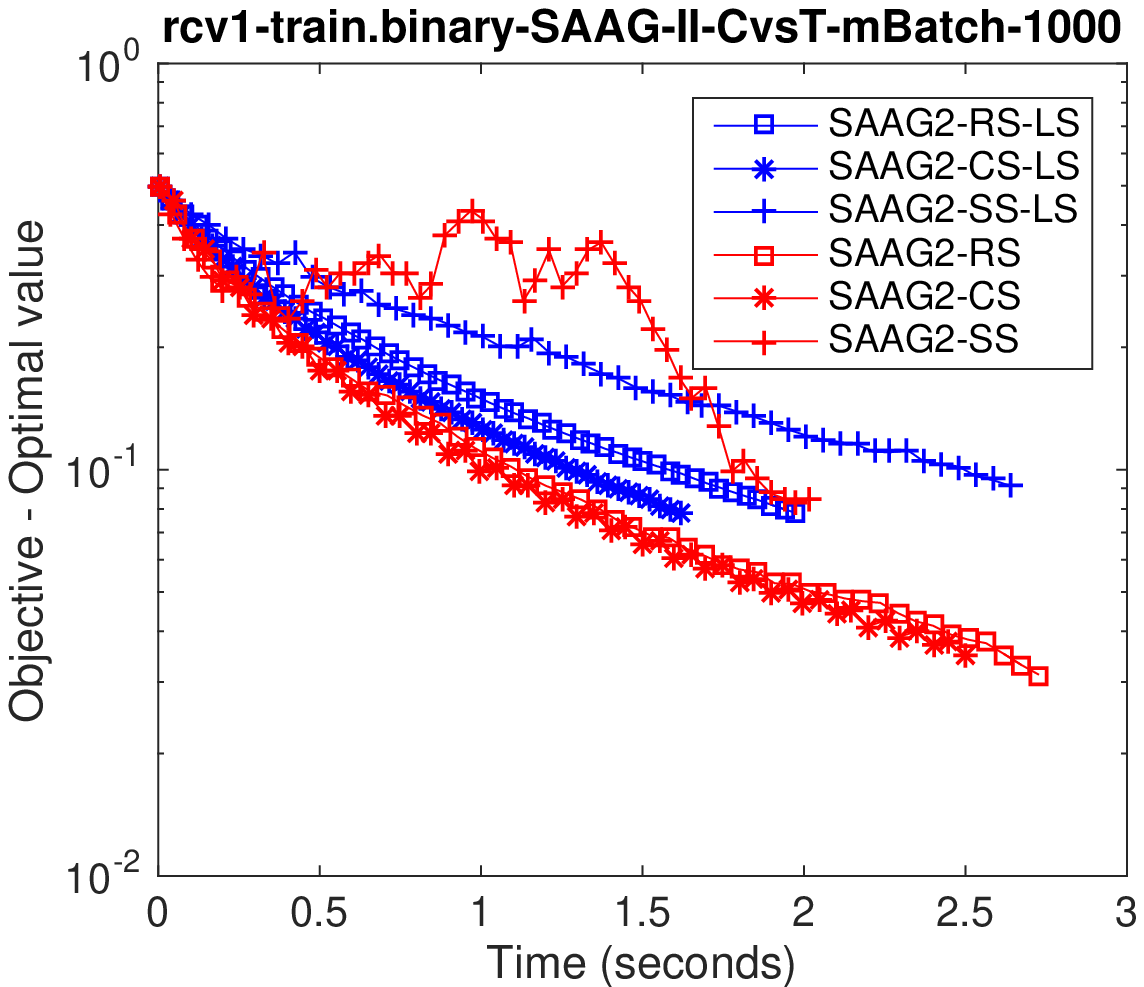}}
	\subfloat{\includegraphics[width=.25\linewidth]{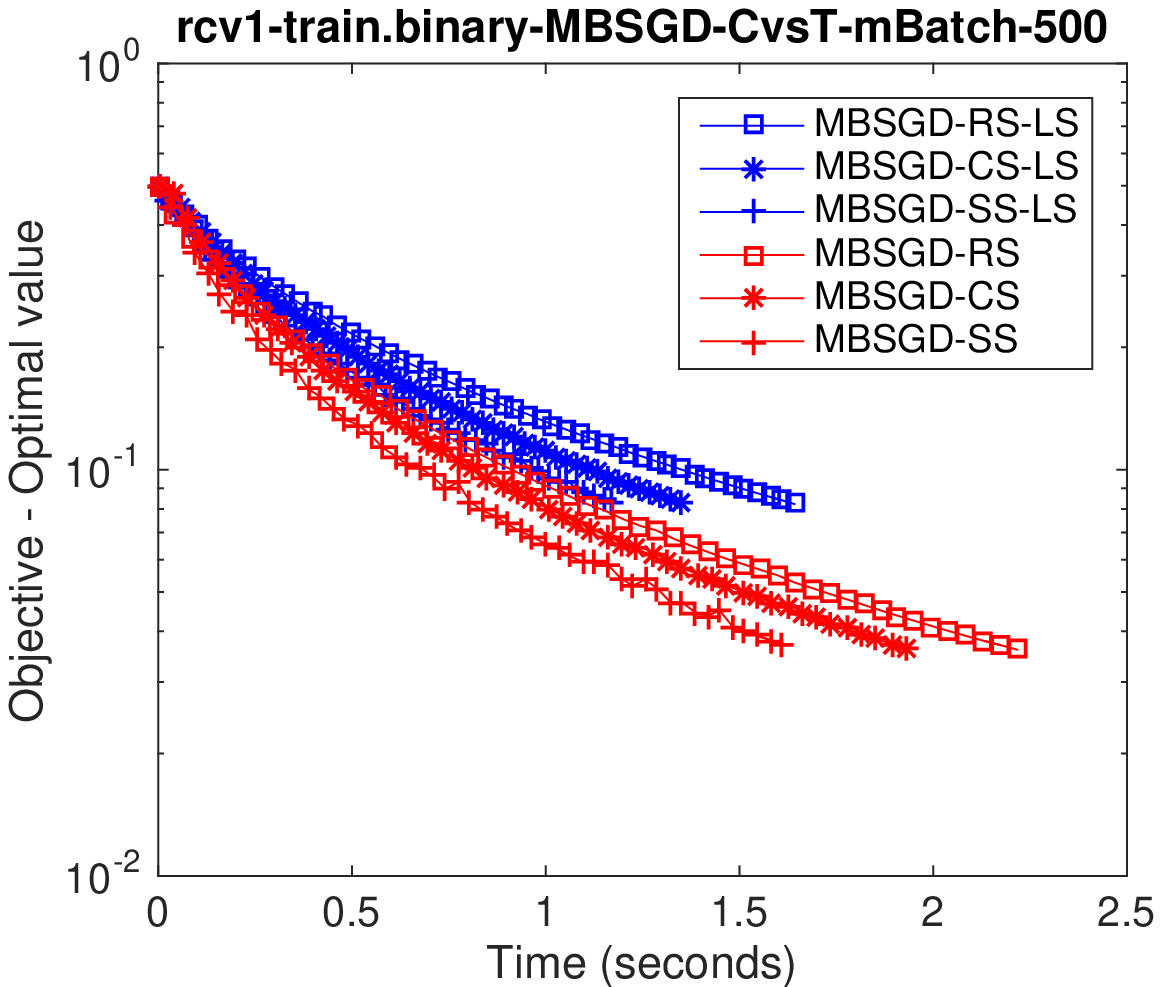}}
	\subfloat{\includegraphics[width=.25\linewidth]{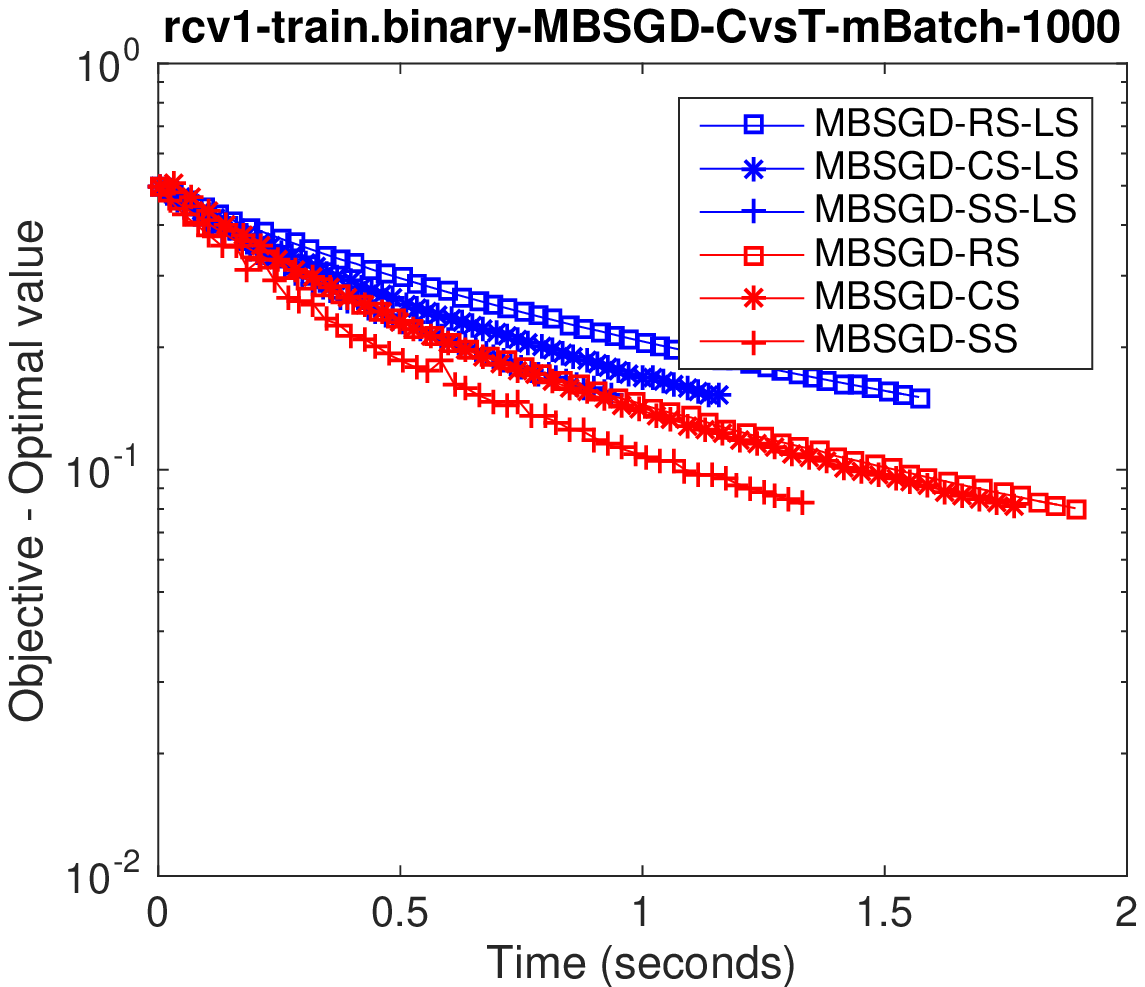}}
	
	\caption{RS, CS and SS are compared using SAG, SAGA, SVRG, SAAG-II and MBSGD, each with two step determination techniques, namely, constant step and backtracking line search, over SUSY and rcv1 datasets with mini-batch of 500 and 1000 data points.}
	\label{fig_1}
\end{figure}

\begin{figure}[htb]
	
	\subfloat{\includegraphics[width=.25\linewidth]{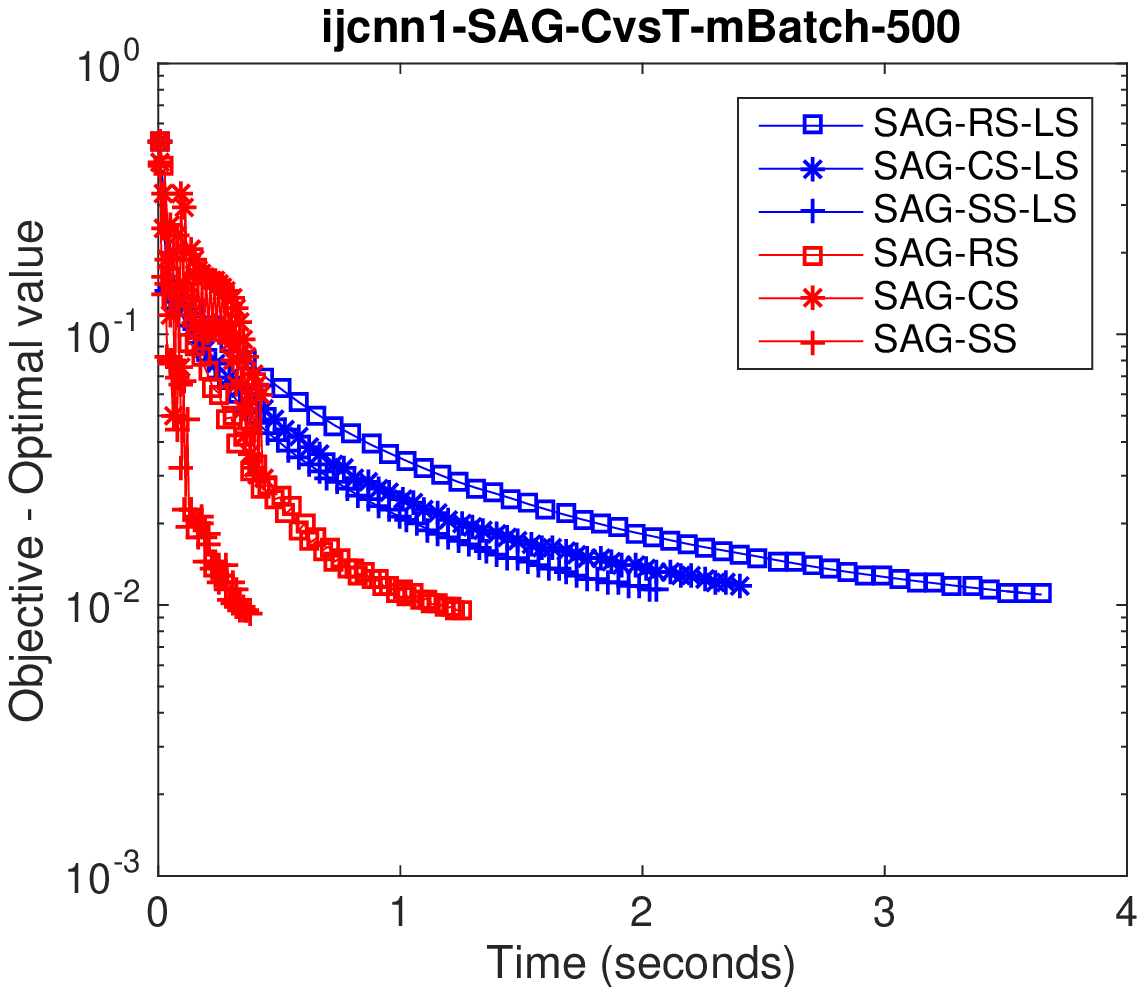}}
	\subfloat{\includegraphics[width=.25\linewidth]{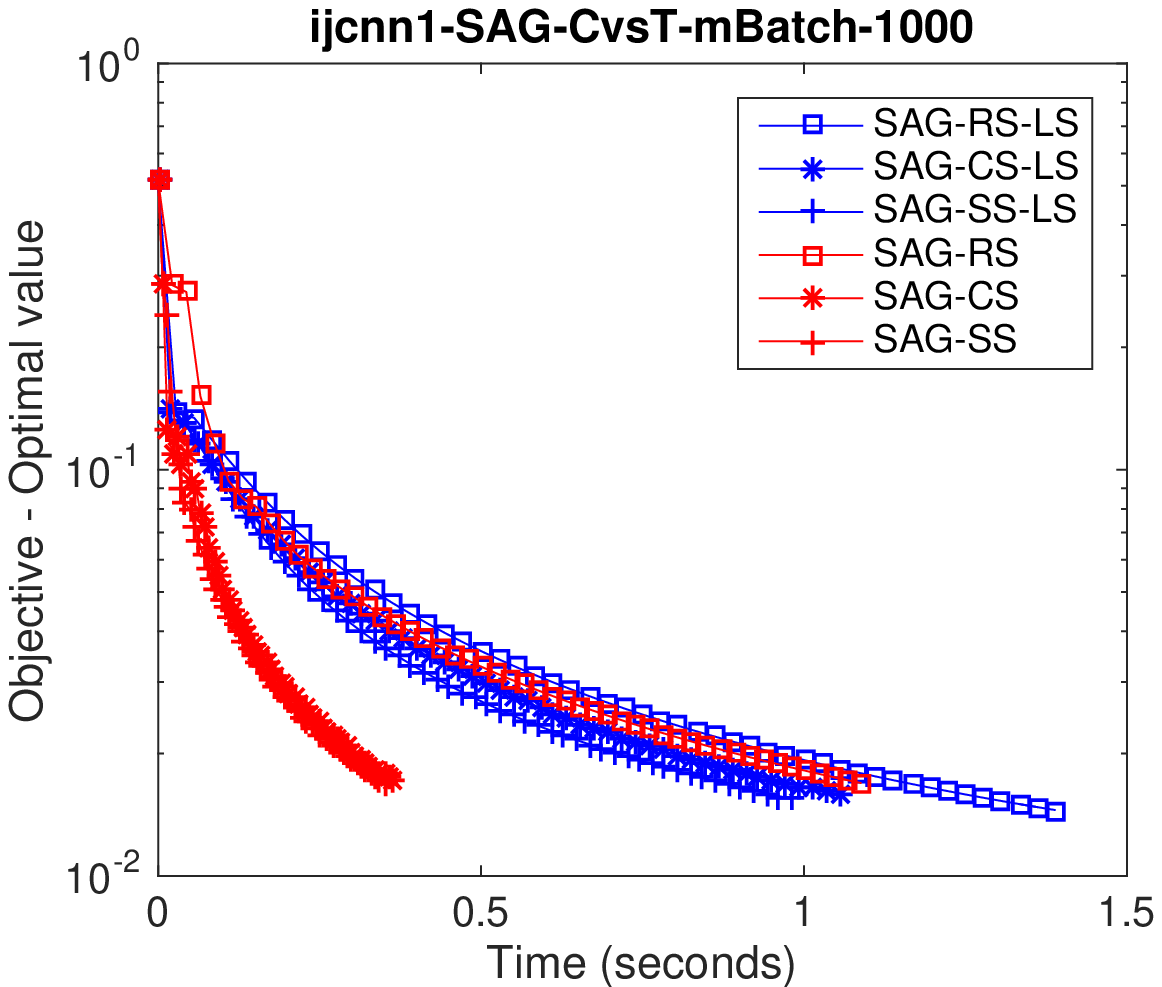}}
	\subfloat{\includegraphics[width=.25\linewidth]{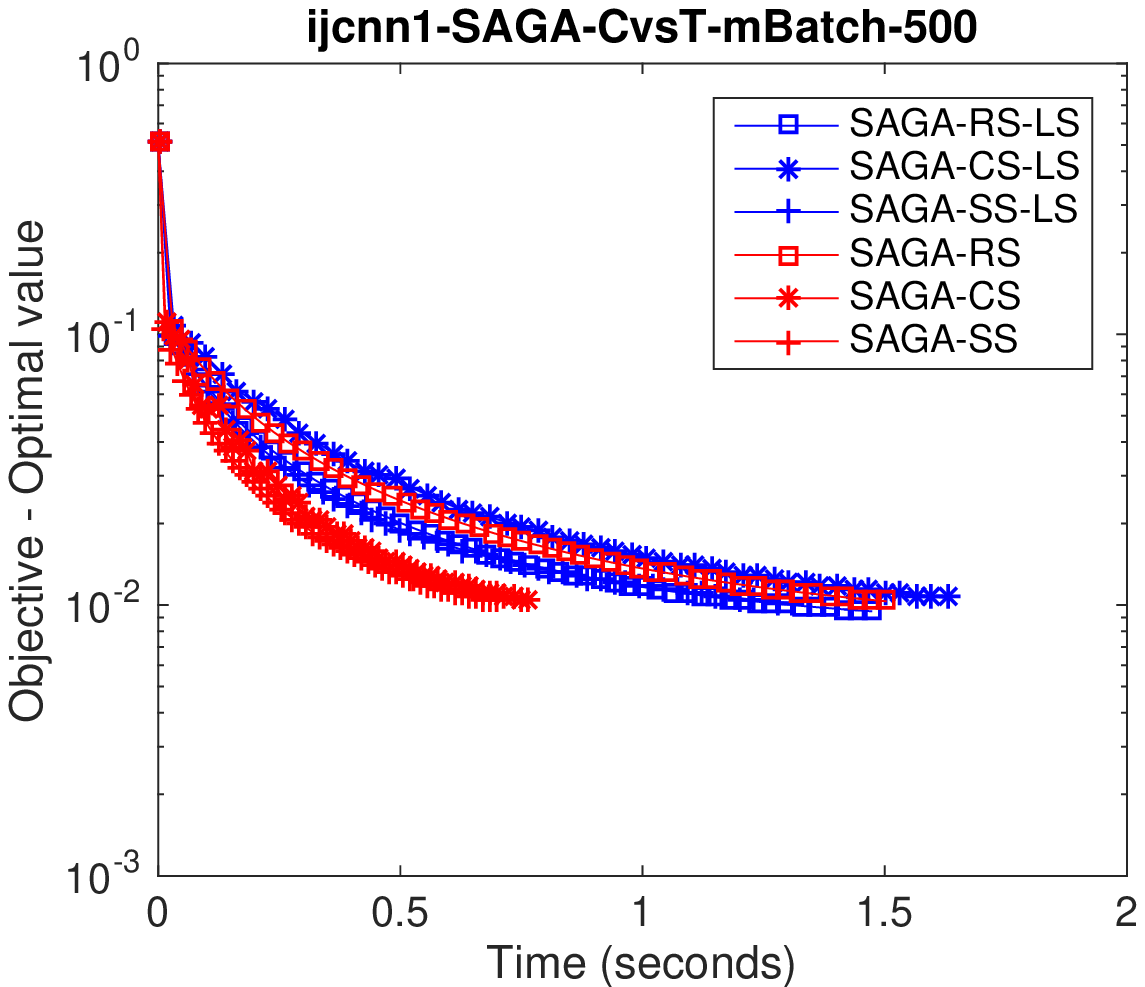}}
	\subfloat{\includegraphics[width=.25\linewidth]{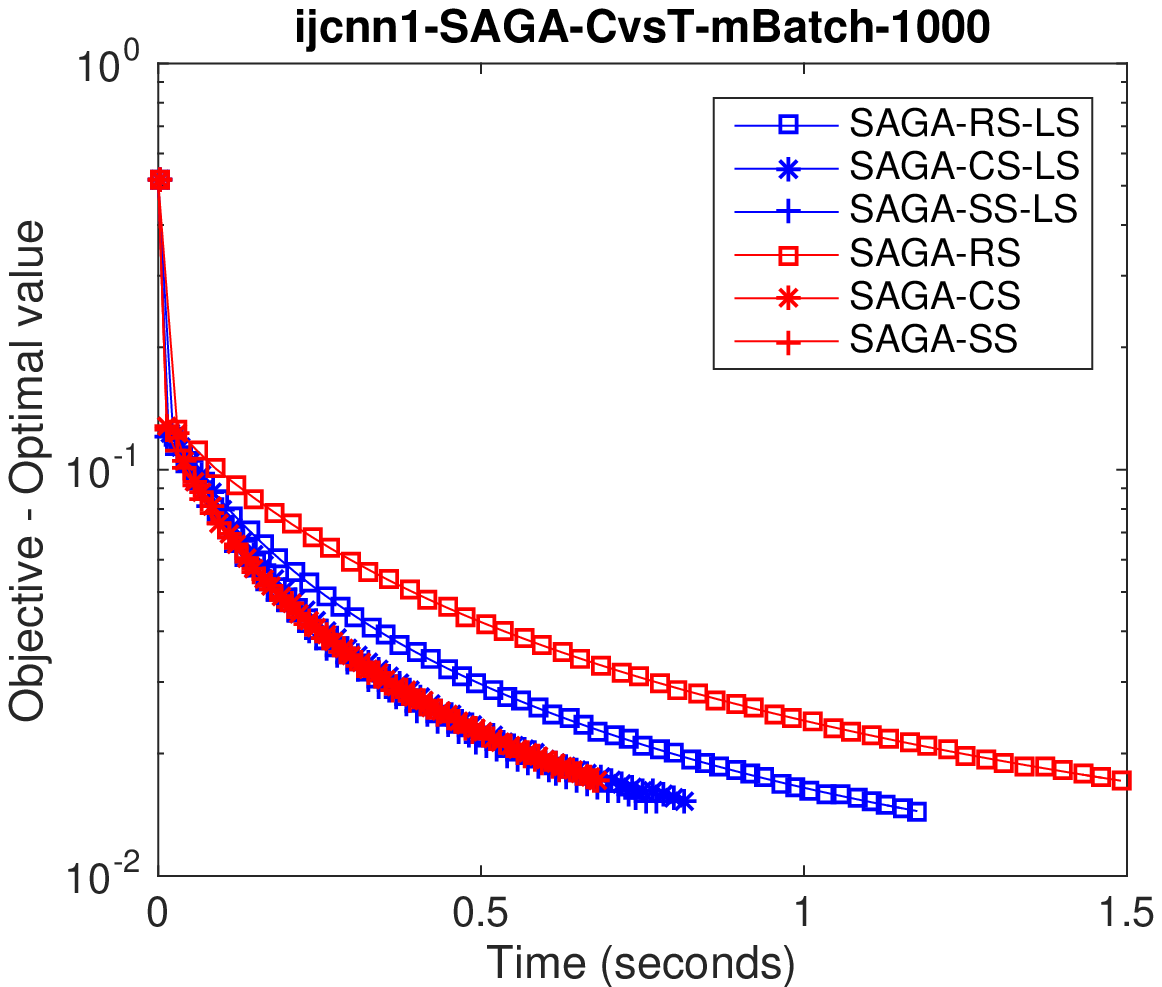}}
	
	\subfloat{\includegraphics[width=.25\linewidth]{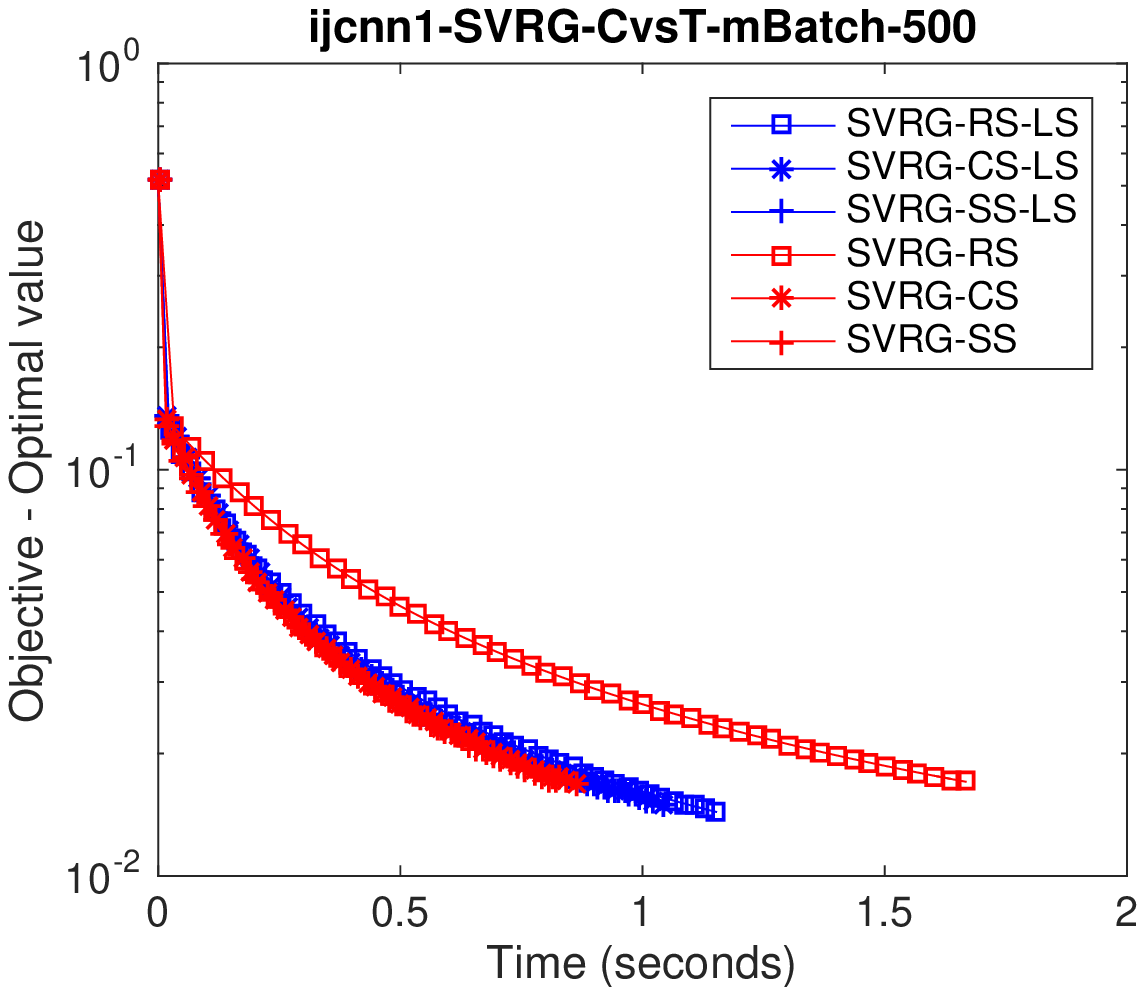}}
	\subfloat{\includegraphics[width=.25\linewidth]{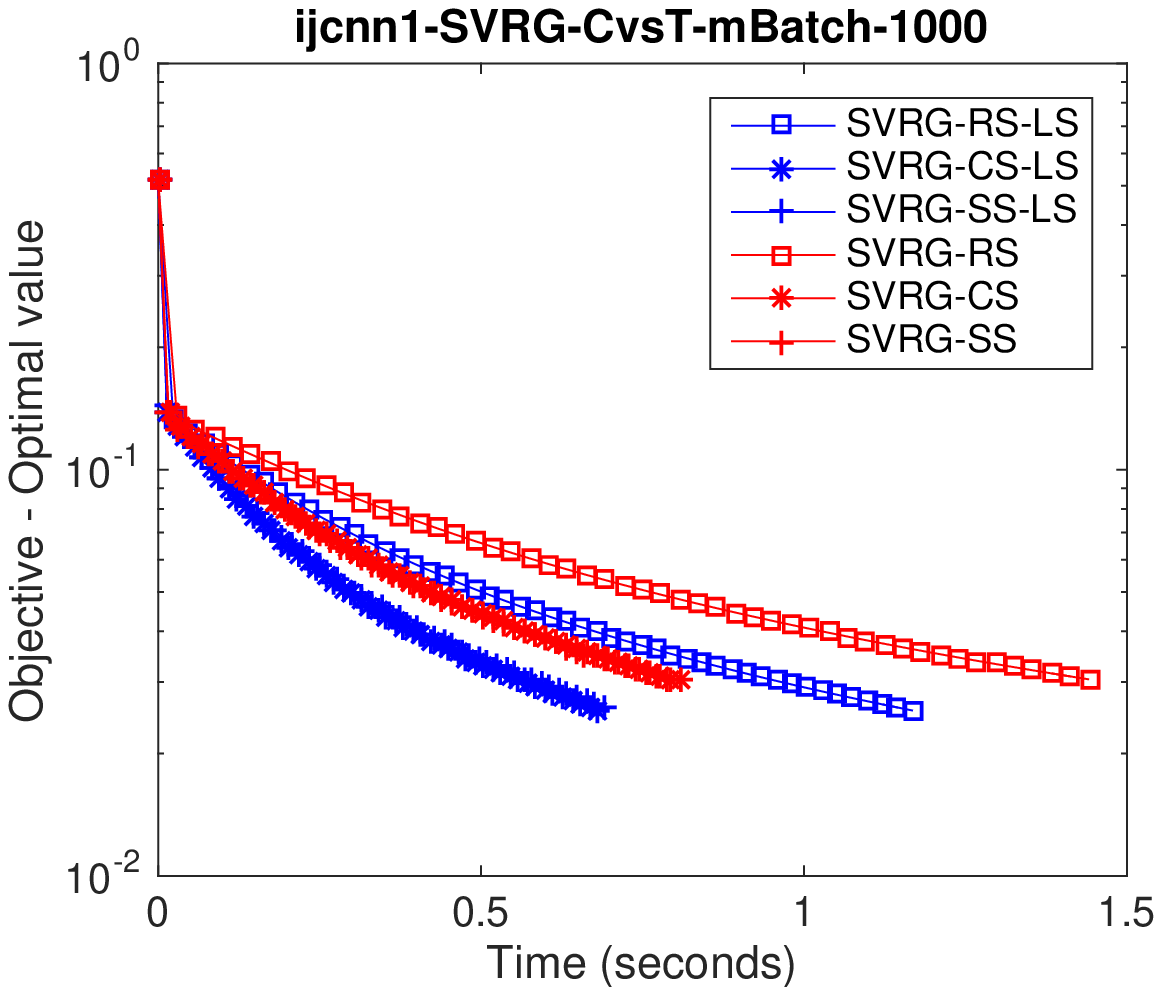}}
	\subfloat{\includegraphics[width=.25\linewidth]{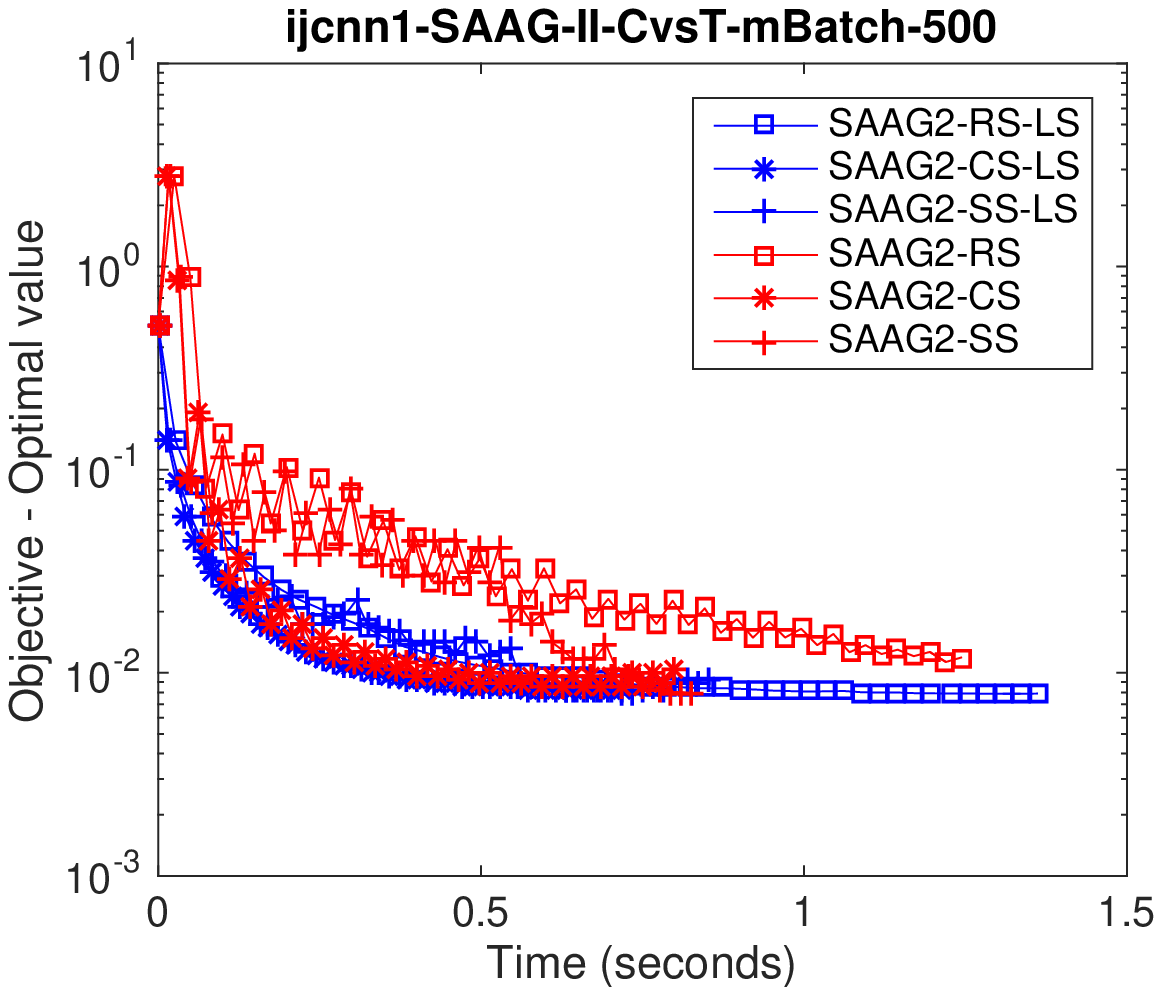}}
	\subfloat{\includegraphics[width=.25\linewidth]{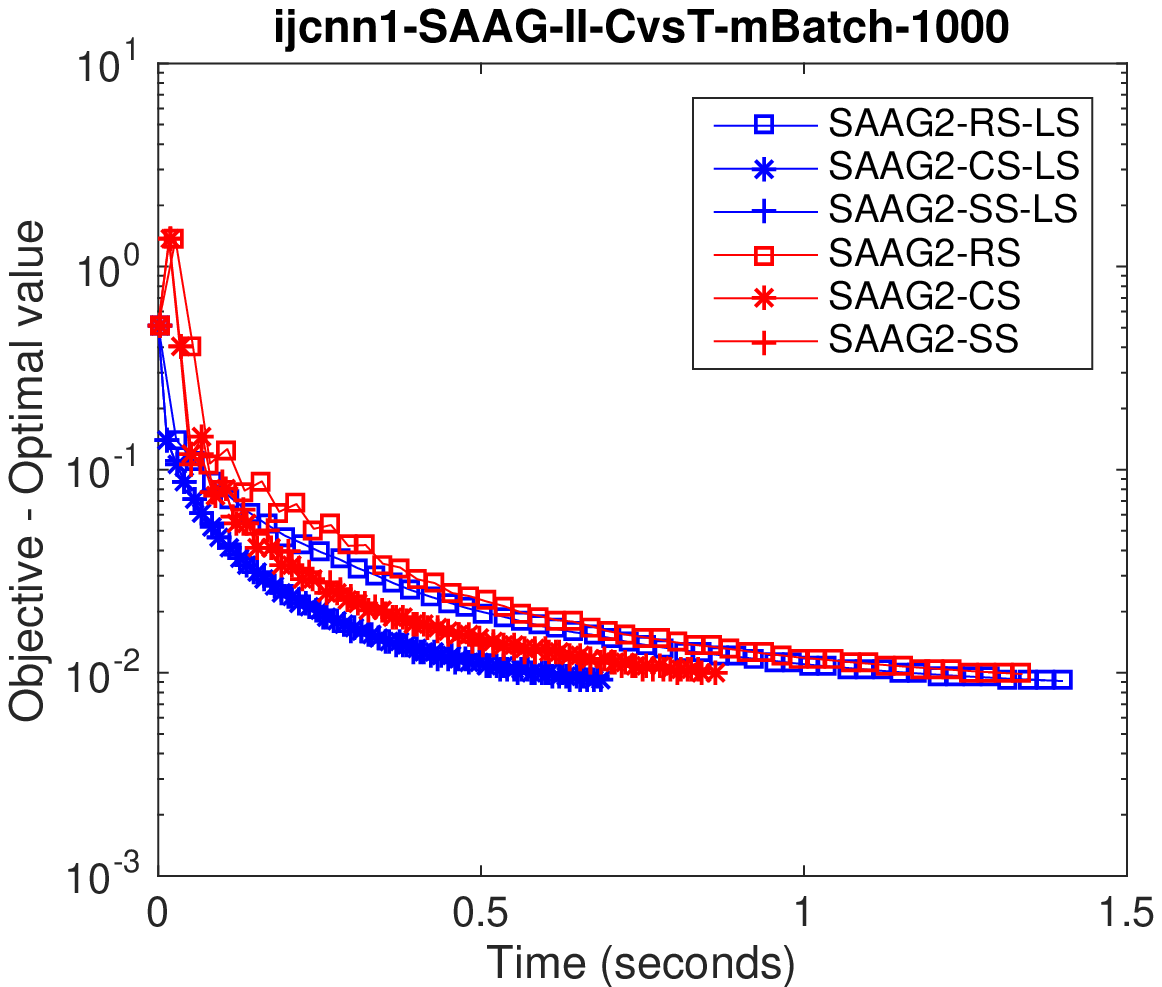}}
	
	\subfloat{\includegraphics[width=.25\linewidth]{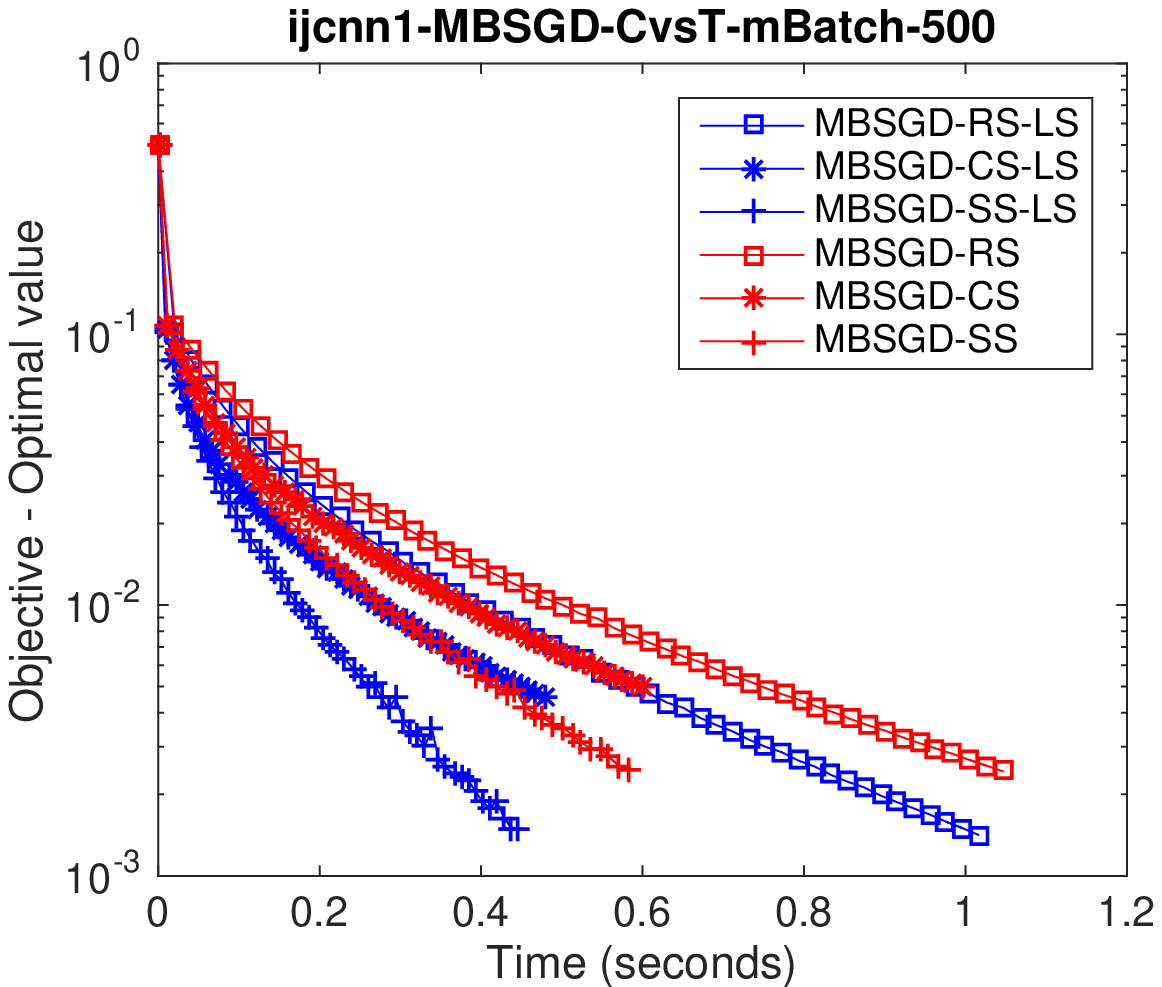}}
	\subfloat{\includegraphics[width=.25\linewidth]{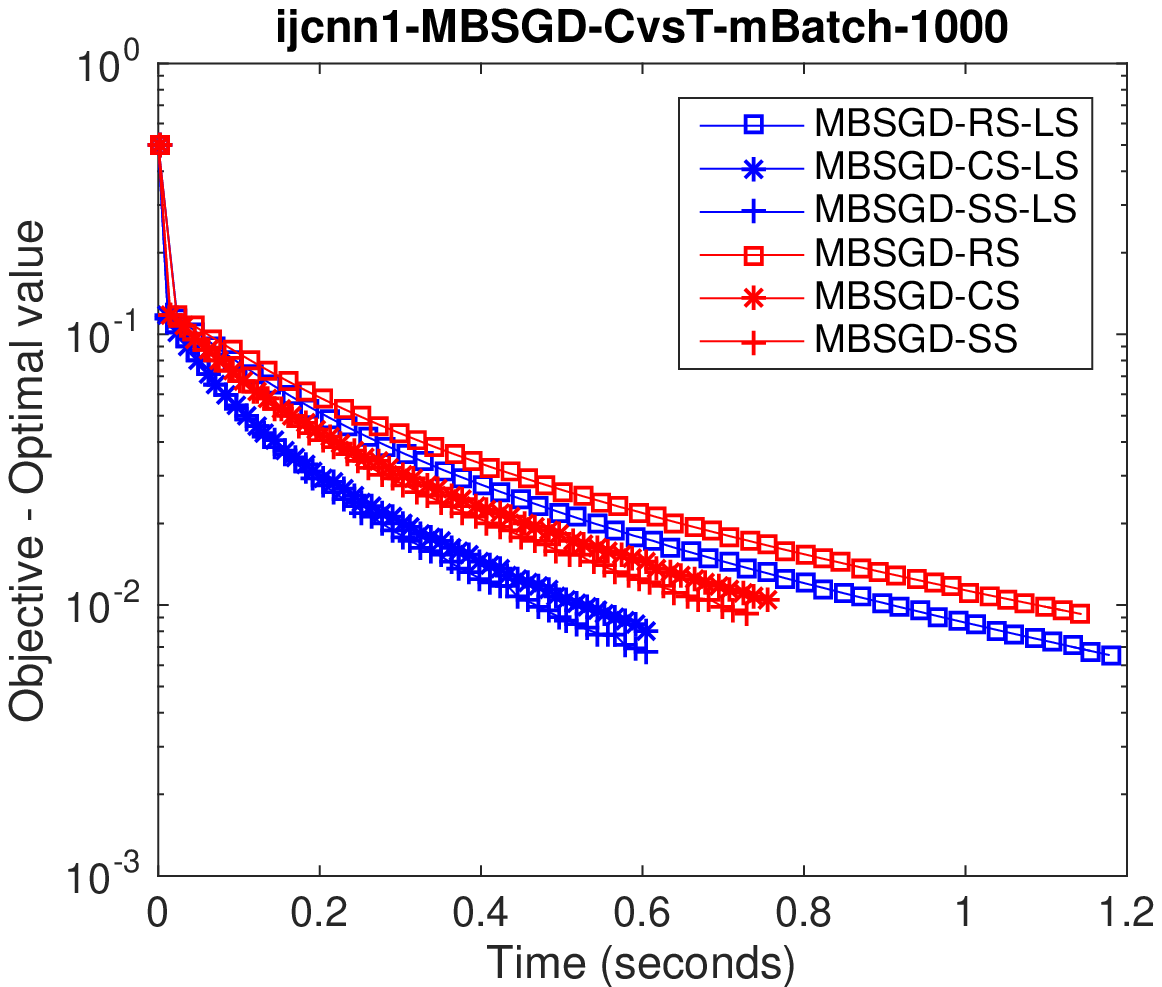}}
	\subfloat{\includegraphics[width=.25\linewidth]{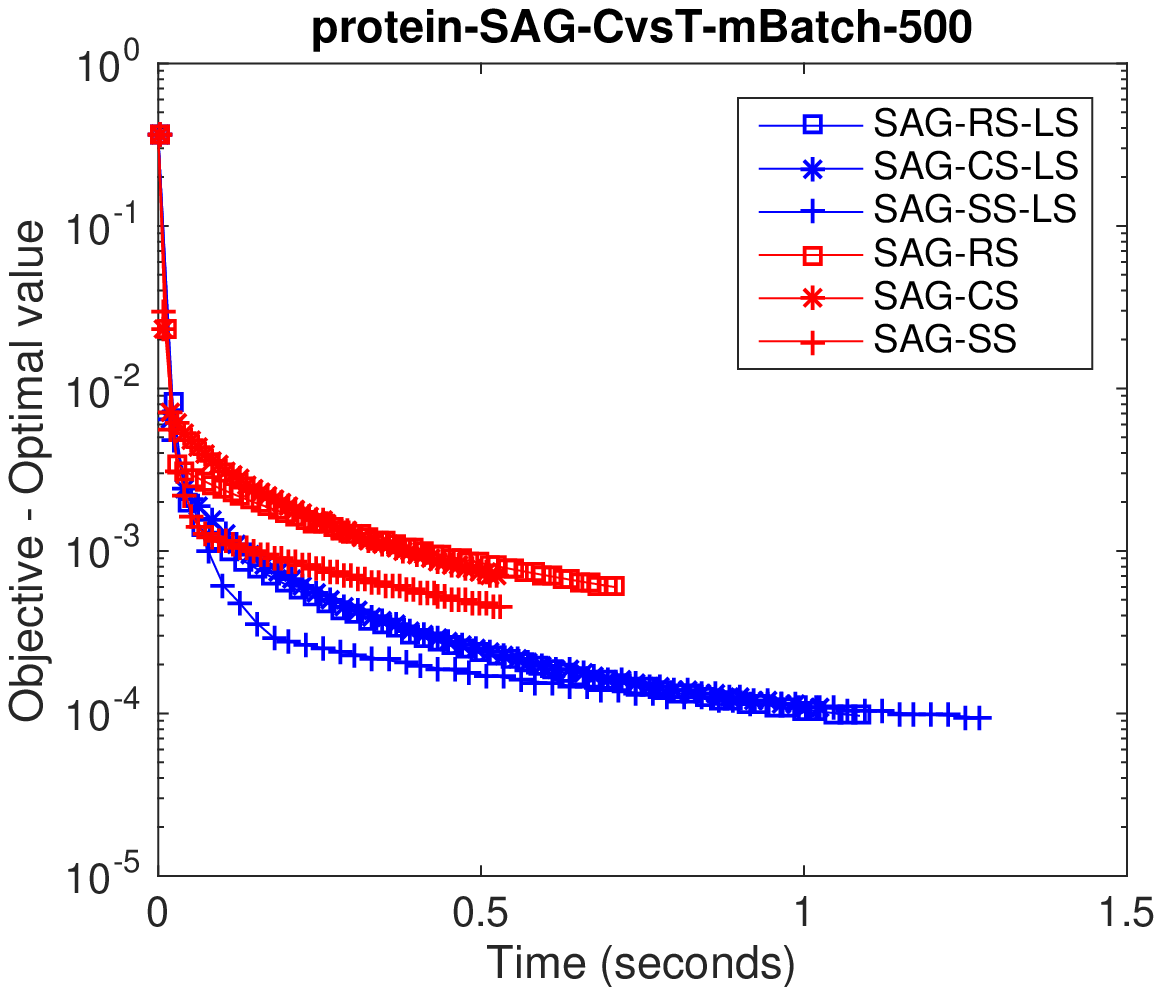}}	
	\subfloat{\includegraphics[width=.25\linewidth]{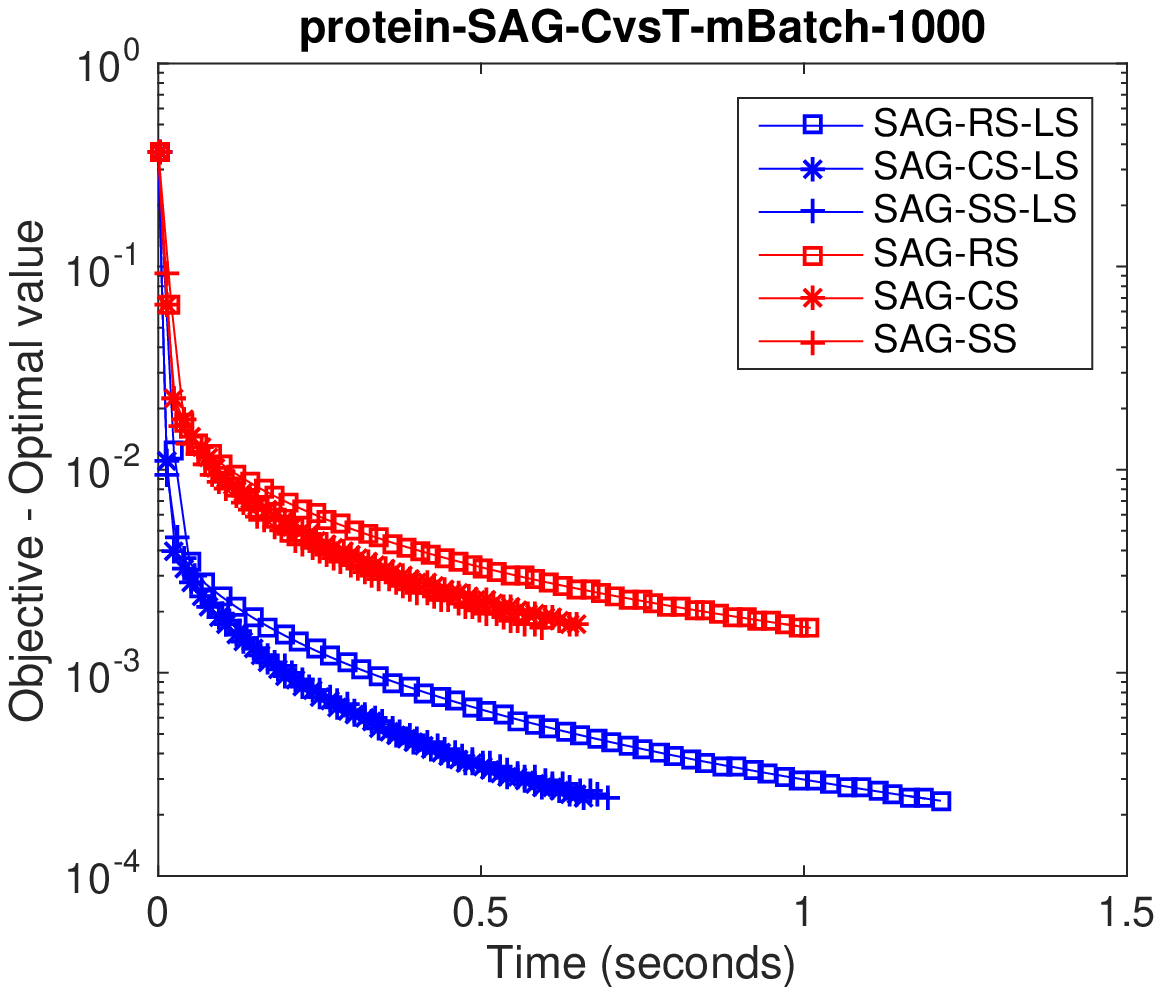}}
	
	\subfloat{\includegraphics[width=.25\linewidth]{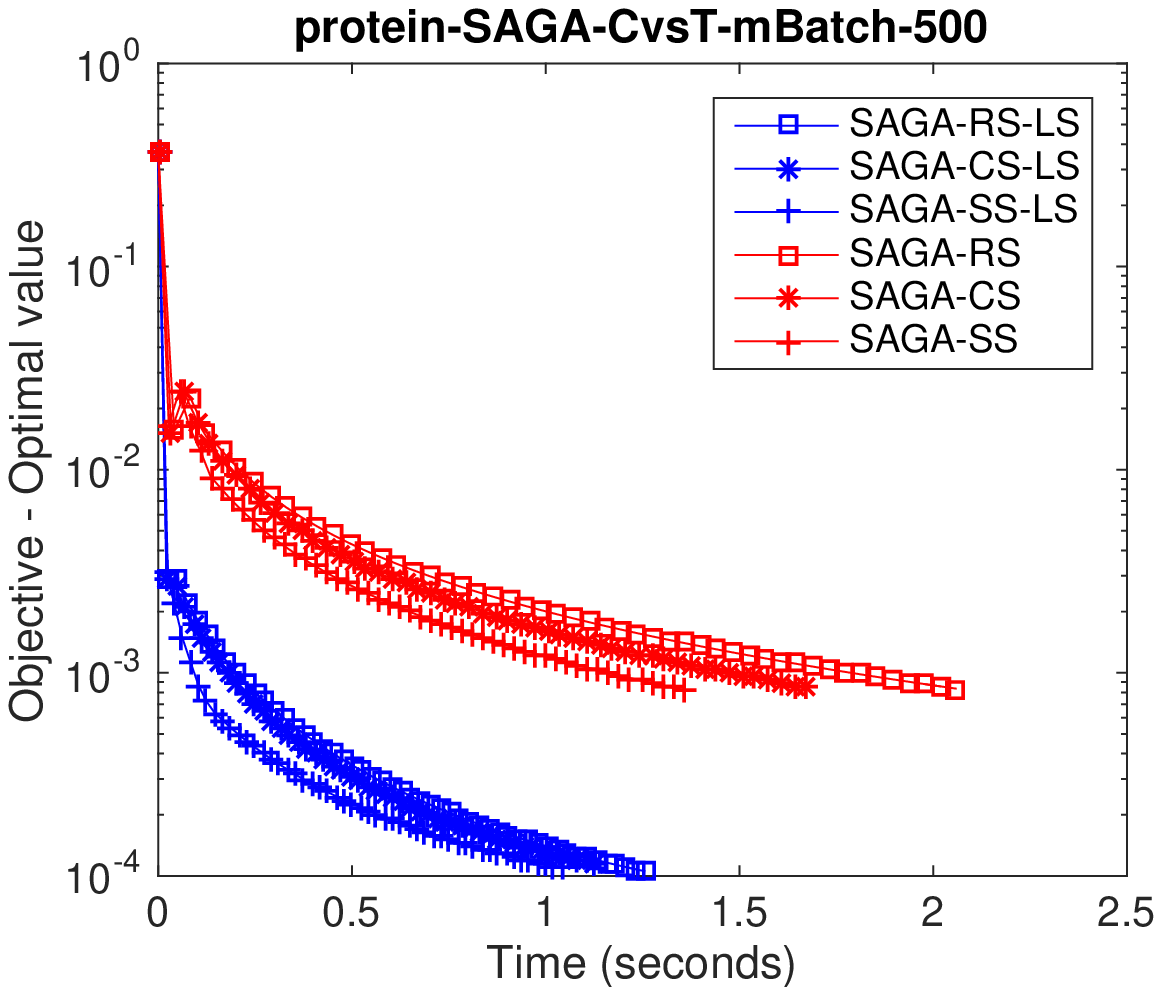}}
	\subfloat{\includegraphics[width=.25\linewidth]{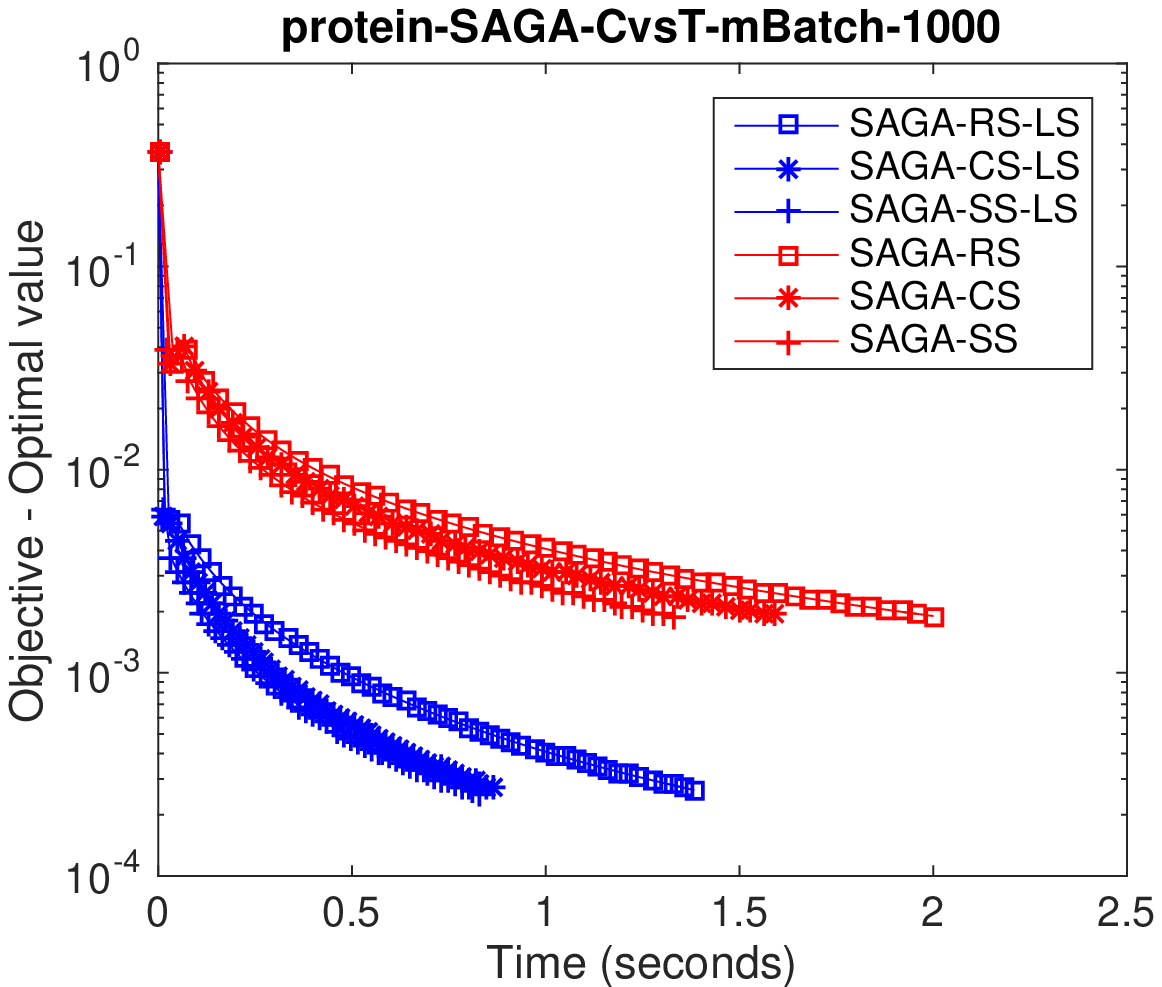}}
	\subfloat{\includegraphics[width=.25\linewidth]{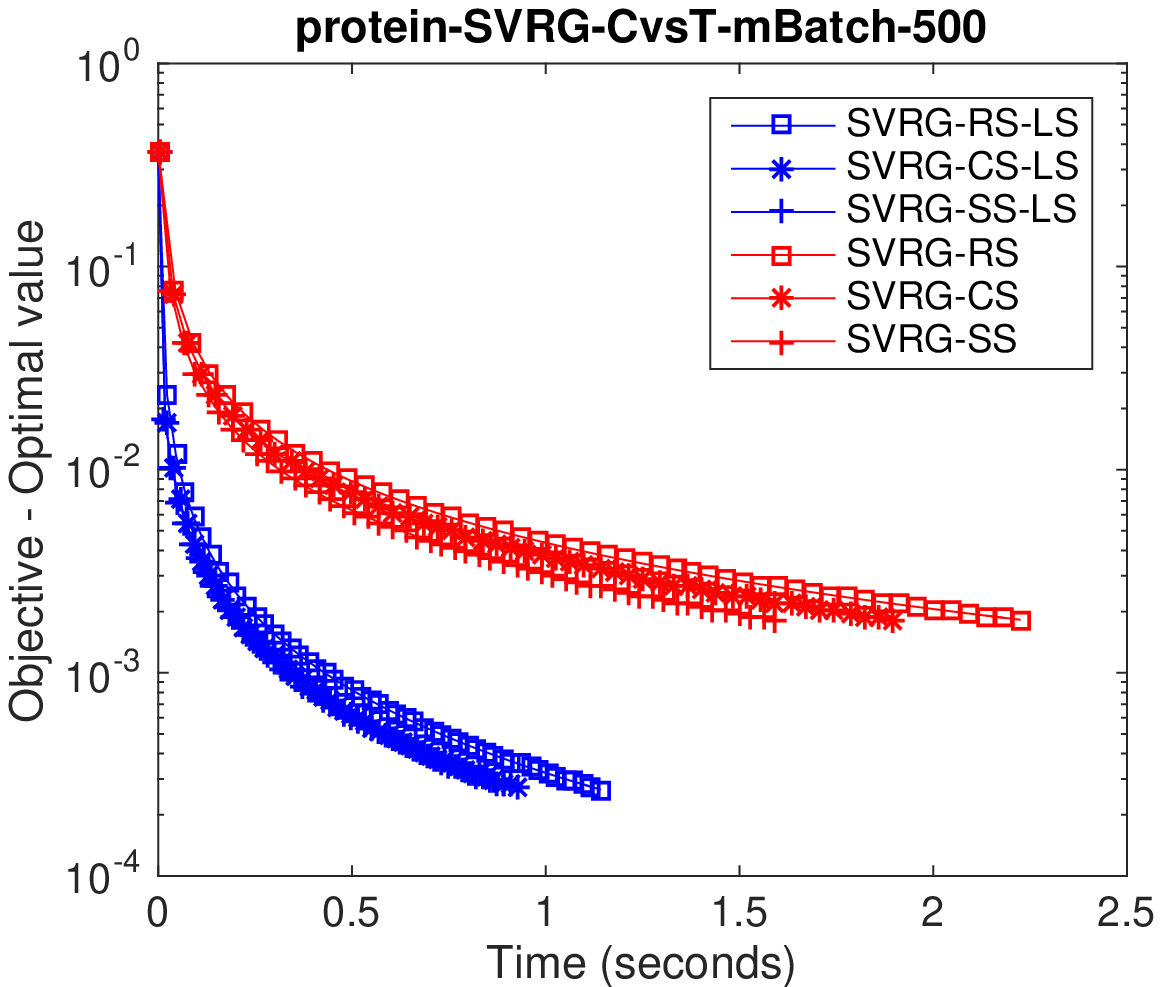}}
	\subfloat{\includegraphics[width=.25\linewidth]{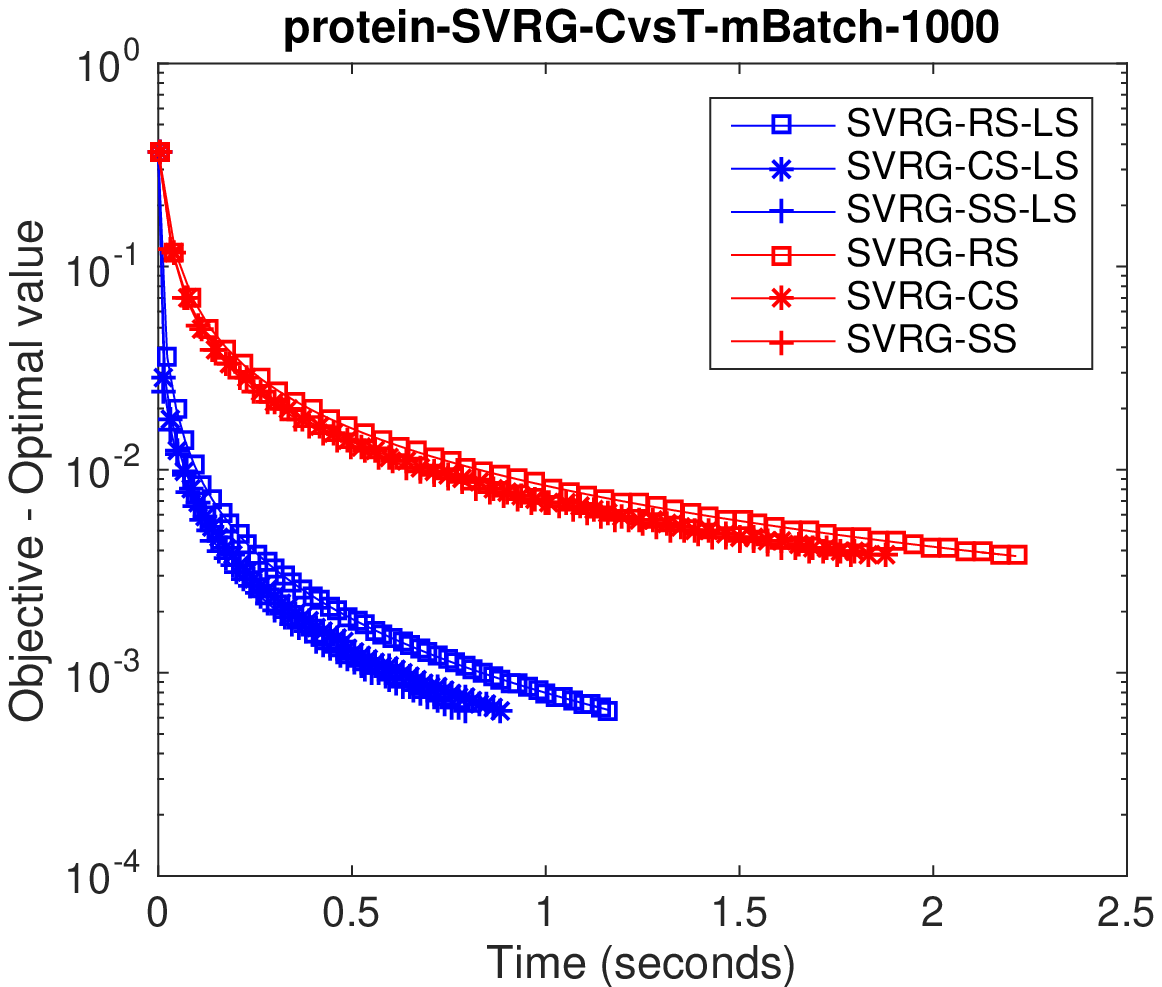}}
	
	\subfloat{\includegraphics[width=.25\linewidth]{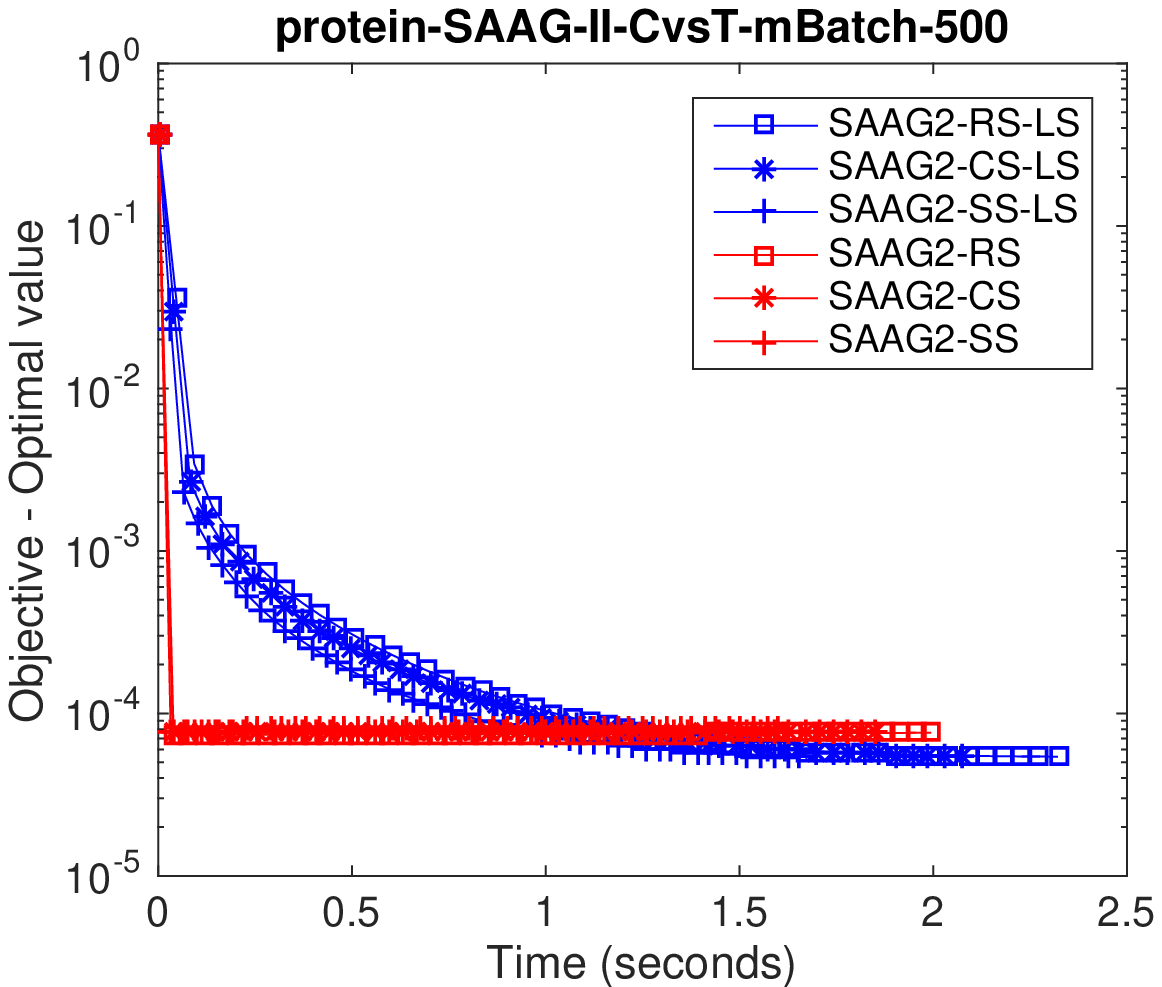}}
	\subfloat{\includegraphics[width=.25\linewidth]{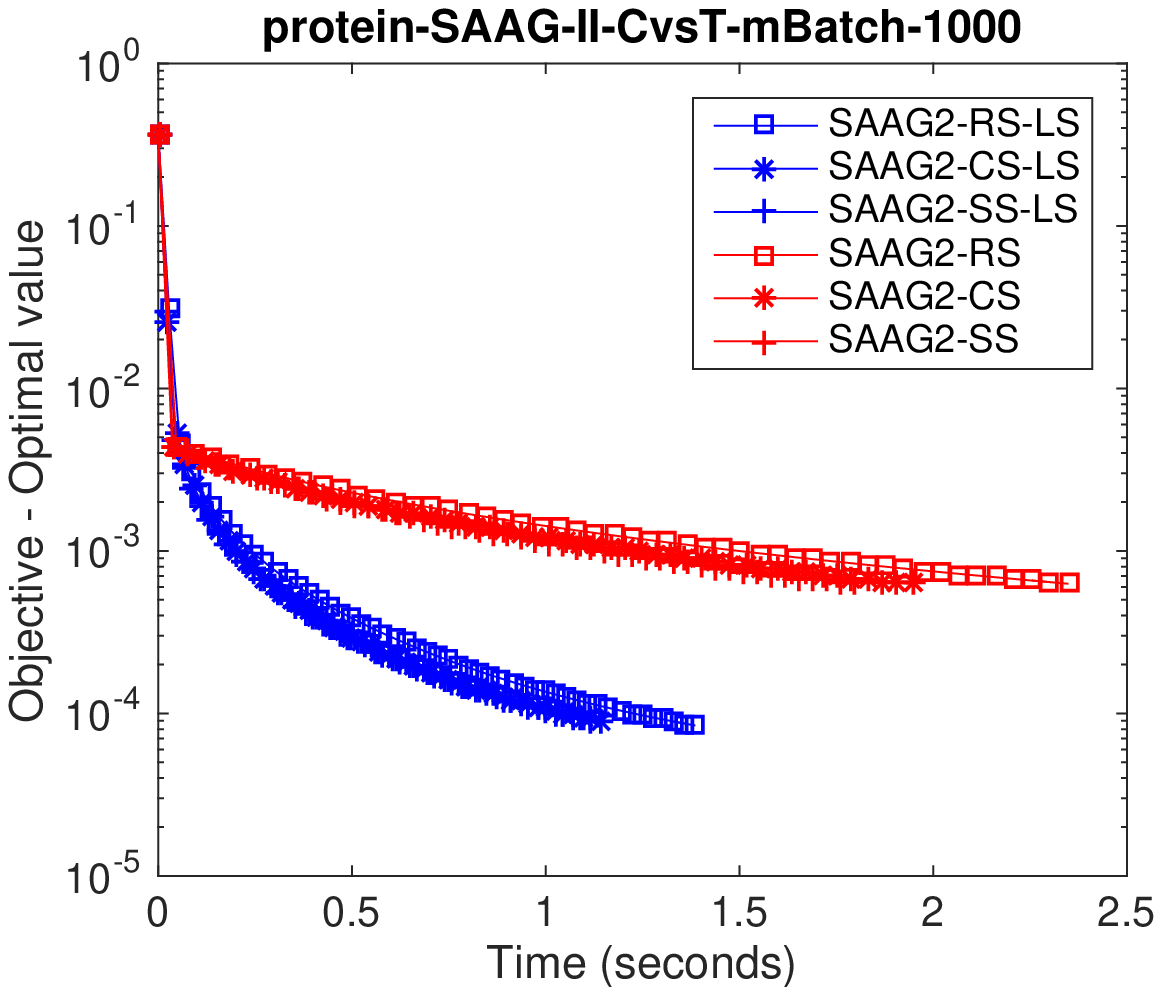}}
	\subfloat{\includegraphics[width=.25\linewidth]{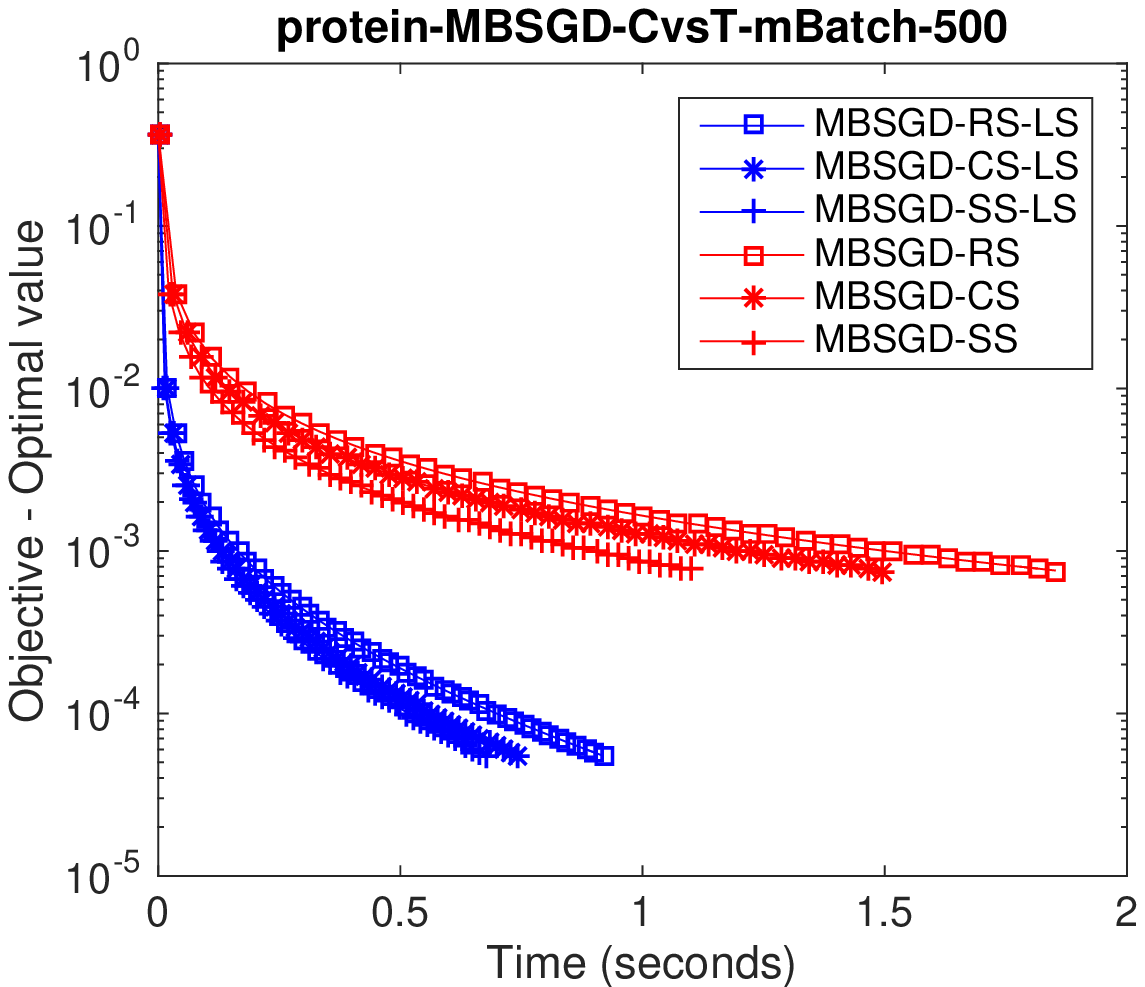}}
	\subfloat{\includegraphics[width=.25\linewidth]{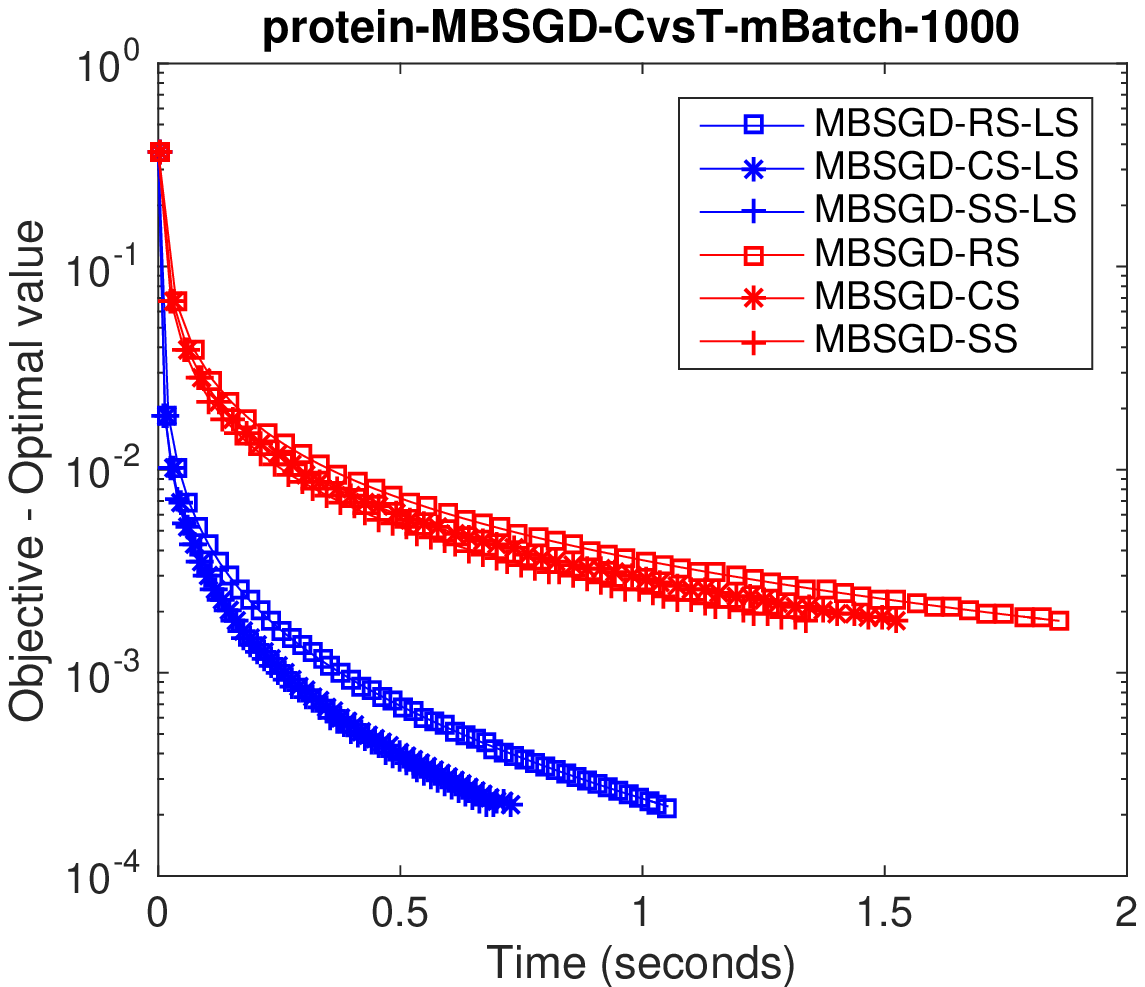}}
	
	\caption{RS, CS and SS are compared using SAG, SAGA, SVRG, SAAG-II and MBSGD, each with two step determination techniques, namely, constant step and backtracking line search, over ijcnn1 and protein datasets with mini-batch of 500 and 1000 data points.}
	\label{fig_2}
\end{figure}

\begin{figure}[htb]
	\subfloat{\includegraphics[width=.25\linewidth]{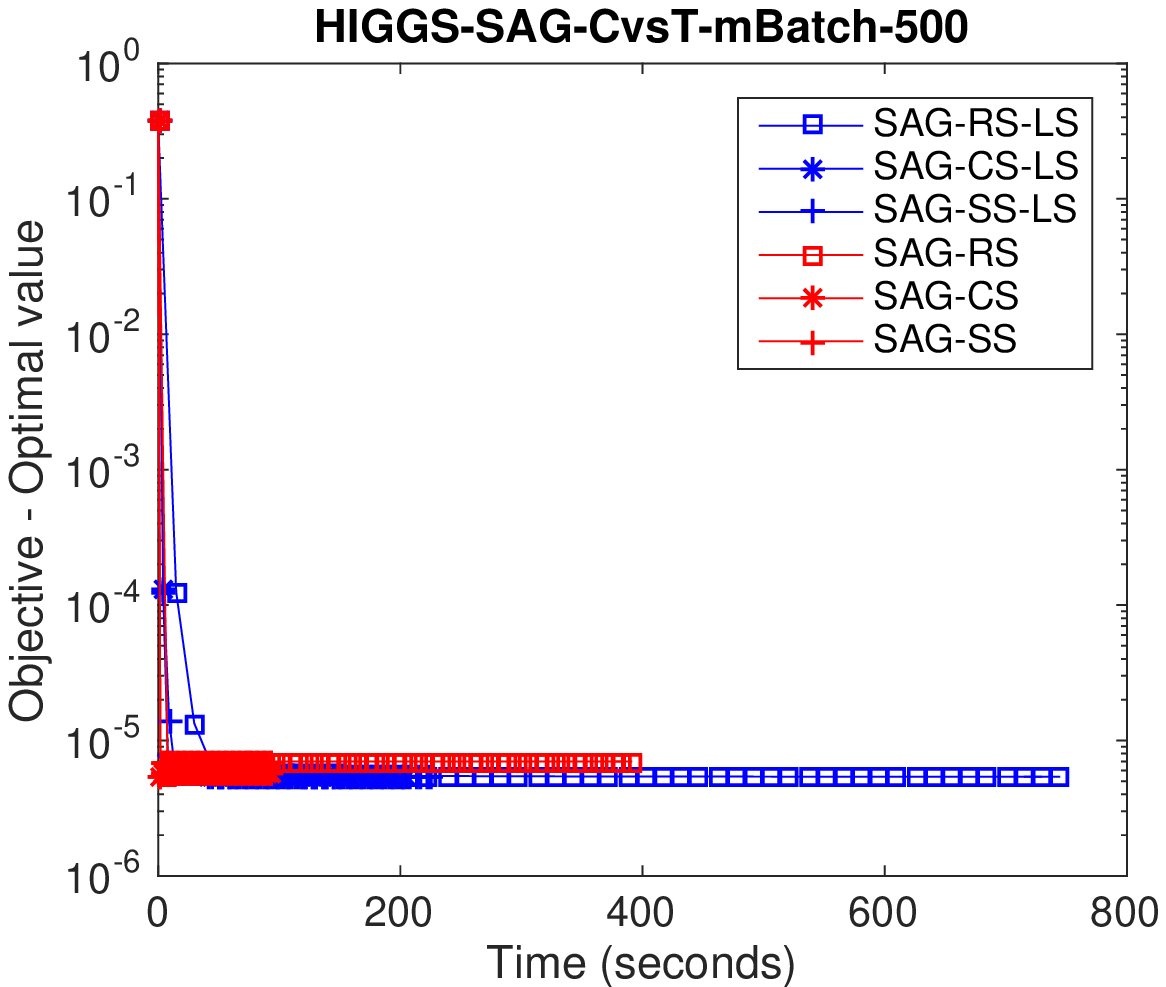}}
	\subfloat{\includegraphics[width=.25\linewidth]{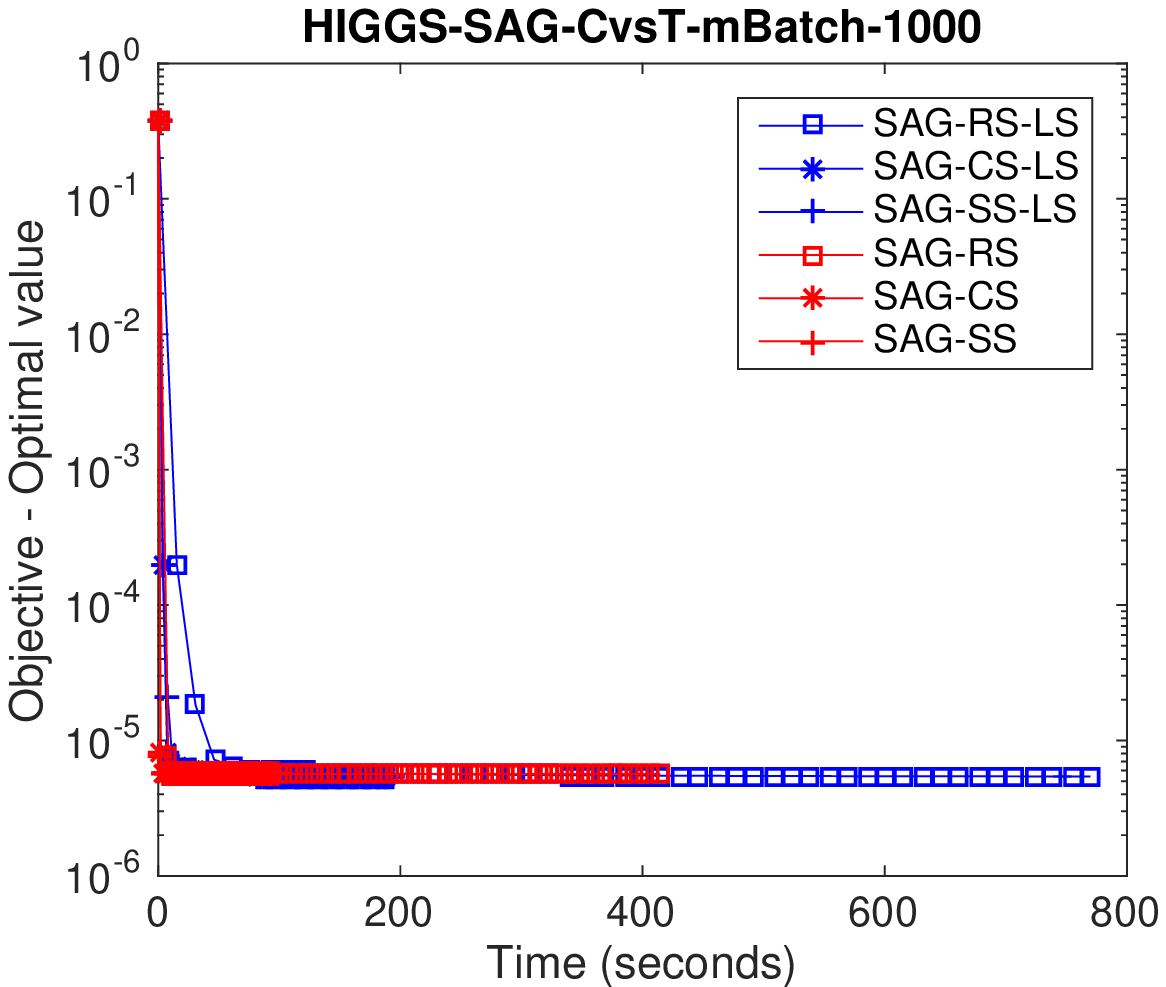}}
	\subfloat{\includegraphics[width=.25\linewidth]{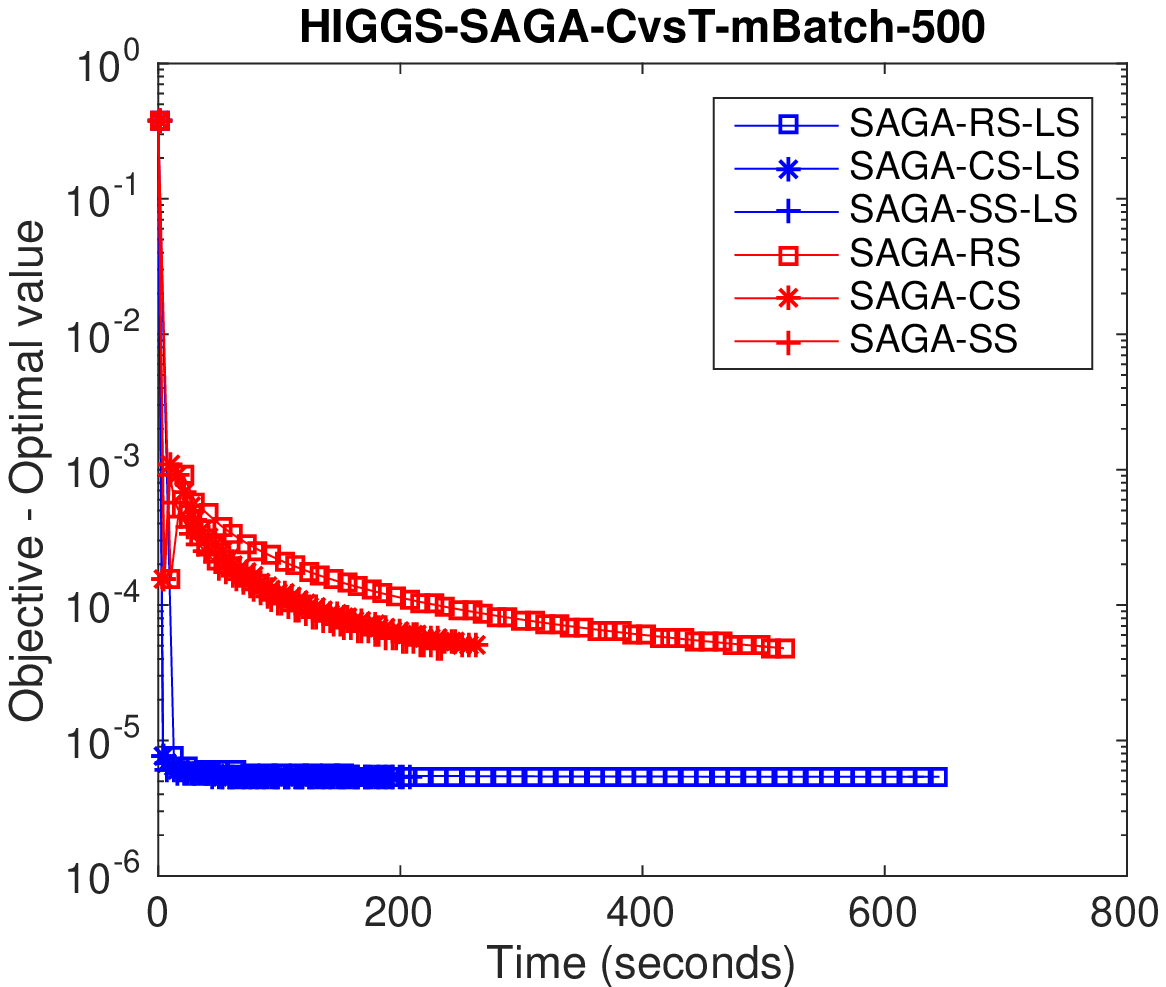}}
	\subfloat{\includegraphics[width=.25\linewidth]{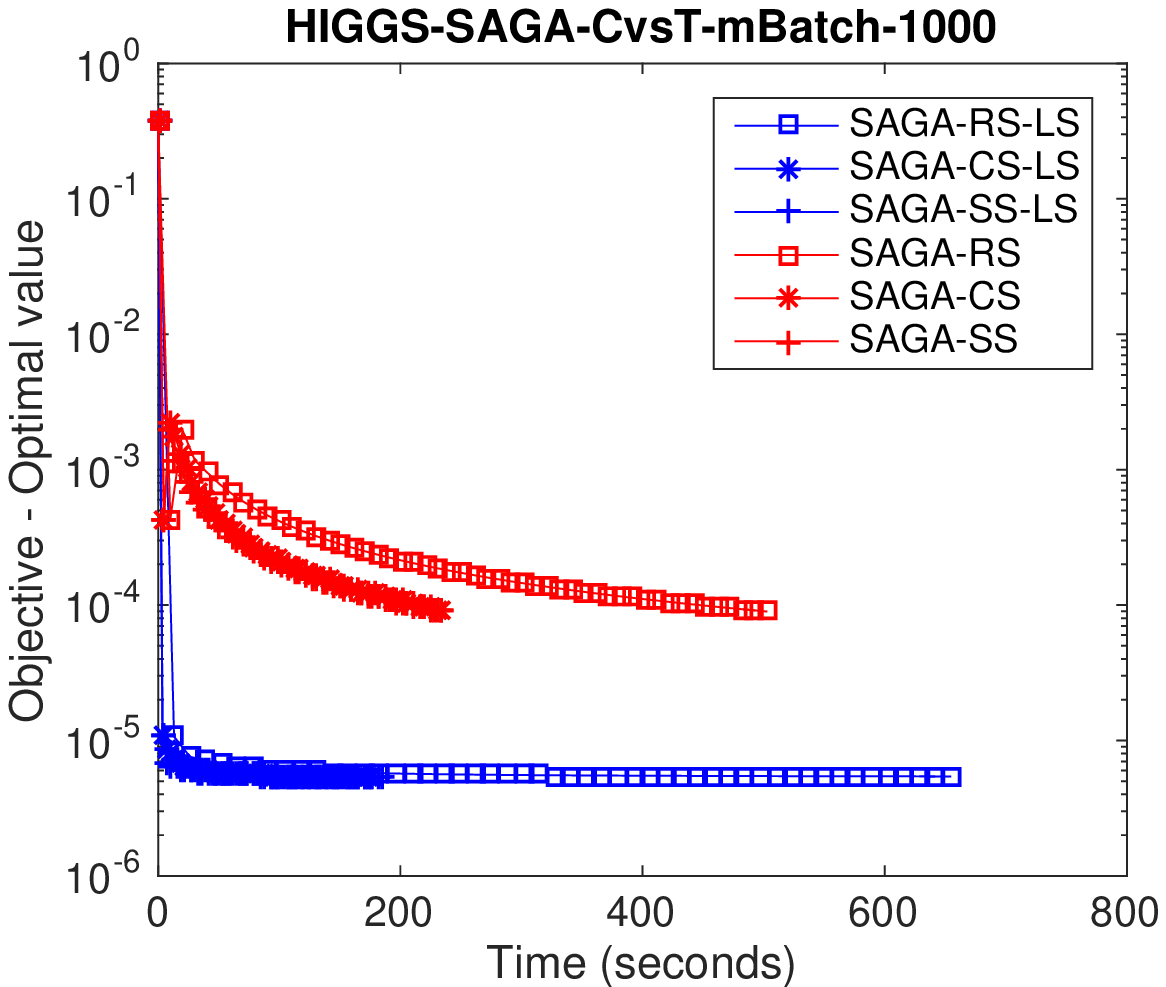}}
	
	\subfloat{\includegraphics[width=.25\linewidth]{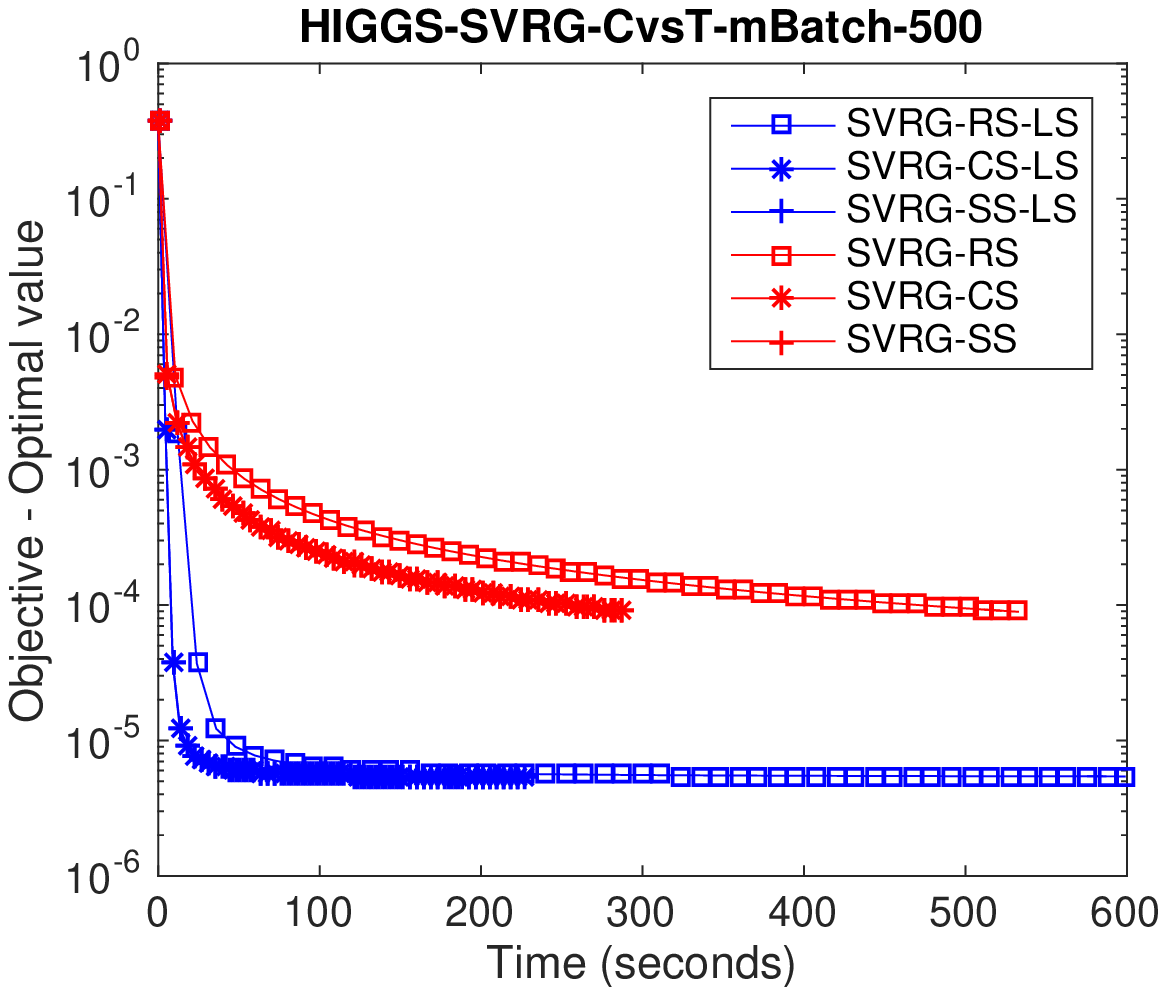}}
	\subfloat{\includegraphics[width=.25\linewidth]{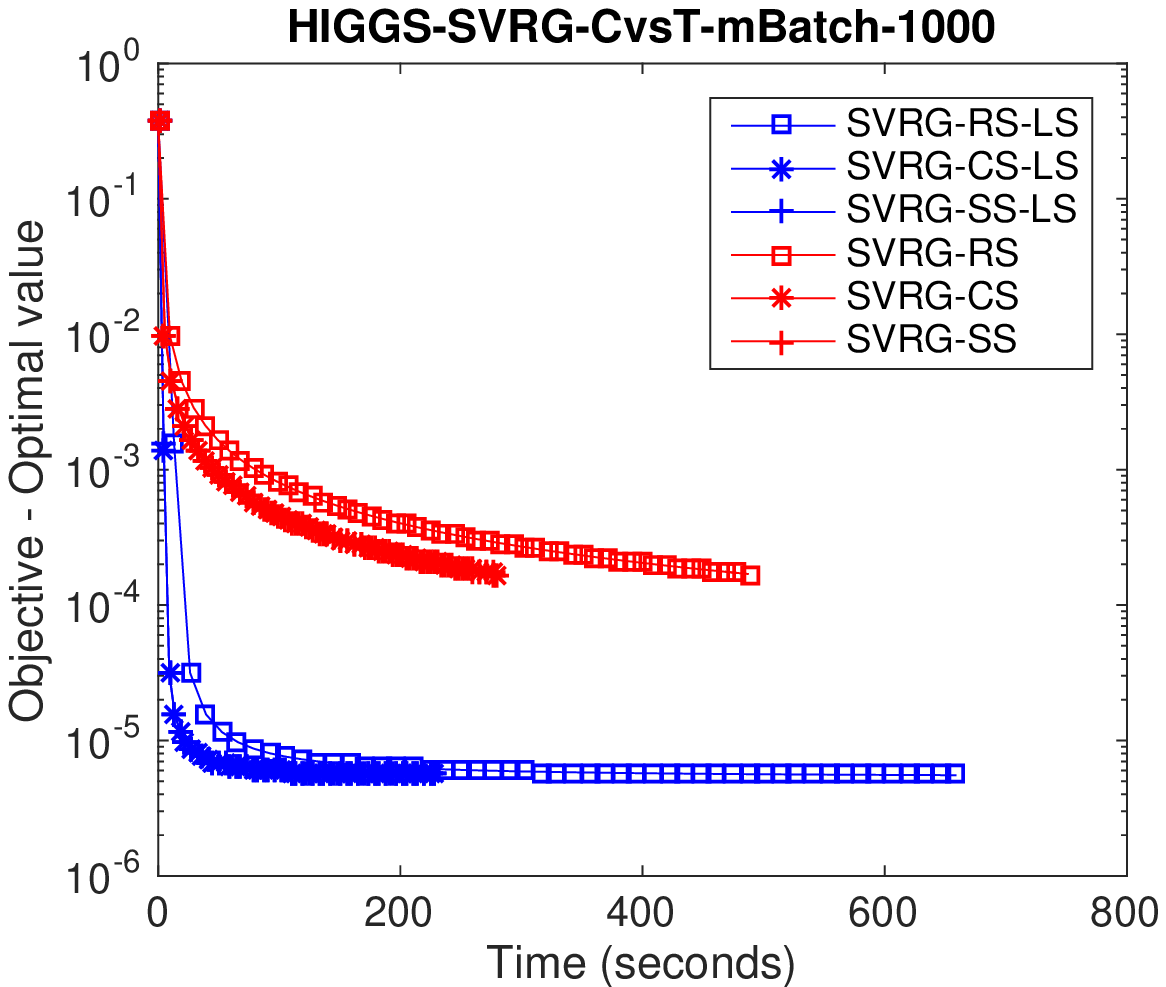}}	
	\subfloat{\includegraphics[width=.25\linewidth]{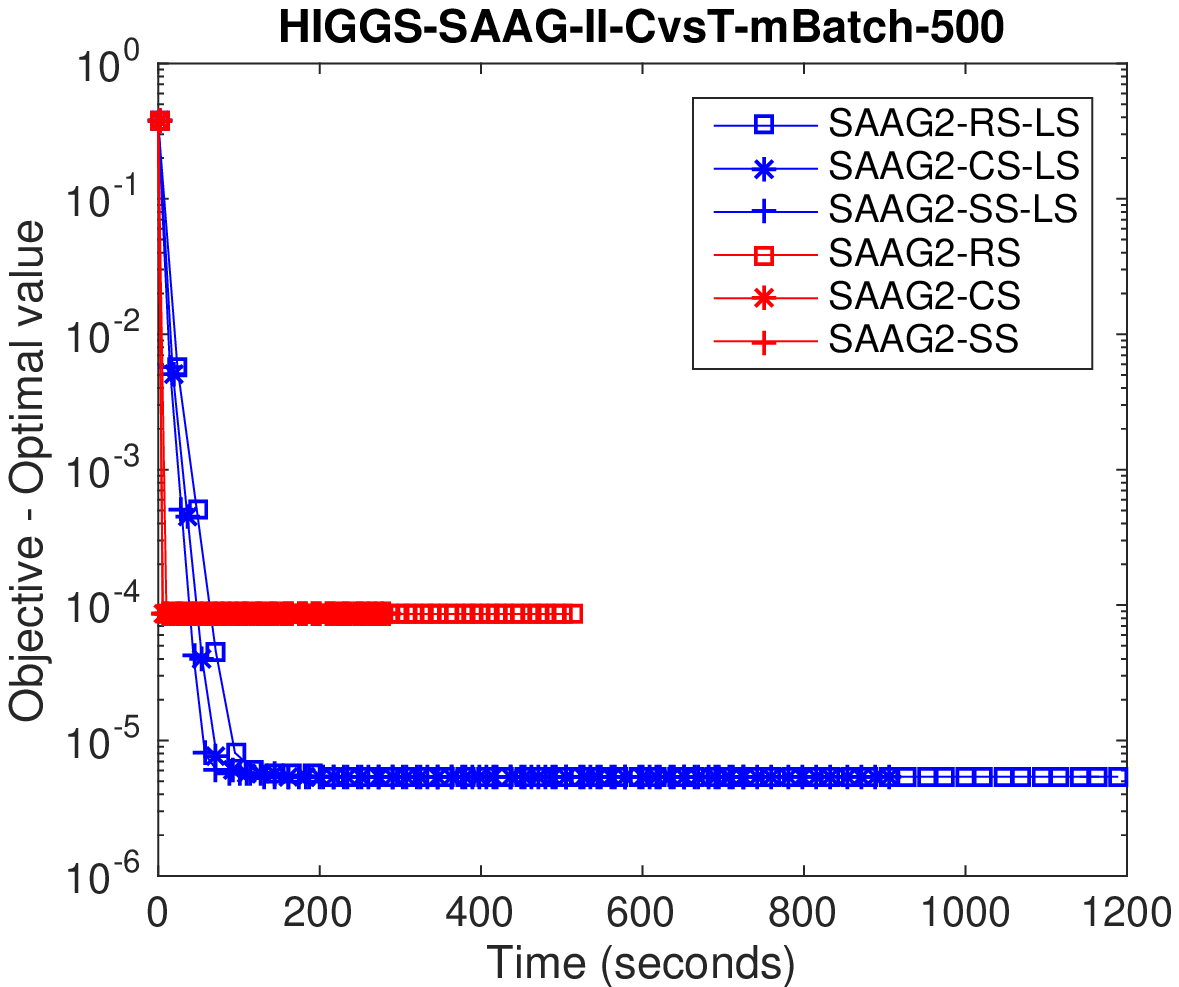}}
	\subfloat{\includegraphics[width=.25\linewidth]{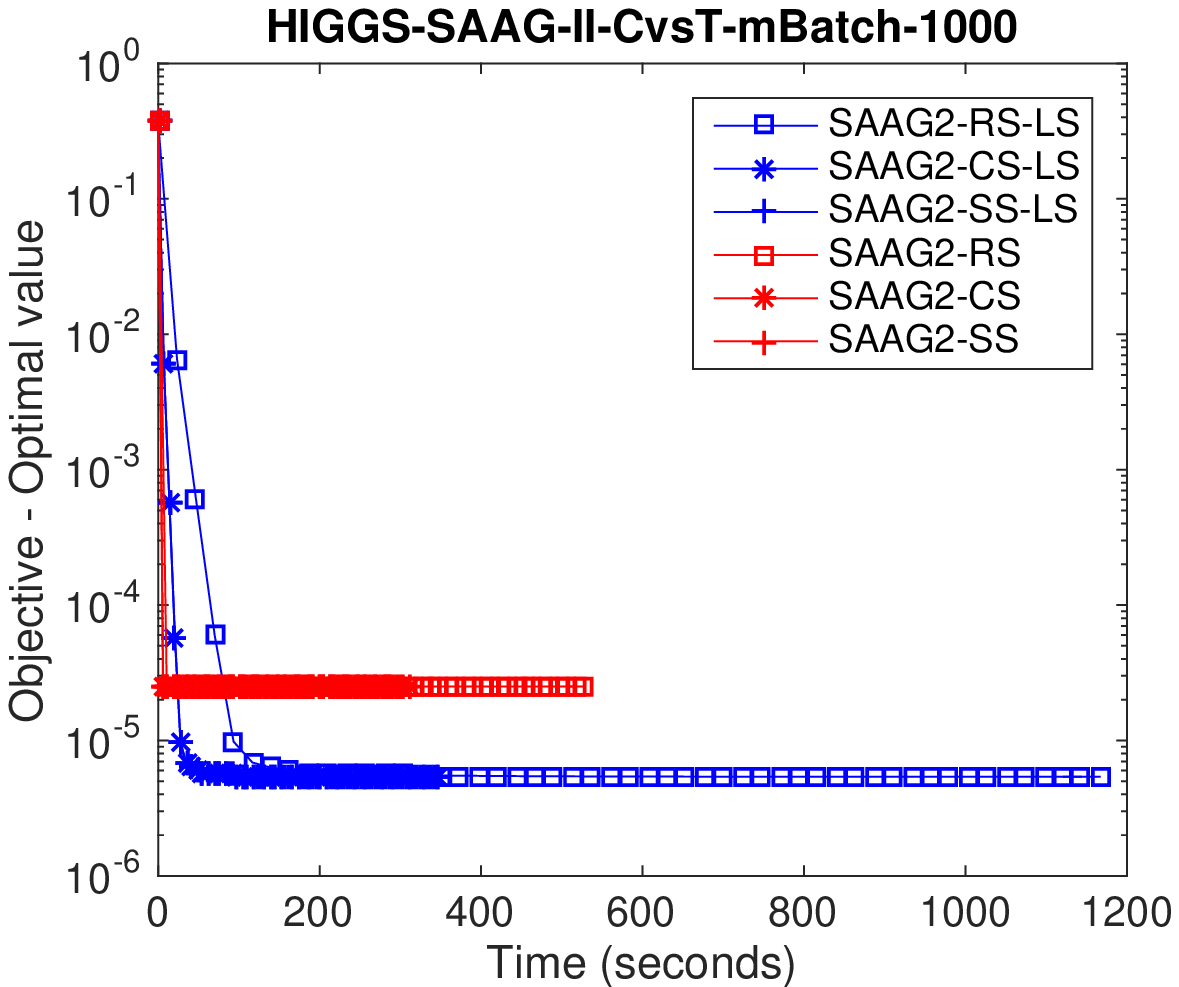}}

	\subfloat{\includegraphics[width=.25\linewidth]{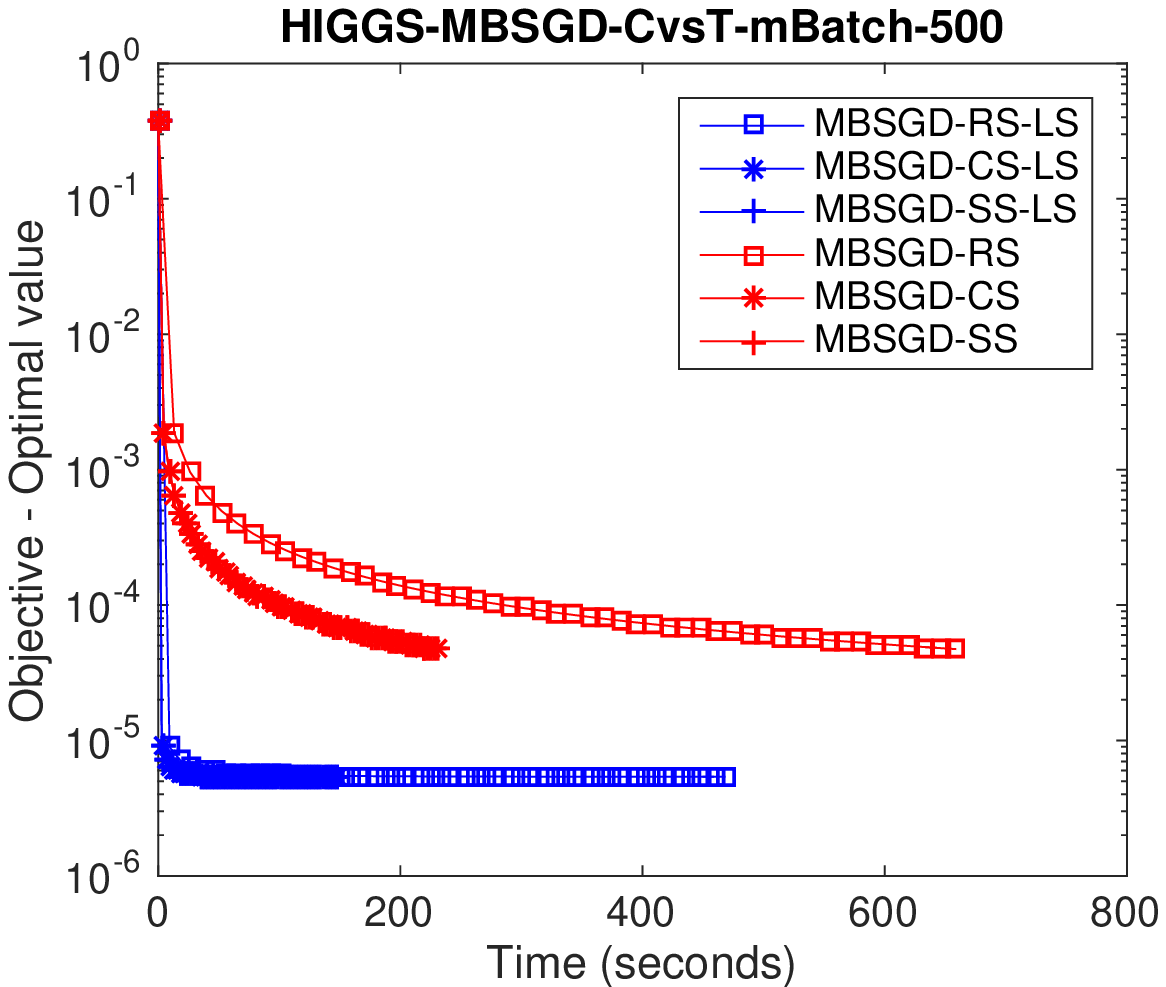}}
	\subfloat{\includegraphics[width=.25\linewidth]{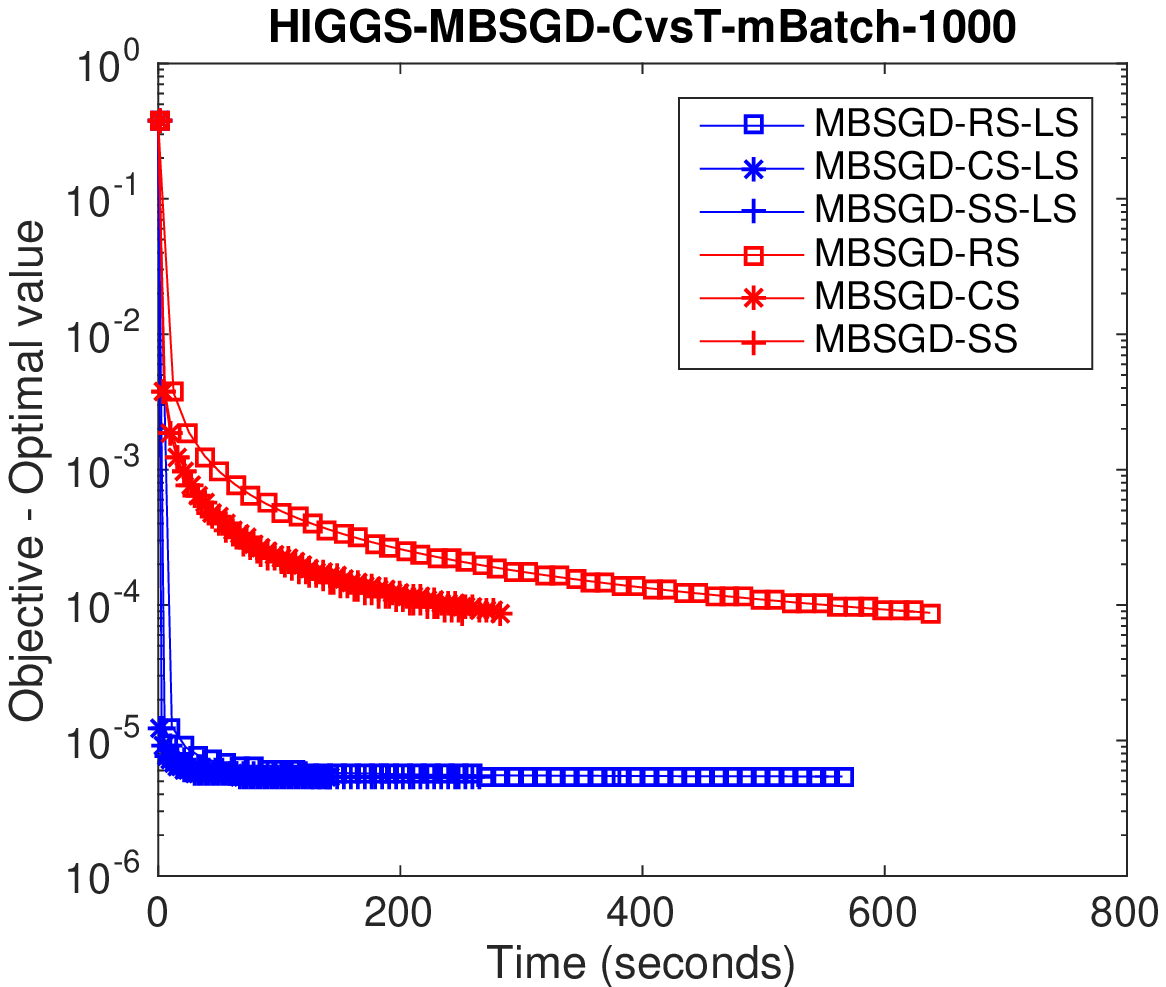}}
	\subfloat{\includegraphics[width=.25\linewidth]{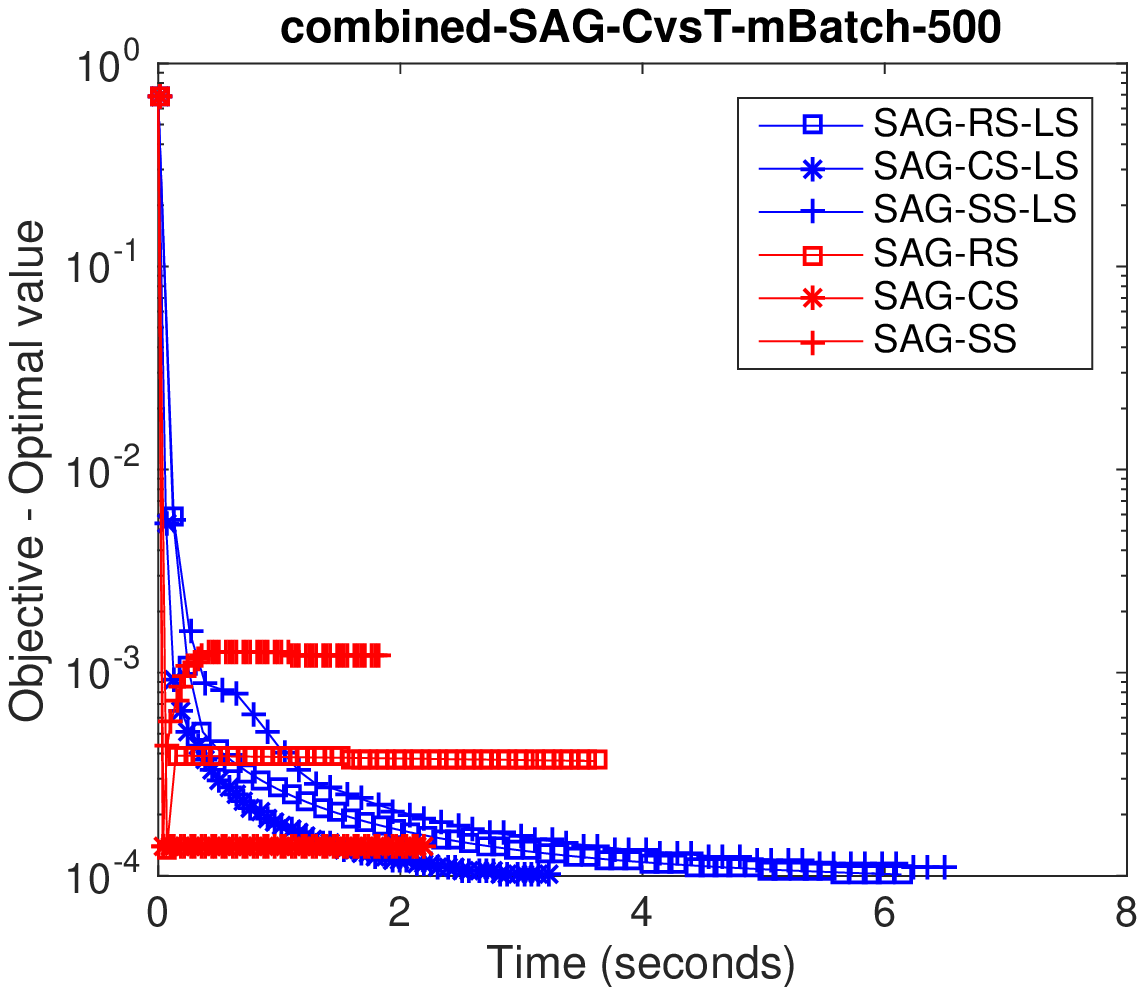}}
	\subfloat{\includegraphics[width=.25\linewidth]{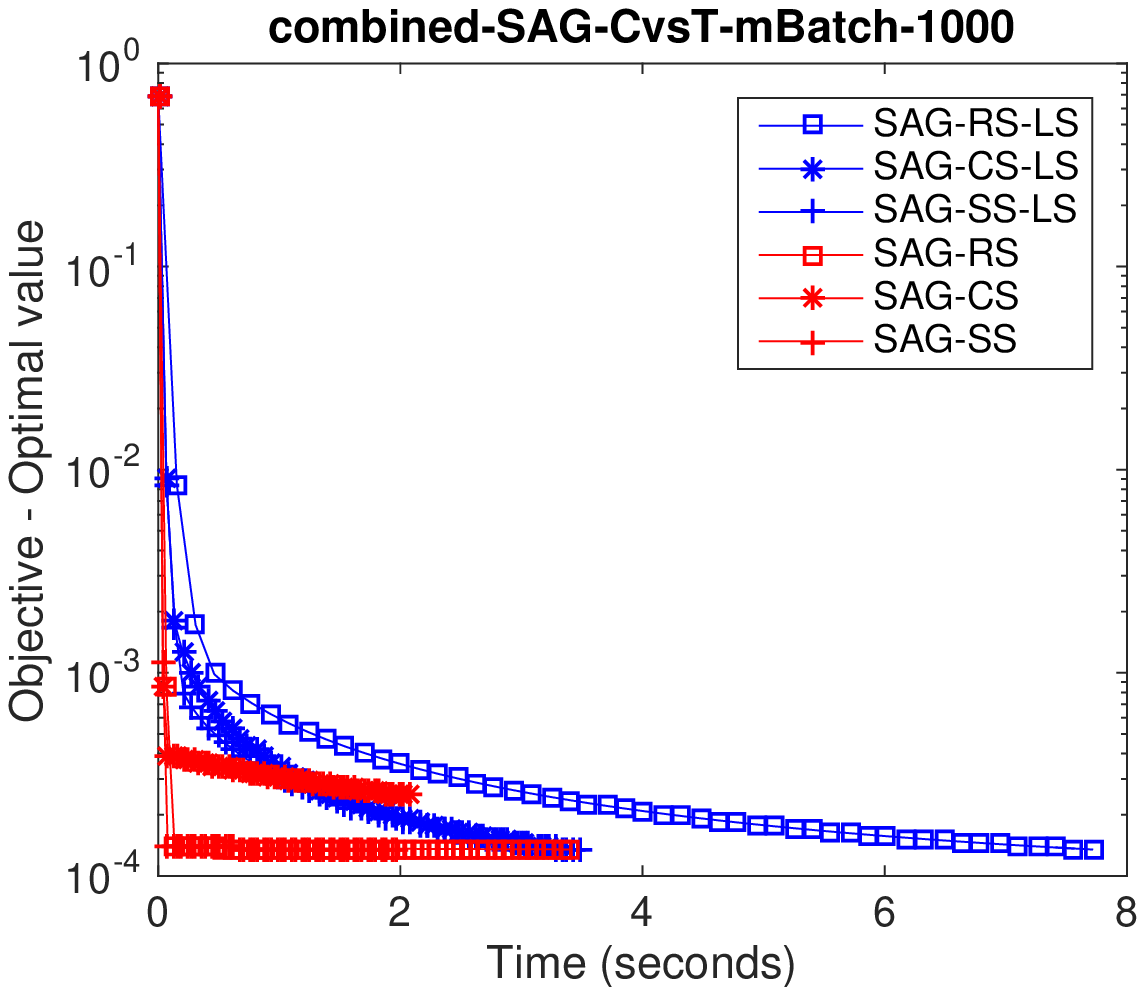}}
	
	\subfloat{\includegraphics[width=.25\linewidth]{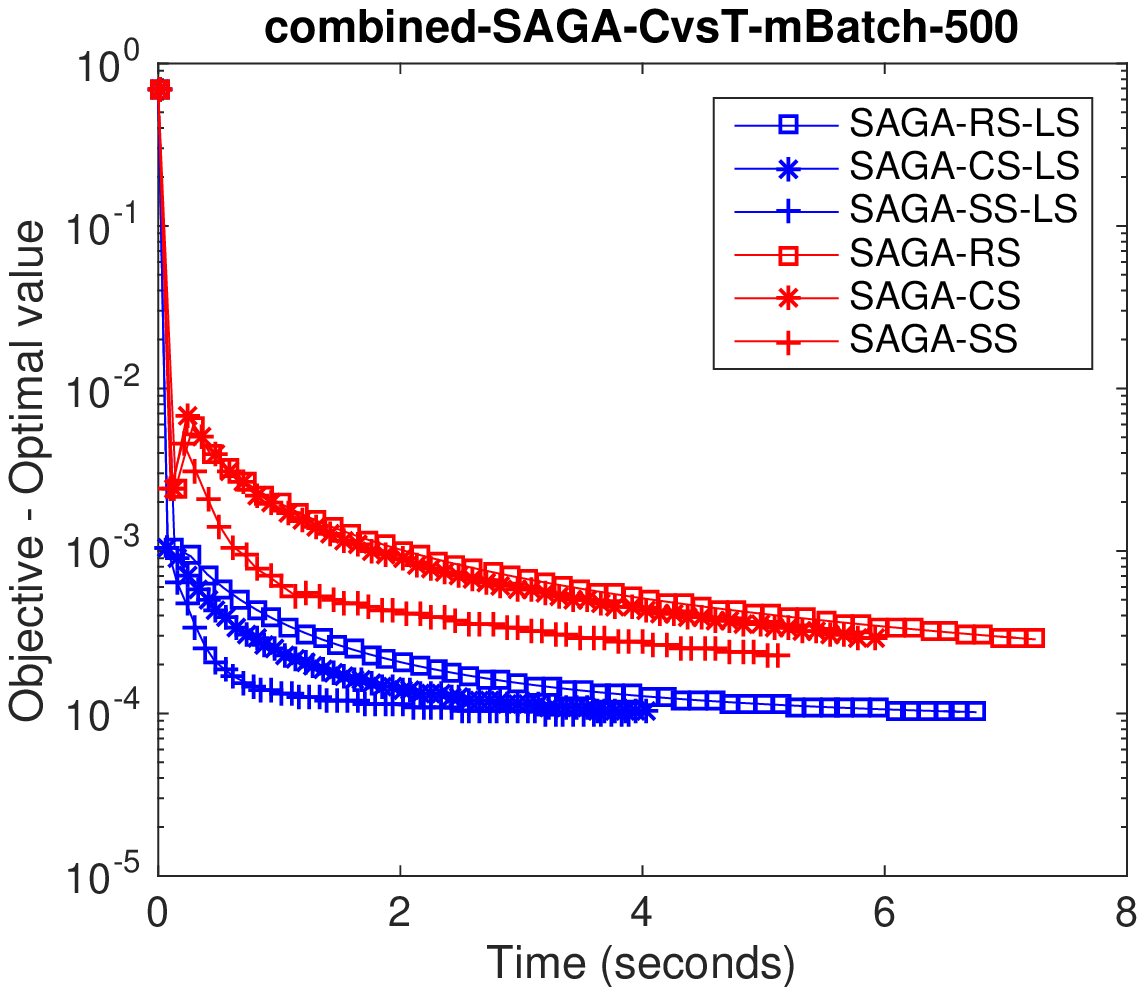}}
	\subfloat{\includegraphics[width=.25\linewidth]{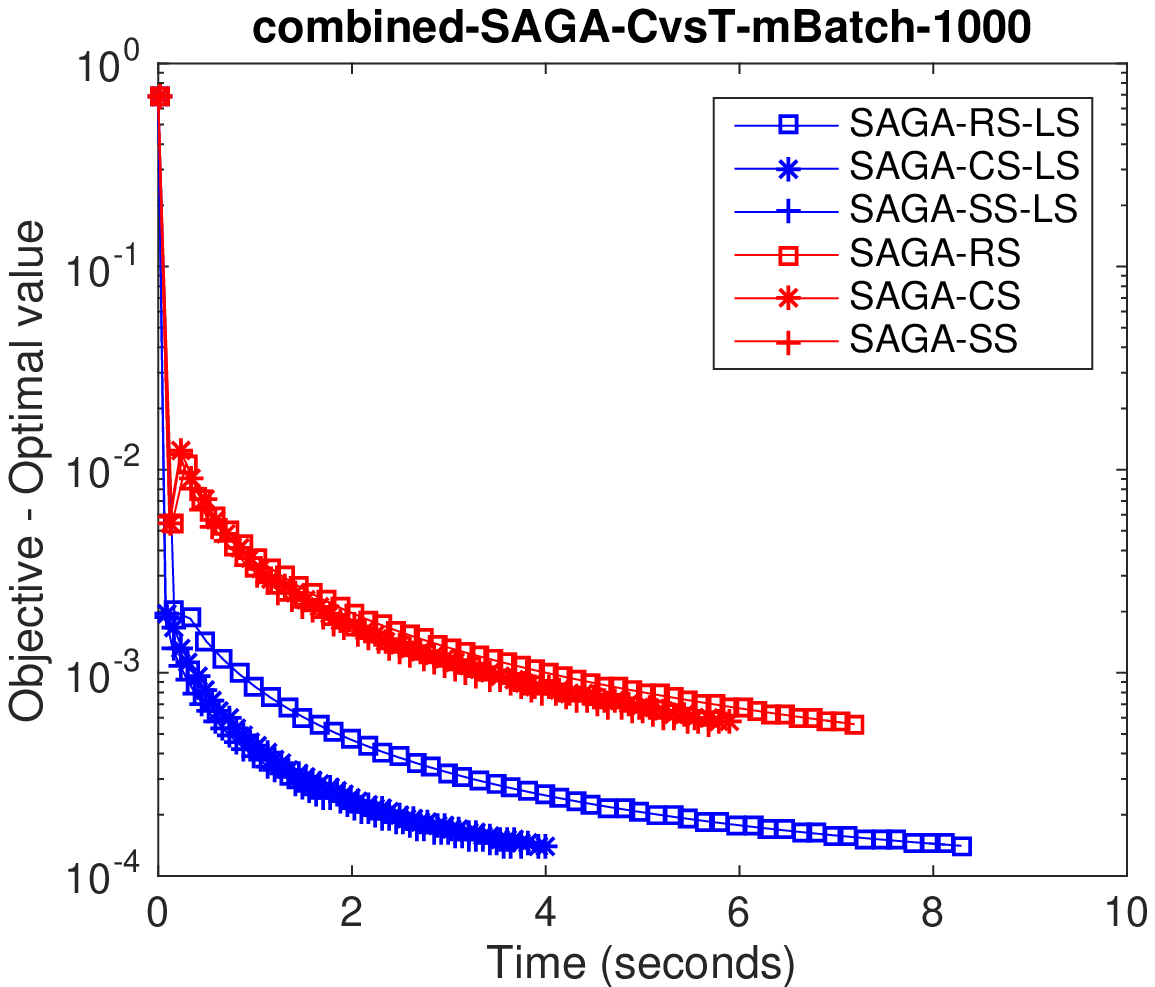}}
	\subfloat{\includegraphics[width=.25\linewidth]{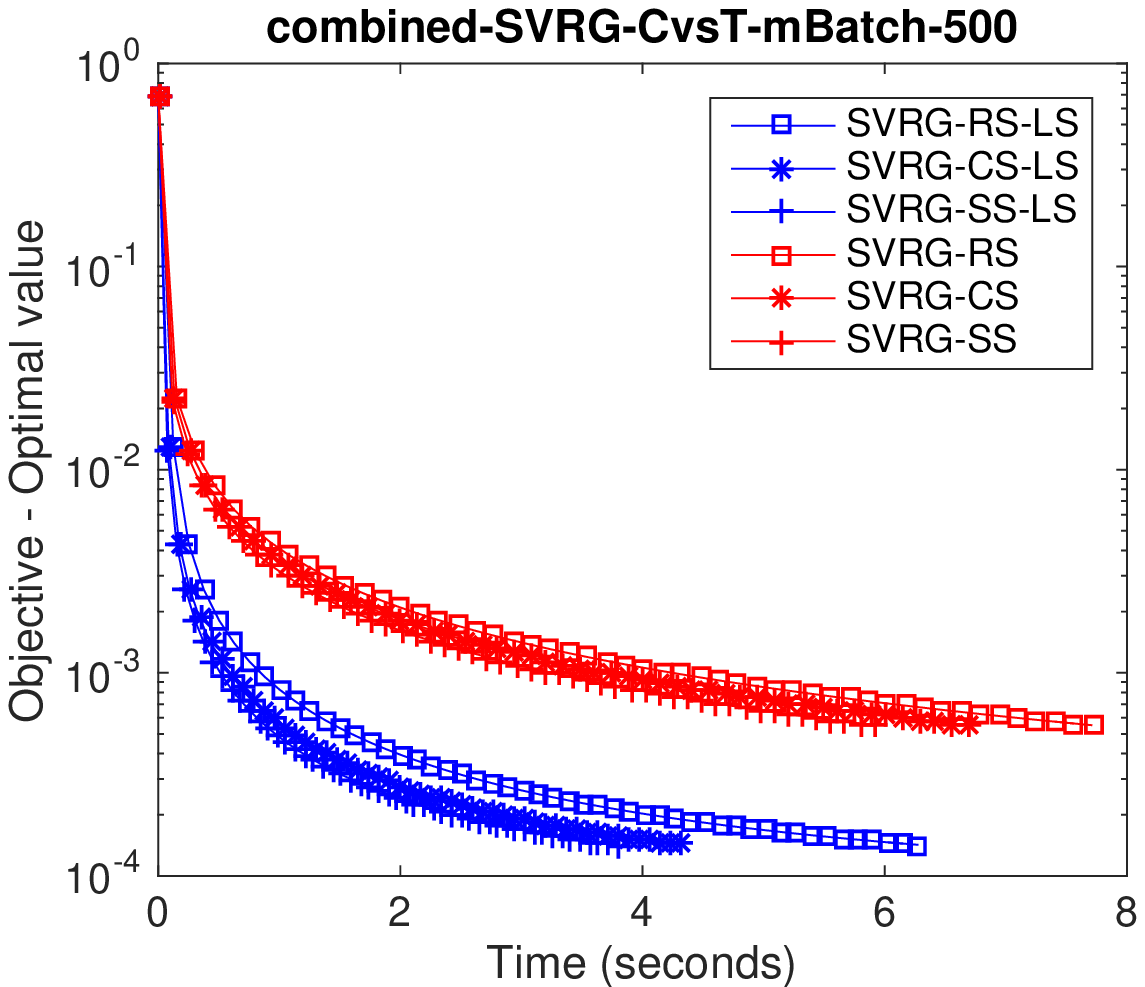}}
	\subfloat{\includegraphics[width=.25\linewidth]{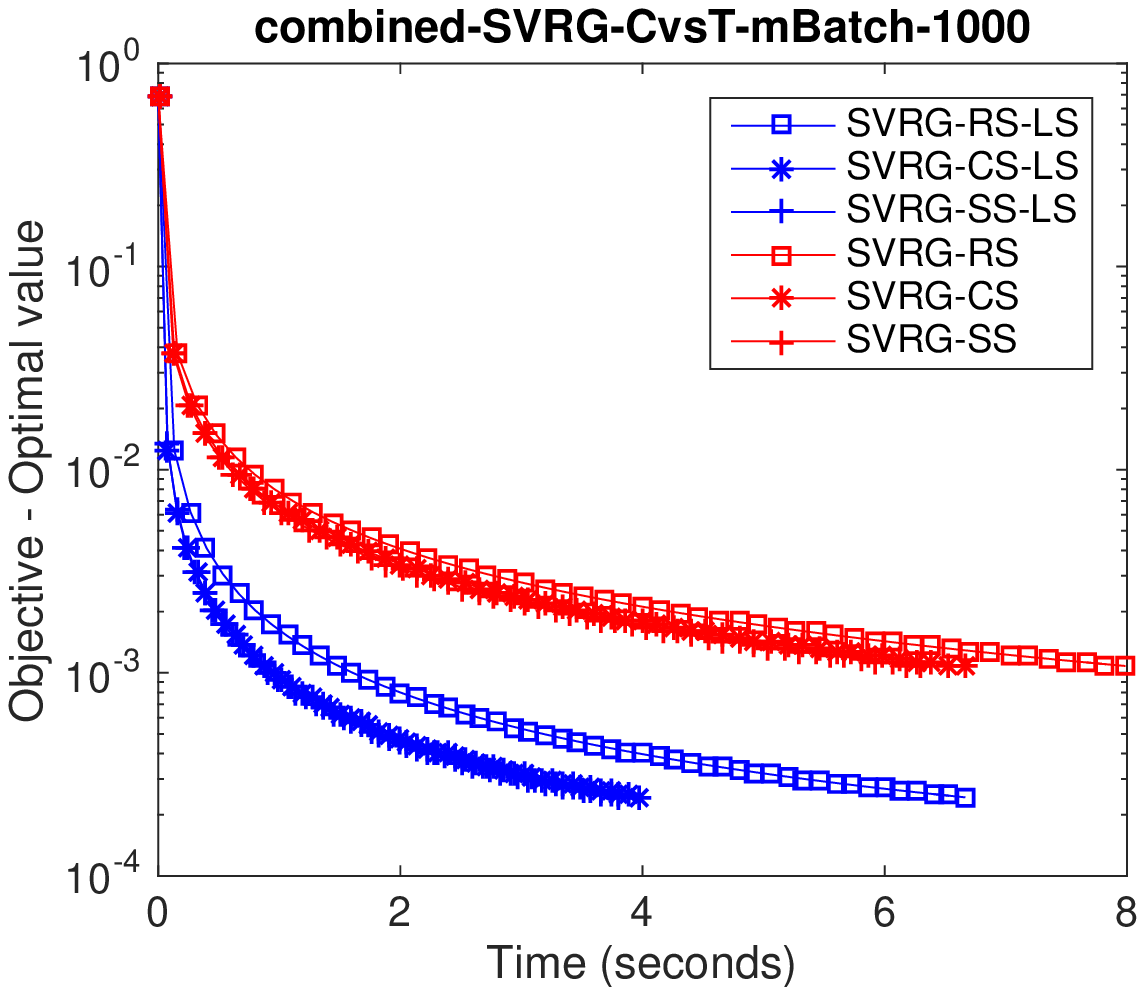}}
	
	\subfloat{\includegraphics[width=.25\linewidth]{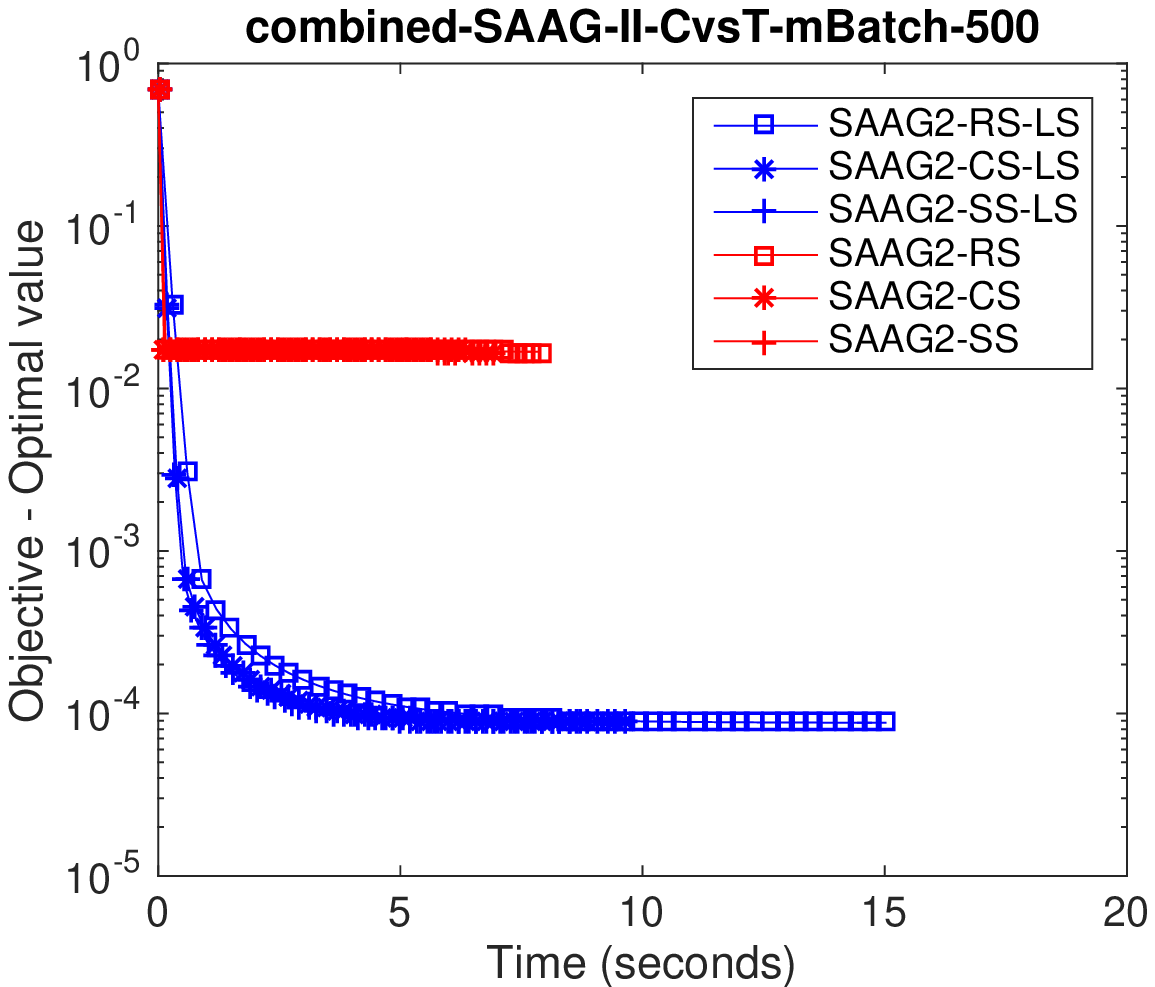}}
	\subfloat{\includegraphics[width=.25\linewidth]{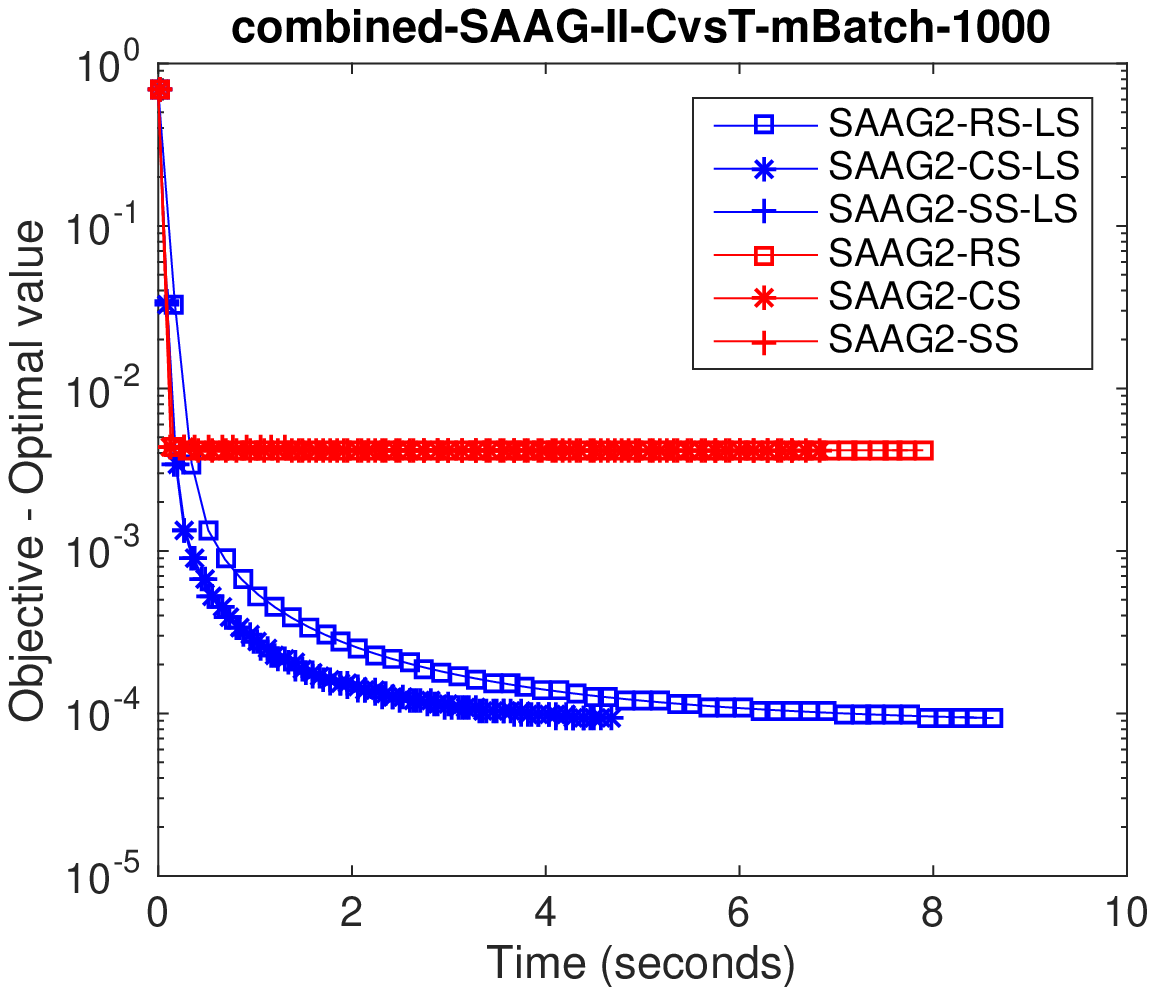}}	
	\subfloat{\includegraphics[width=.25\linewidth]{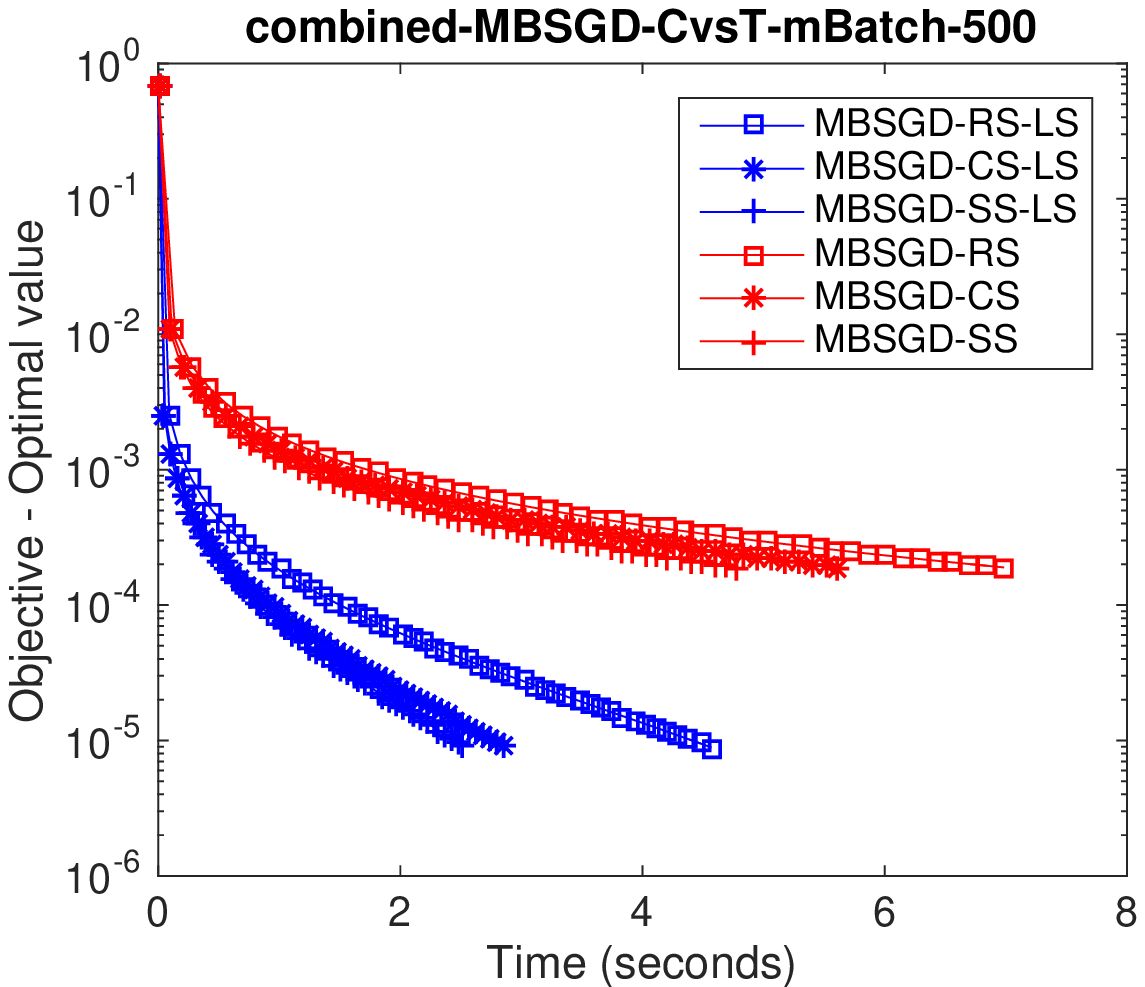}}
	\subfloat{\includegraphics[width=.25\linewidth]{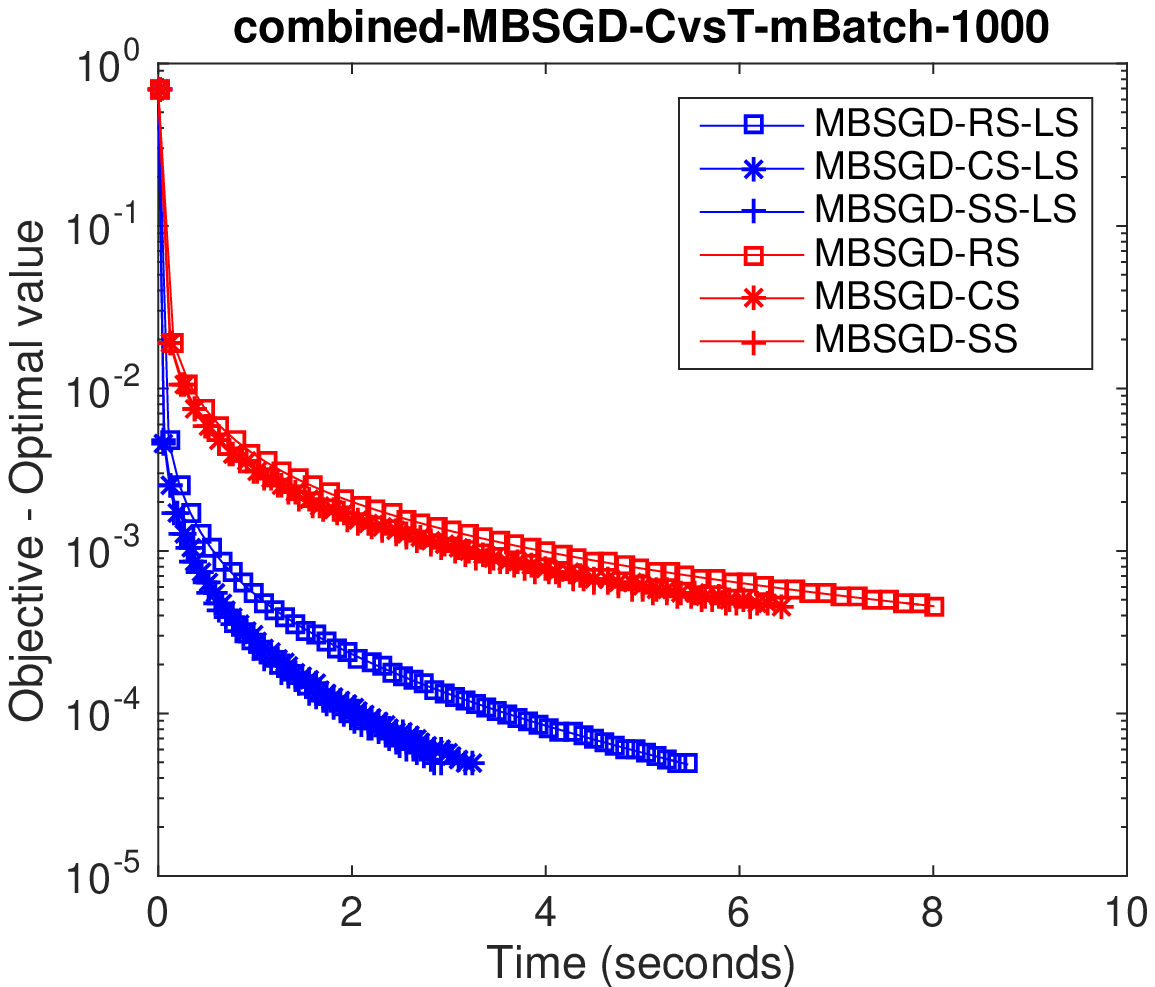}}
	
	\caption{RS, CS and SS are compared using SAG, SAGA, SVRG, SAAG-II and MBSGD, each with two step determination techniques, namely, constant step and backtracking line search, over HIGGS and SensIT (combined) datasets with mini-batch of 500 and 1000 data points.}
	\label{fig_3}
\end{figure}

\begin{figure}[htb]
	
	\subfloat{\includegraphics[width=.25\linewidth]{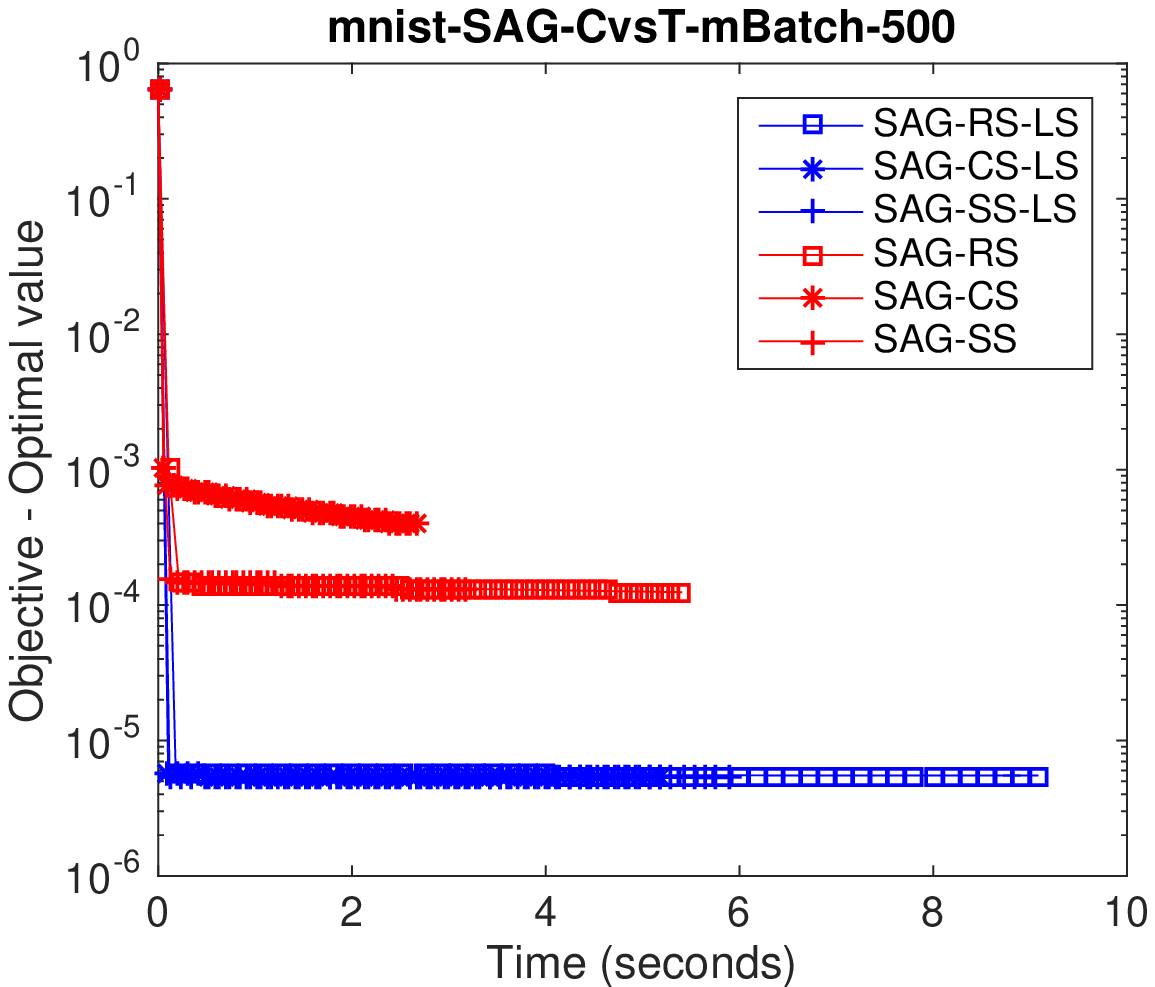}}
	\subfloat{\includegraphics[width=.25\linewidth]{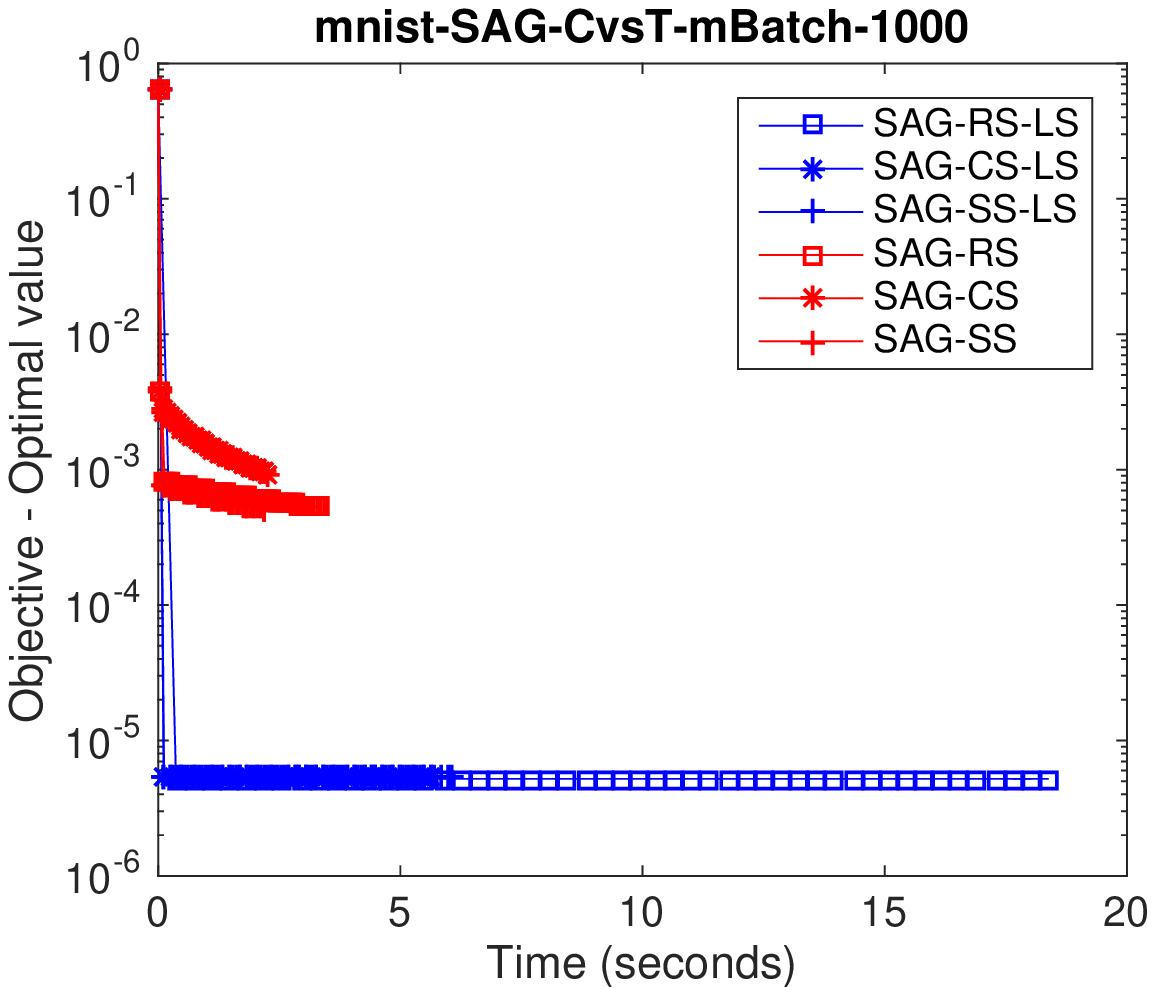}}
	\subfloat{\includegraphics[width=.25\linewidth]{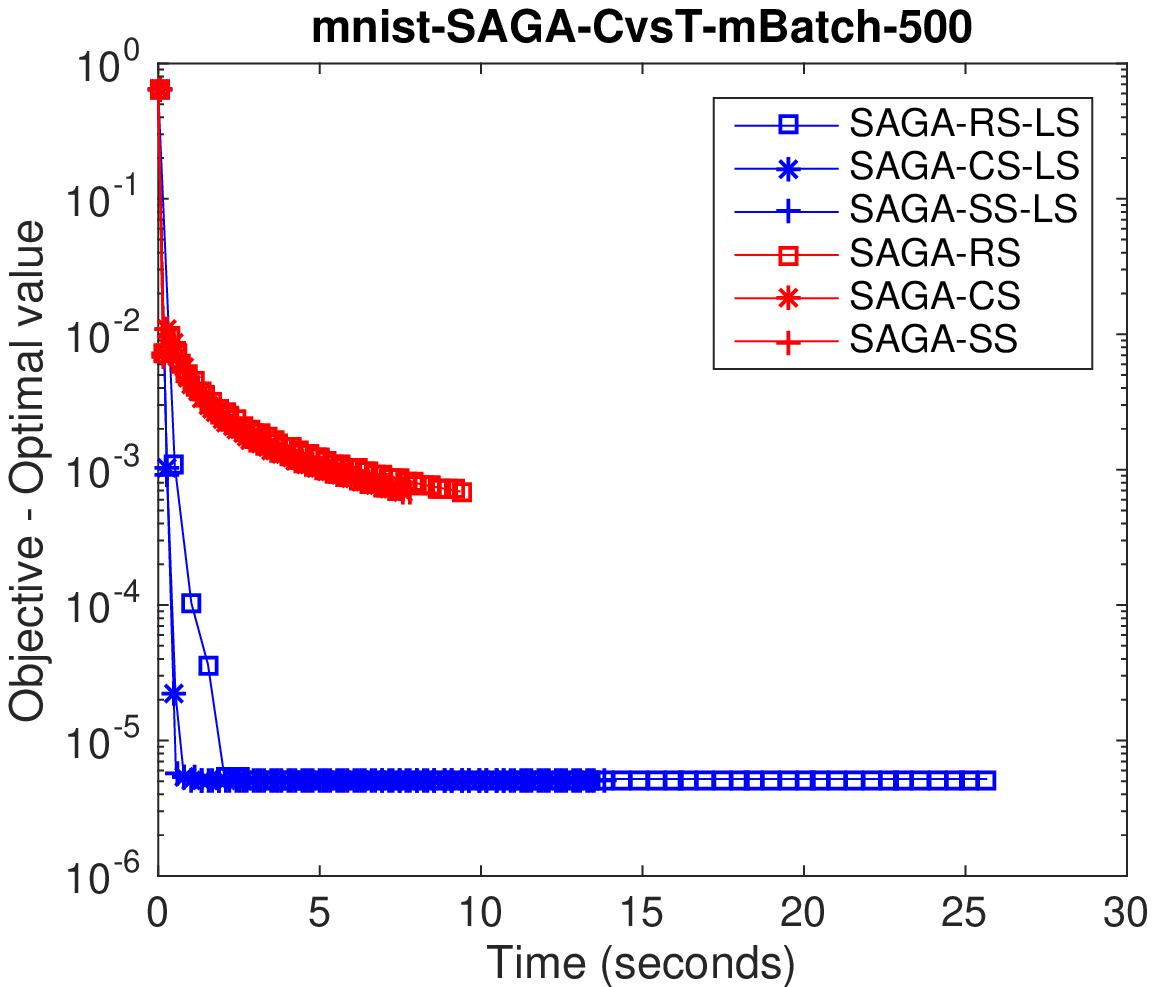}}
	\subfloat{\includegraphics[width=.25\linewidth]{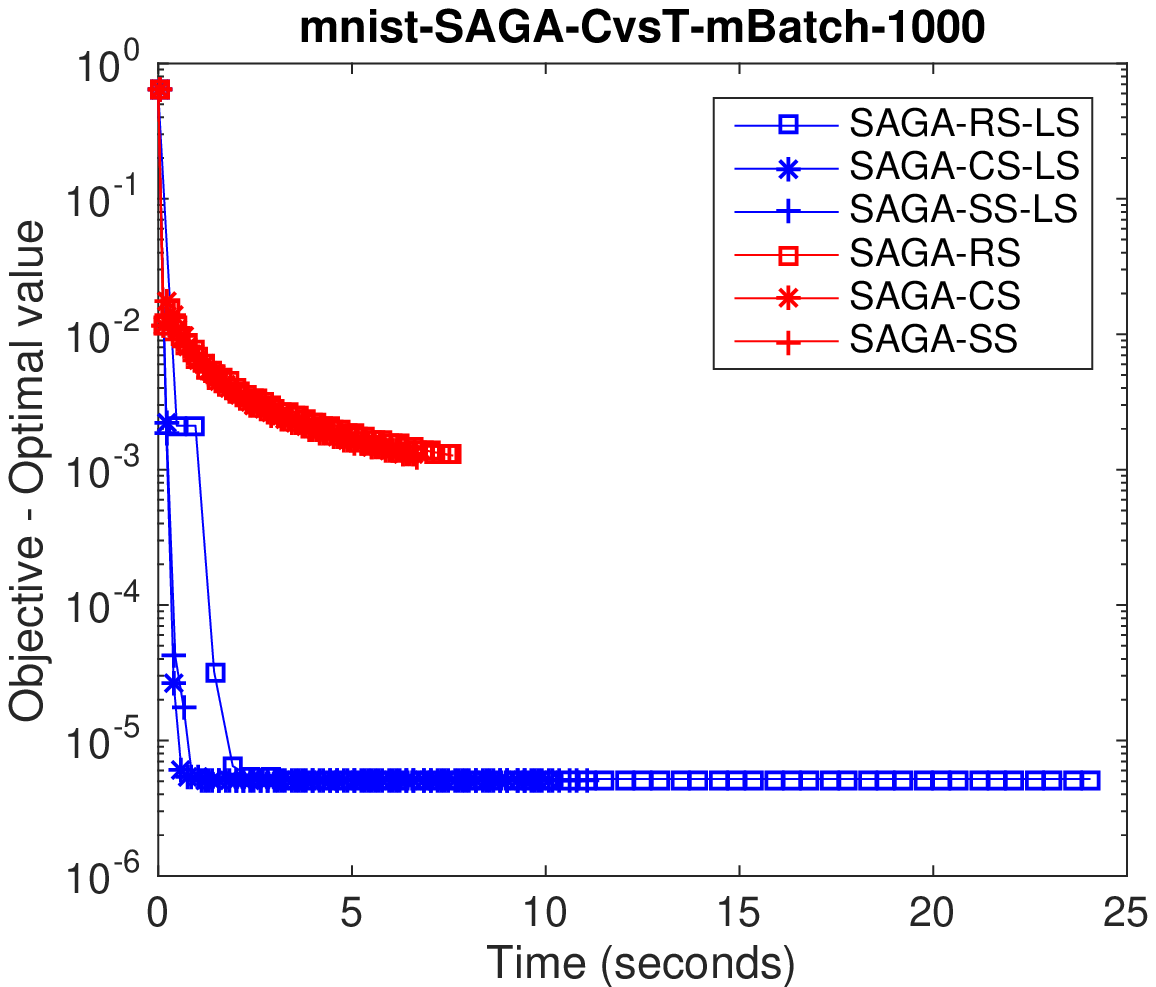}}
	
	\subfloat{\includegraphics[width=.25\linewidth]{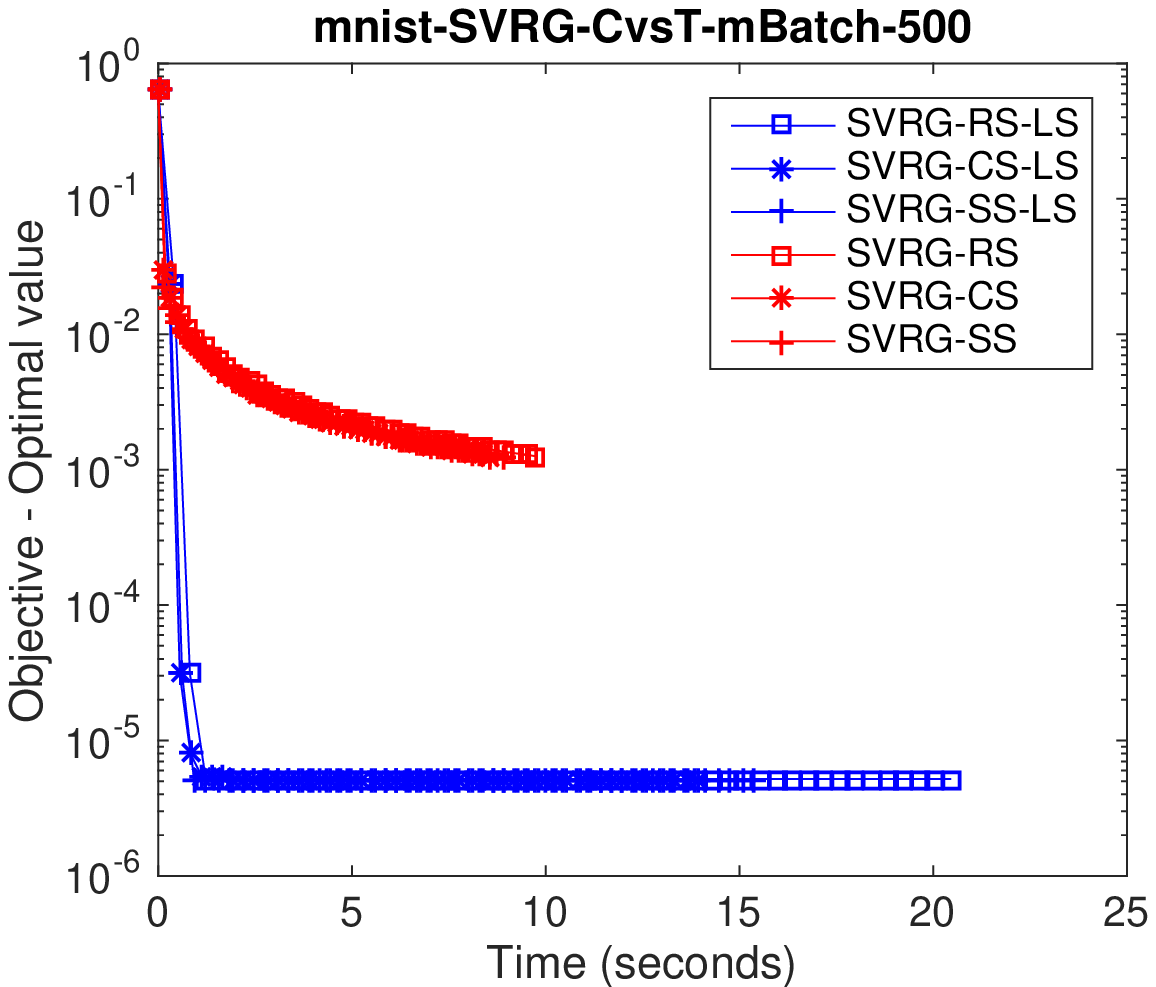}}
	\subfloat{\includegraphics[width=.25\linewidth]{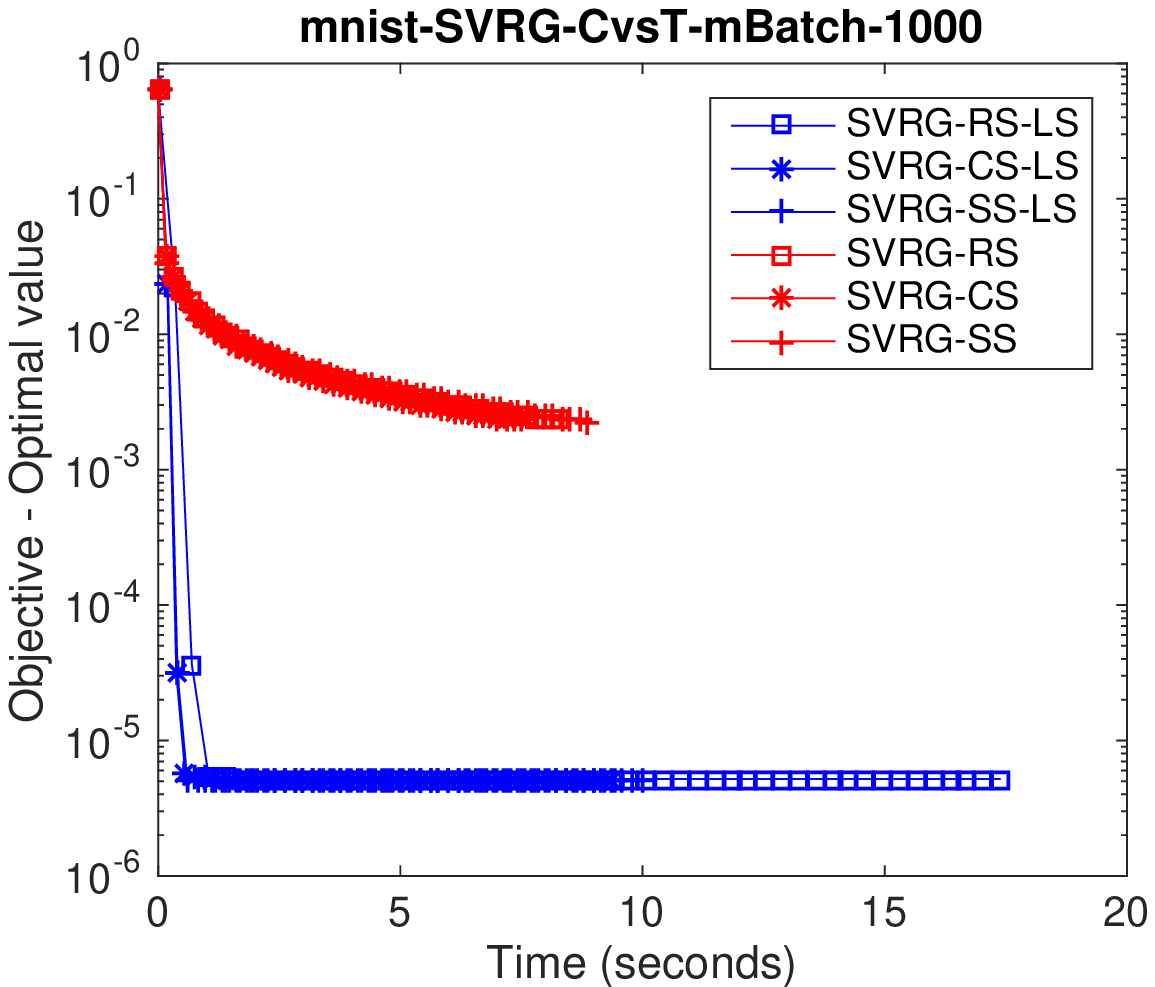}}
	\subfloat{\includegraphics[width=.25\linewidth]{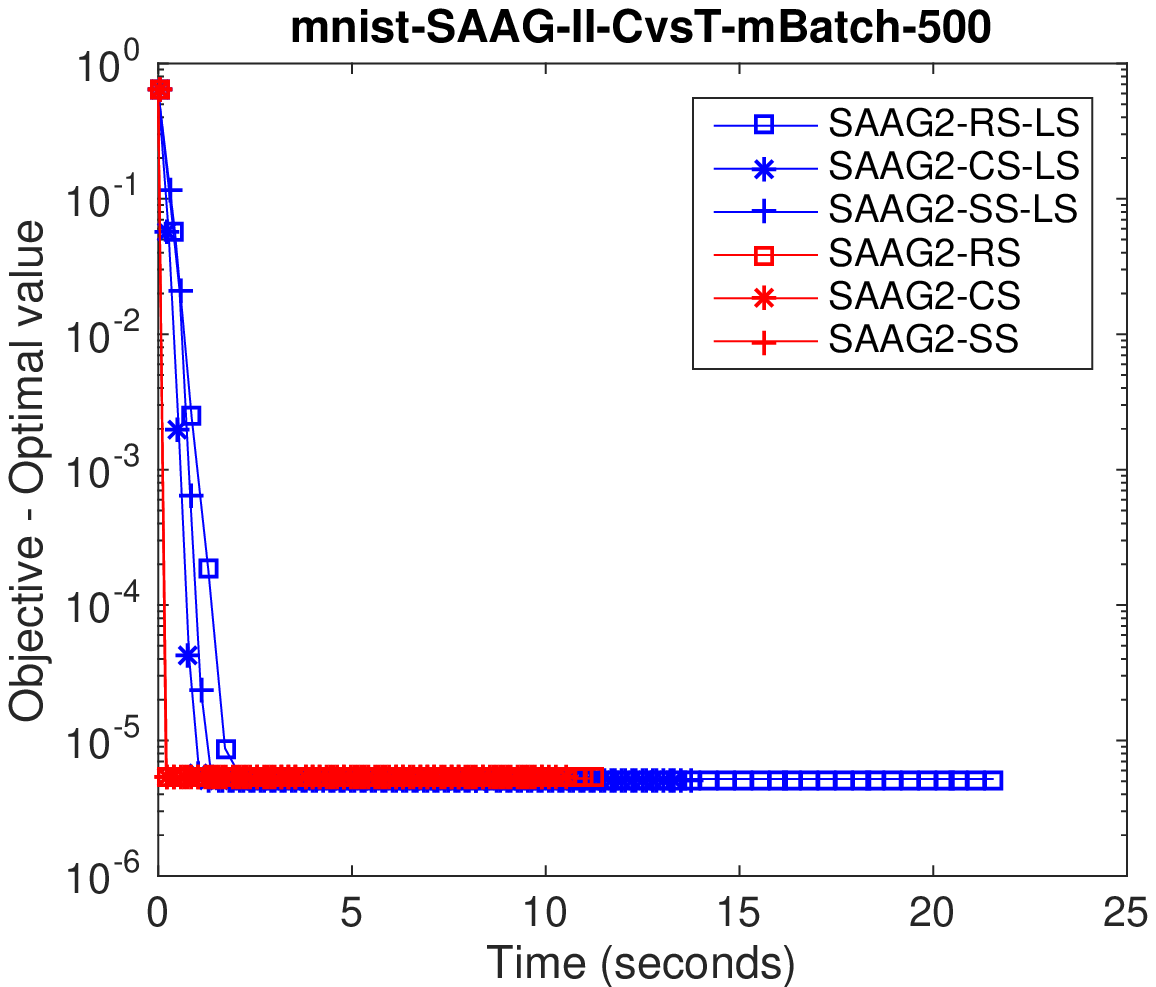}}
	\subfloat{\includegraphics[width=.25\linewidth]{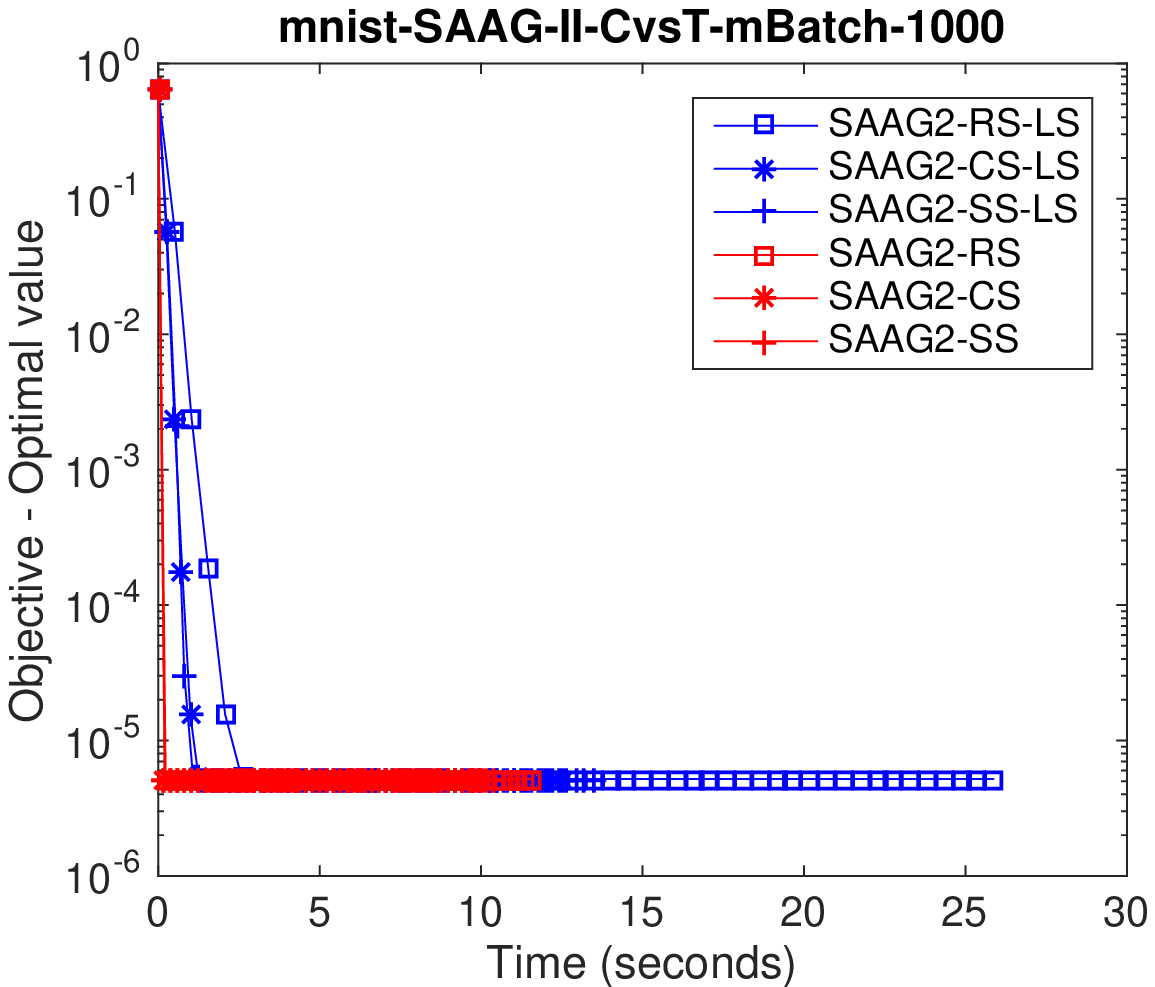}}
	
	\subfloat{\includegraphics[width=.25\linewidth]{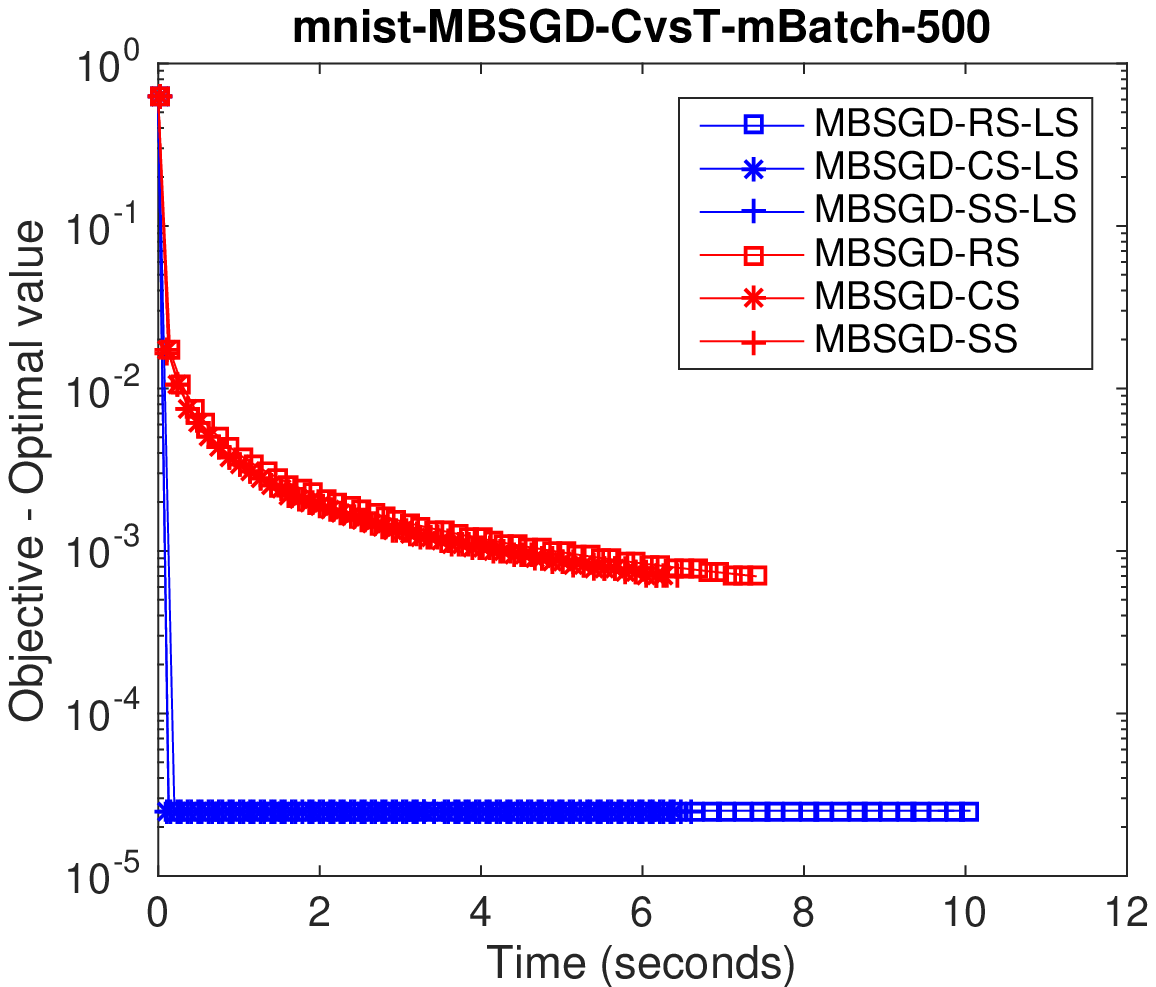}}
	\subfloat{\includegraphics[width=.25\linewidth]{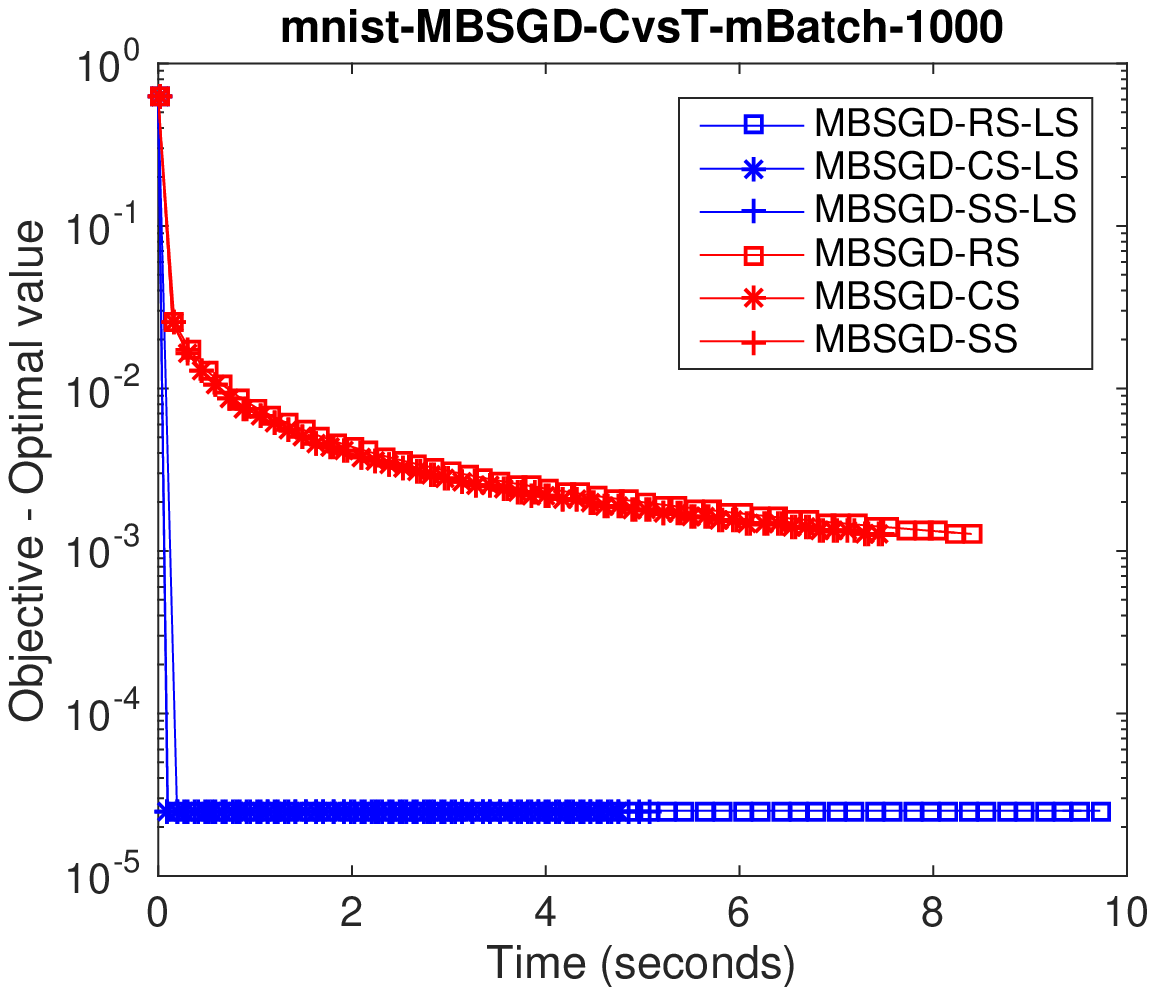}}
	\subfloat{\includegraphics[width=.25\linewidth]{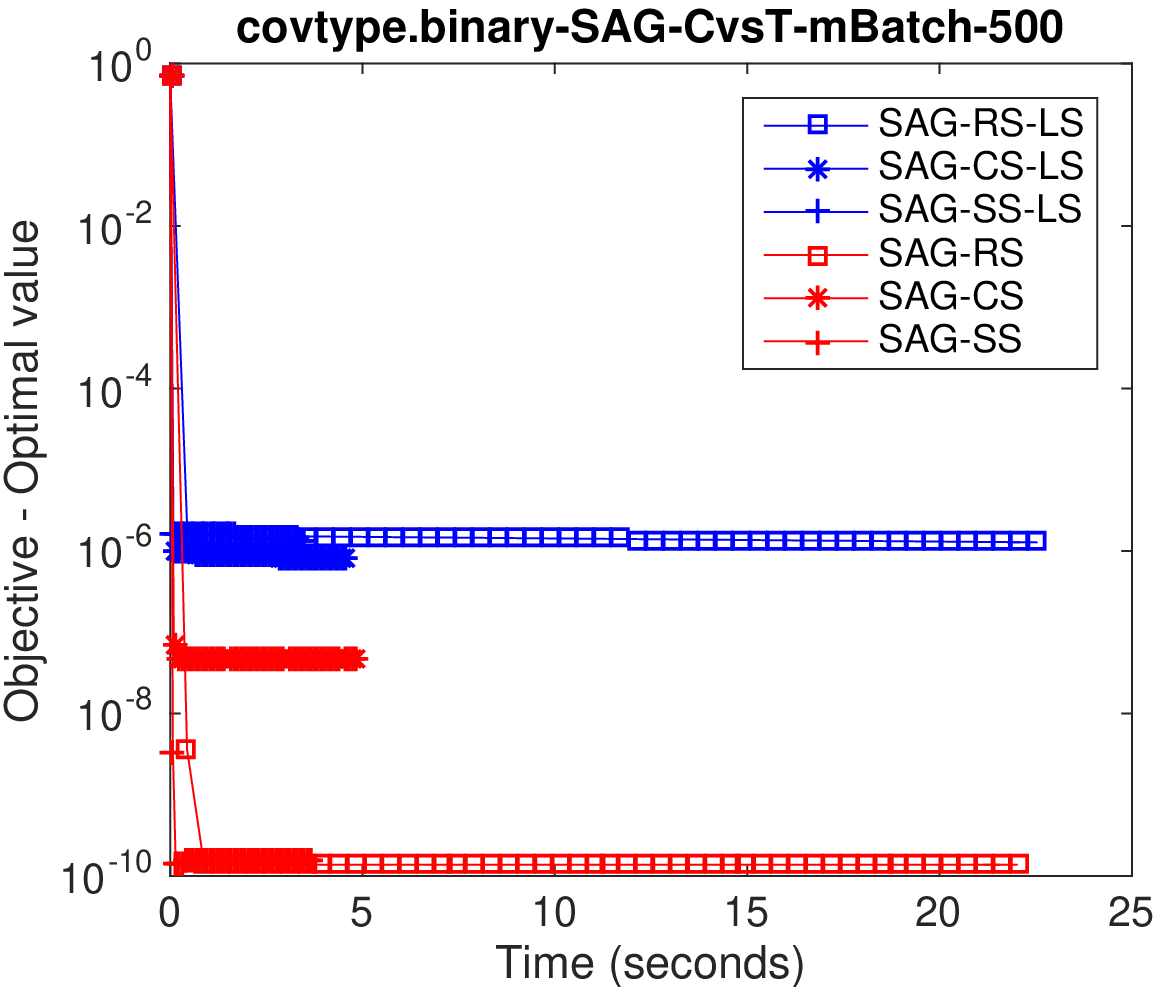}}
	\subfloat{\includegraphics[width=.25\linewidth]{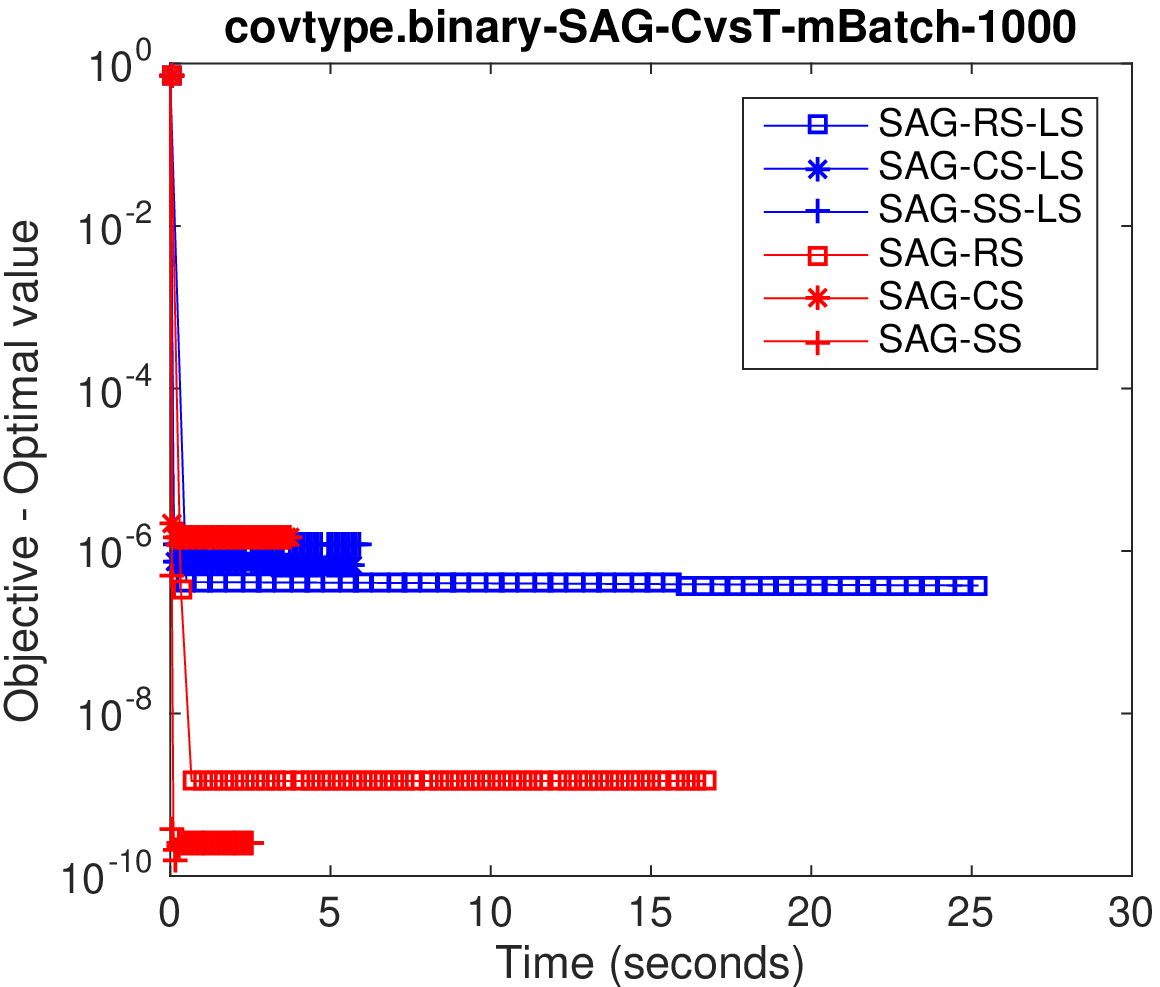}}
	
	\subfloat{\includegraphics[width=.25\linewidth]{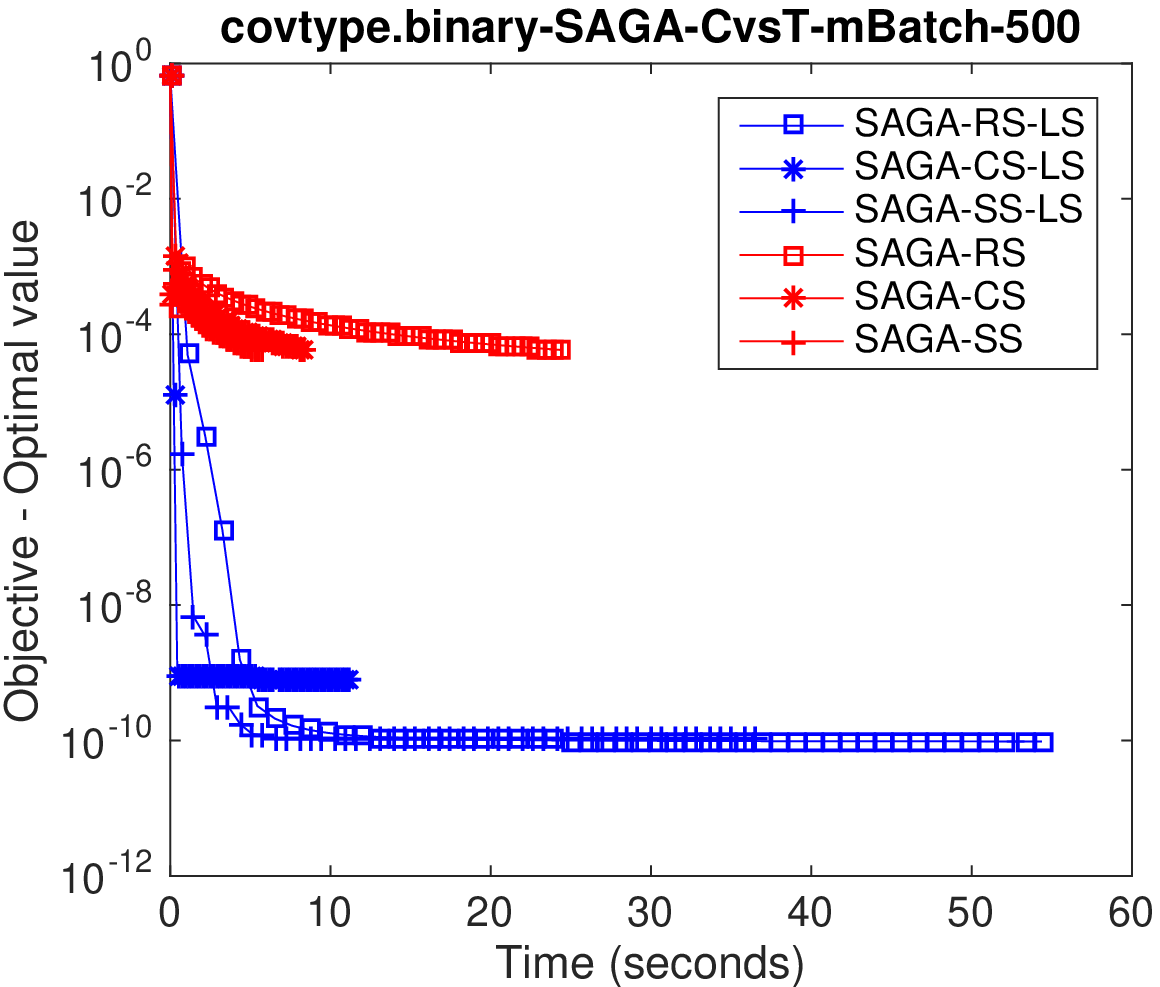}}
	\subfloat{\includegraphics[width=.25\linewidth]{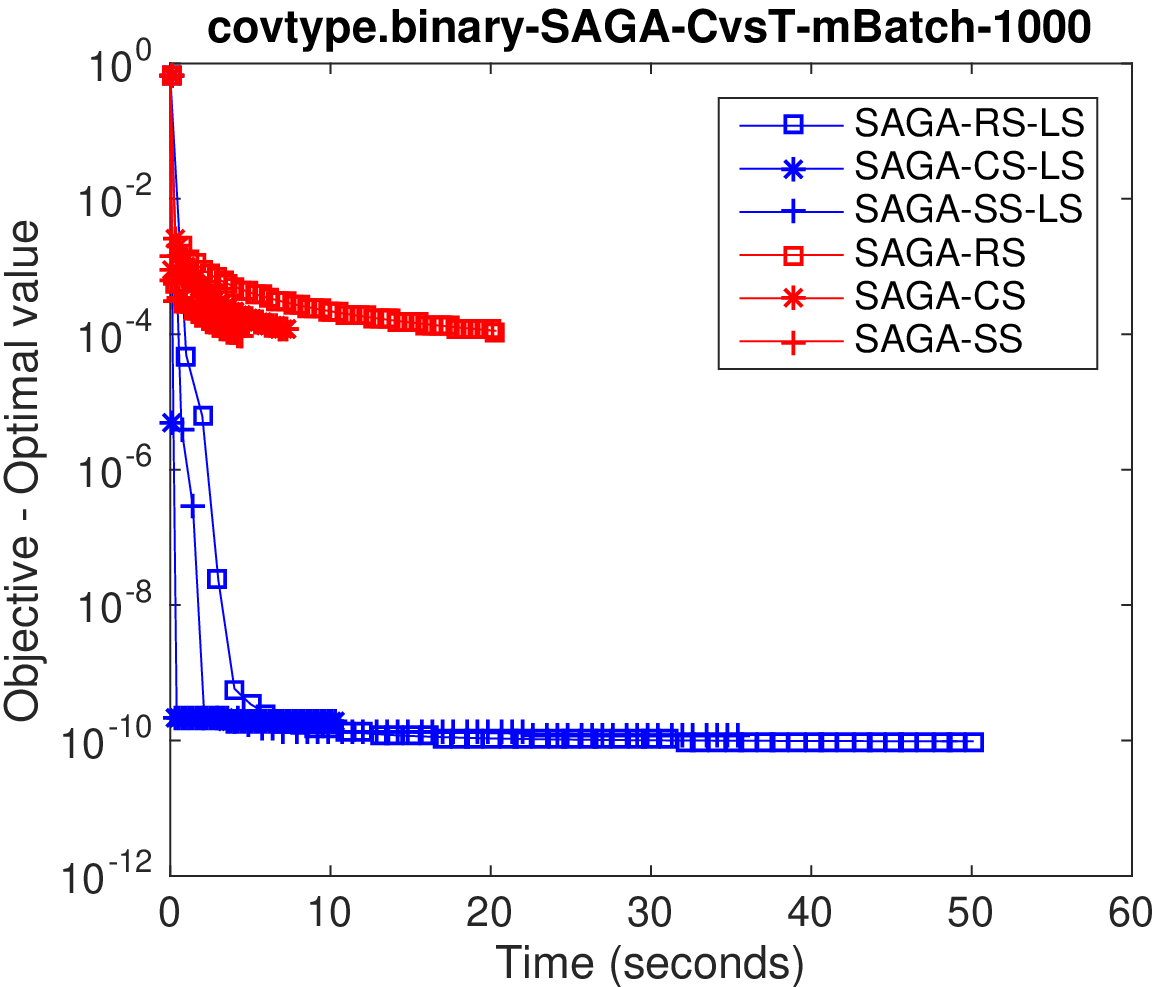}}
	\subfloat{\includegraphics[width=.25\linewidth]{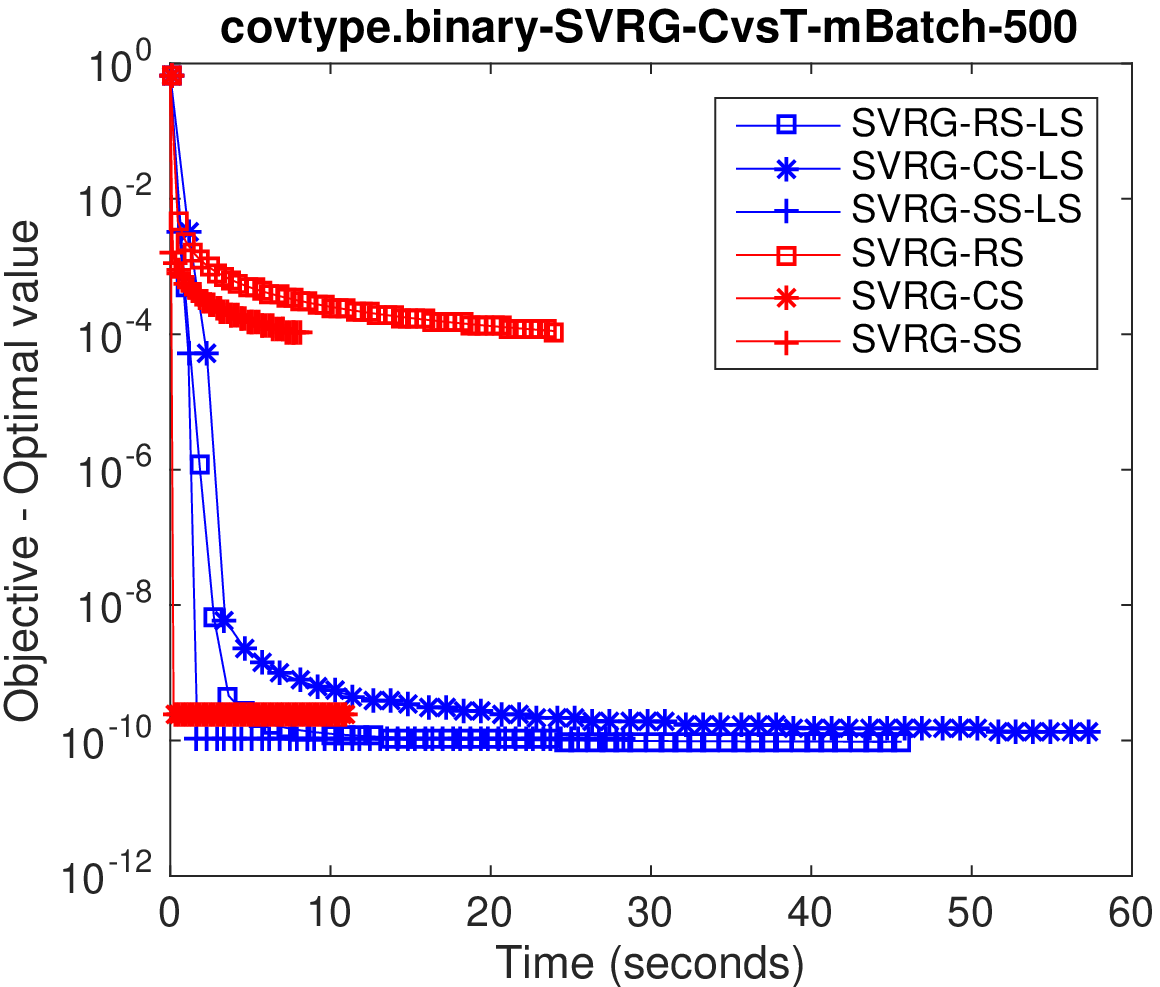}}
	\subfloat{\includegraphics[width=.25\linewidth]{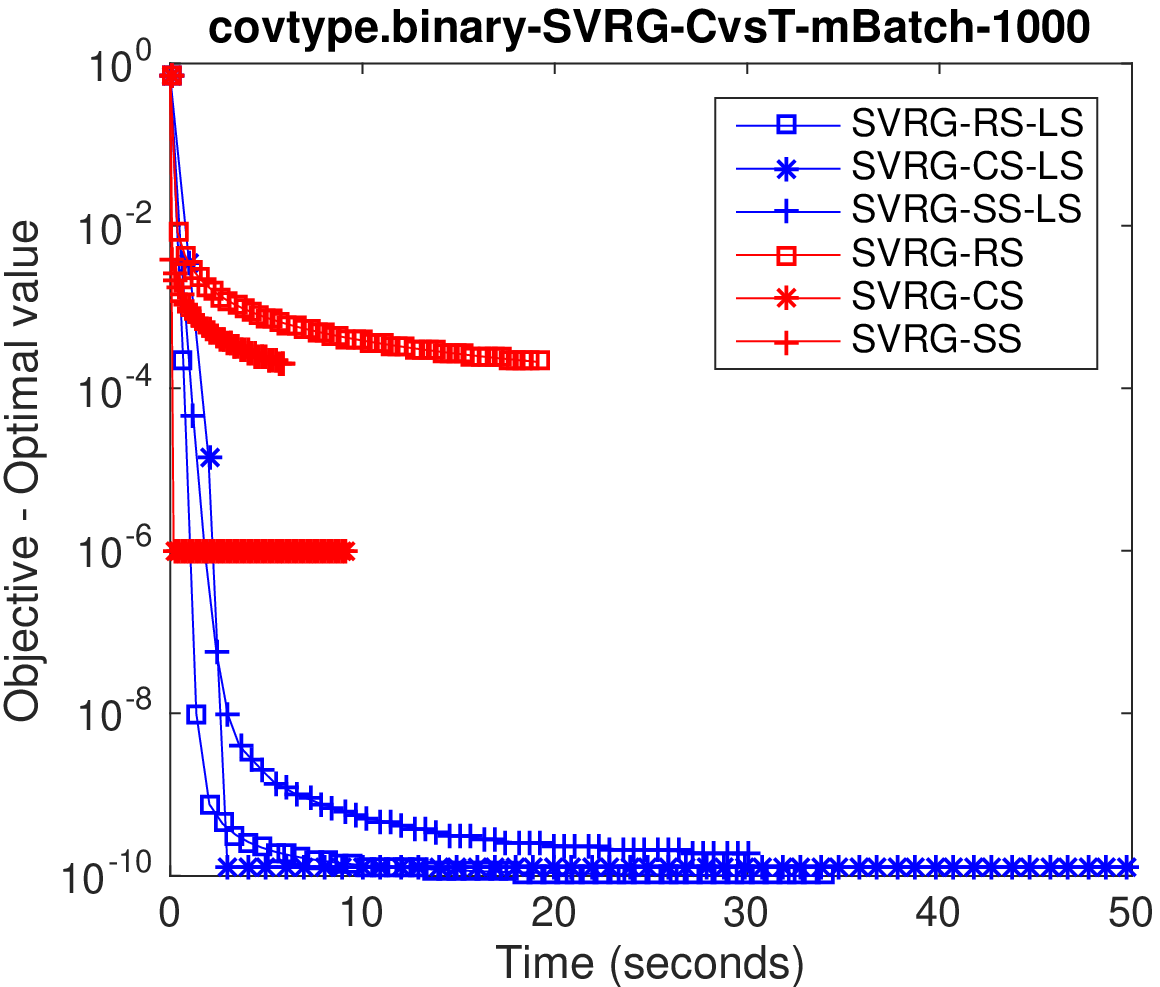}}
	
	\subfloat{\includegraphics[width=.25\linewidth]{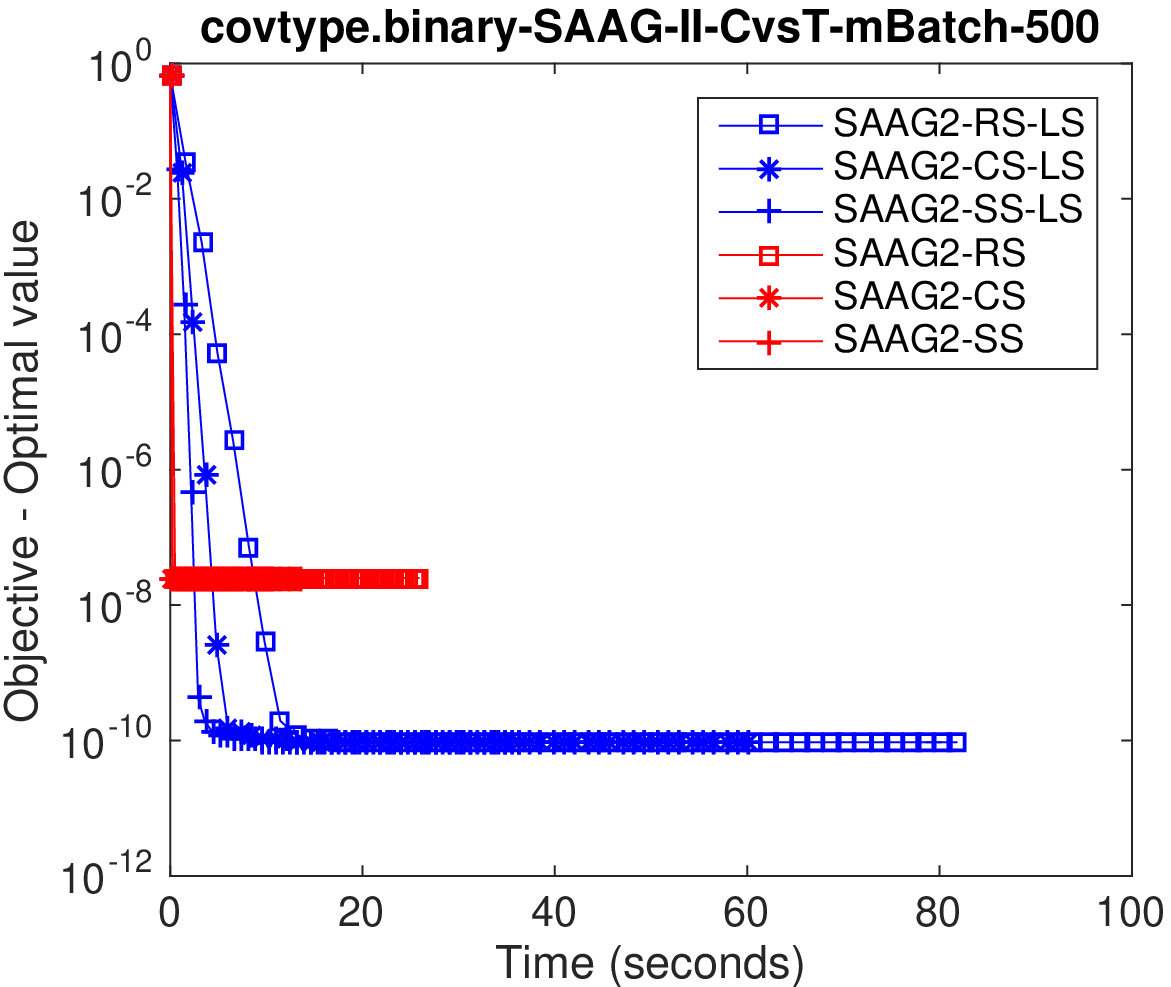}}
	\subfloat{\includegraphics[width=.25\linewidth]{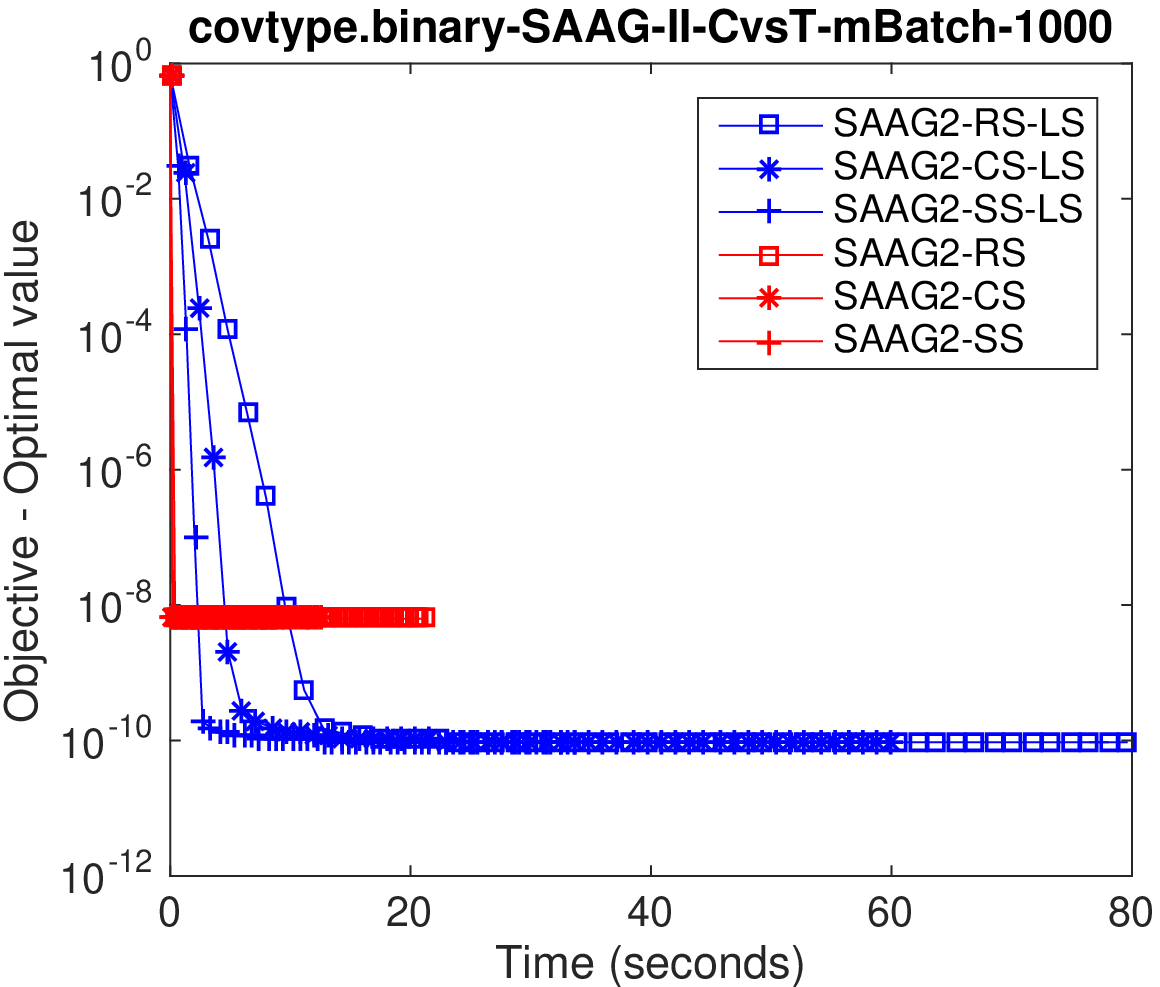}}
	\subfloat{\includegraphics[width=.25\linewidth]{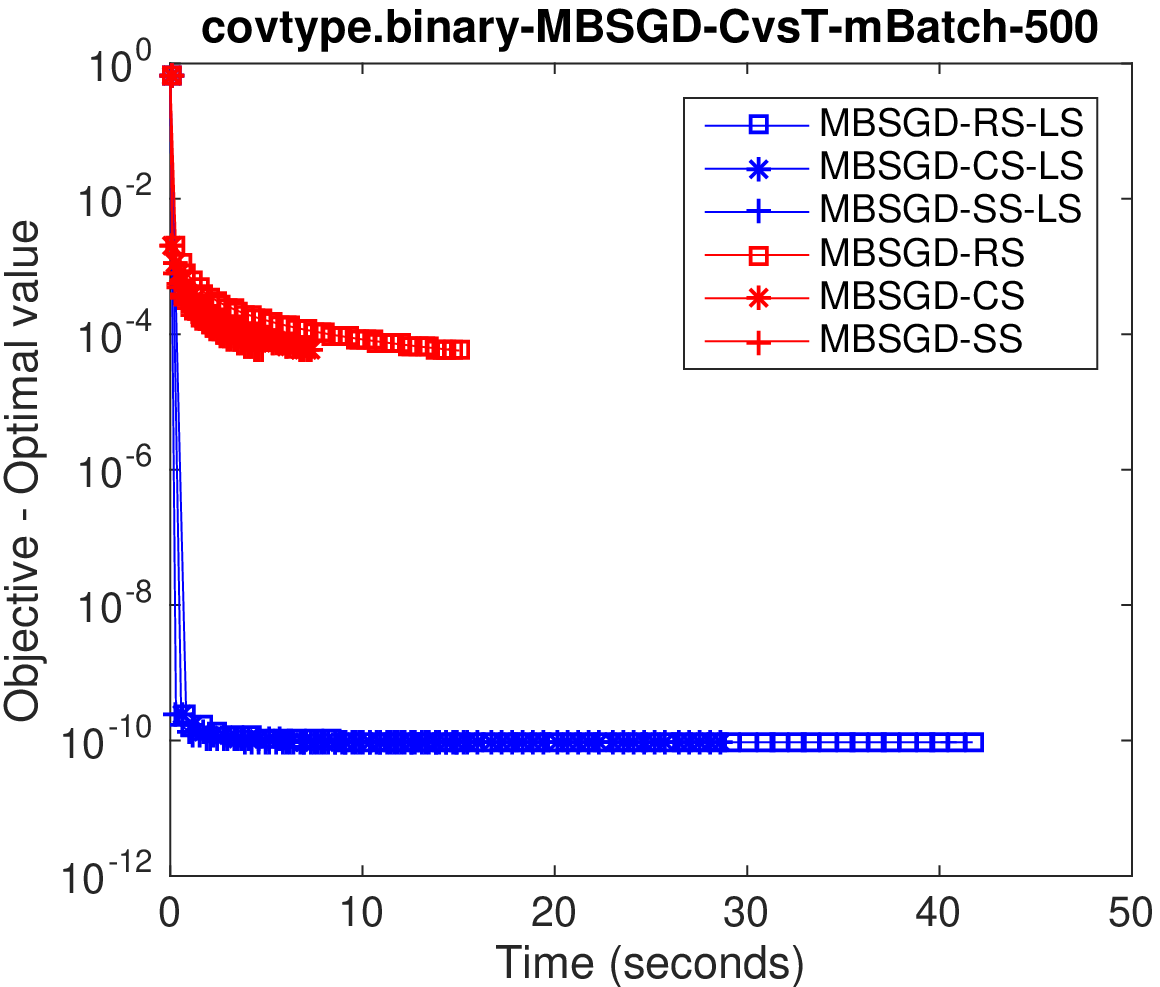}}
	\subfloat{\includegraphics[width=.25\linewidth]{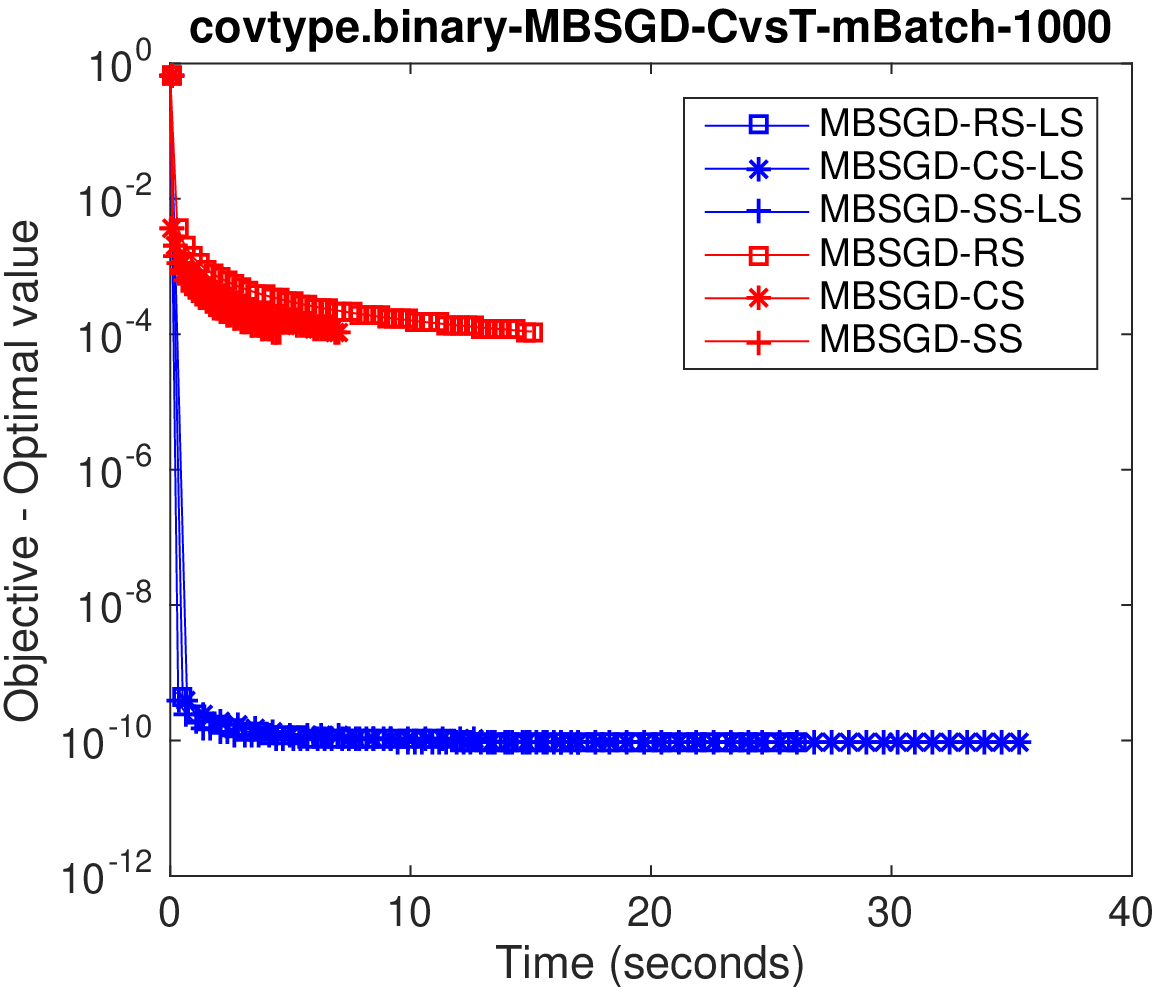}}
	
	\caption{RS, CS and SS are compared using SAG, SAGA, SVRG, SAAG-II and MBSGD, each with two step determination techniques, namely, constant step and backtracking line search, over mnist and covtype datasets with mini-batch of 500 and 1000 data points.}
	\label{fig_4}
\end{figure}

Experimental results\footnote{Experimental results can be reproduced using the code available at following link: \url{https://sites.google.com/site/jmdvinodjmd/code}} plot the difference between objective function and optimum value against training time for three sampling techniques, namely, Random Sampling (RS), Cyclic/Sequential Sampling (CS) and Systematic Sampling (SS), which are represented by Figs.\ref{fig_1}--\ref{fig_4}. To save space, results for different samplings with constant step size and backtracking line search methods are plotted in same figure (red color for constant step size and blue for line search (LS)). As it is clear from figures, 20 different settings over one dataset compare CS, SS and RS, and prove that methods with CS and SS converges faster than with RS. In general, SS gives the best results as per the intuition but sometimes CS produces better results than SS since CS and SS are quite similar. The results with constant step size and backtracking line search methods show similar results. For larger datasets, like SUSY and HIGGS, SS and CS show clear advantage over RS, than for smaller dataset and thus prove the efficacy/suitability of CS and SS for large-scale learning problems. For some of the results, like, with HIGGS dataset and SAG method, learning algorithms with all the sampling techniques converge quickly to same value but careful examination reveals that CS and SS converges earlier than RS.\\
Experimental results can be presented using tables comparing the training time and objective function value, for given number of epochs (number of passes through dataset), for CS, SS and RS. Since it is easy to understand the graphs so experiments are documented using figures. Results for SUSY, HIGGS and covtype.binary are also represented using Tables~\ref{tab_HIGGS}, \ref{tab_SUSY} and \ref{tab_covtype}, which prove faster training for CS and SS than RS, while their values are same up to certain decimal places. As it is clear from the table, SAG method with mini-batch of size 200 and constant step size, the training time for Systematic Sampling (SS) and Cyclic Sampling (CS) are faster than for Random Sampling (RS) by a factor of more than three while the values of objective function are same up to four decimal places. The training times for SS and CS are very close as expected. For mini-batch of 1000 data points and constant step size, SAG method runs three and half times faster for SS and CS as compared with RS while the values of objective function are same up to six decimal places. With backtracking line search for determining the step size, SAG method runs more than two times faster with mini-batch of 200 and more than three and half times with mini-batch of size 1000, for SS and CS as compared with RS while the values of objective function are same up to eight decimal places. SAGA with constant step size runs more than two times faster with mini-batches of 200 and 1000 for CS and SS as compared with RS while the values of objective function are same up to five and four decimal places for mini-batches of 200 and 1000, respectively. With backtracking line search, SAGA performs even better for CS and SS and runs more than two times faster with mini-batch of 200 and more than three and half times with mini-batch of size 1000 as compared with RS while the values of objective function are same up to eight decimal places with mini-batch of size 200 and up to six decimal places with mini-batch of size 1000. Similar trend are followed for SAAG-II, SVRG and MBSGD methods where CS and SS are around two to more than three and half times faster and objective values are same up to six to nine decimal places as compared  with RS with constant step and backtracking line search (LS).\\
Results with SUSY dataset are presented in Table~\ref{tab_SUSY}. As it is clear from the table, it shows good results like HIGGS dataset. CS and SS techniques are faster than RS technique by around two to four times while the objective function values are same up to three to eight decimal places. Methods with constant step and with backtracking line search show similar trends for the results.\\
Results with covtype.binary dataset are presented in Table~\ref{tab_covtype}. As it is clear from the table, it gives good results. CS and SS techniques are faster than RS technique by around one and half to six times while the objective function values are same up to three to ten decimal places. Methods with constant step and with backtracking line search show similar trends but for line search, objective function values are close among each other.\\

\begin{table*}[h]
	\centering
	\caption{Comparison of Training Time (in seconds) and objective function values after 30 epochs using dataset HIGGS}
	\label{tab_HIGGS}      
	\begin{tabular}{|l|c|r|r|r|r|r|}
		\hline
		\multirow{2}{*}{Method} & \multirow{2}{*}{Sampling} & \multirow{2}{*}{Batch} & \multicolumn{2}{c|}{Constant Step} & \multicolumn{2}{c|}{Line Search}\\
		& &  & Time & Objective & Time & Objective\\
		\hline
		
		\multirow{6}{*}{SAG} & RS  & \multirow{3}{*}{200} &  229.220102   &  0.3258410619&495.873963  &0.3258353956  \\
		
		& CS &  & 66.378361 &0.3258375894 & 214.083371  &  0.3258353935\\
		& SS &   & 67.867812  & 0.3258410840  & 214.533017    & 0.3258353956 \\

		& RS& \multirow{3}{*}{1000} & 234.129248   & 0.3258356616   & 535.821883  & 0.3258354793 \\
		
		& CS     & & 63.957239  &  0.3258358353 &  126.807235    & 0.3258354785 \\
		
		&  SS    &   &  65.908254  &  0.3258356562 &  148.786983  &  0.3258354792 \\
		\hline
		
		\multirow{6}{*}{SAGA} & RS  & \multirow{3}{*}{200} &  302.318358  & 0.3258637650 &432.787547  & 0.3258353922 \\
		
		& CS &  &147.235962  & 0.3258659974&  182.545531 & 0.3258353924 \\
		& SS &   & 148.402540  &  0.3258636638 &  177.698227  & 0.3258353937  \\

		& RS& \multirow{3}{*}{1000} &  301.003253  &  0.3259746991  & 445.741781  & 0.3258354946 \\
		
		& CS     &  &   145.917197 &  0.3259814282  & 112.433248  & 0.3258355048 \\
		
		&  SS    &   &  147.646753  &  0.3259748885 & 120.481113   & 0.3258354828  \\
		\hline
		
		%
		%
		%
		%
		%
		
		\multirow{6}{*}{SAAG-II} & RS  & \multirow{3}{*}{200} & 297.134694   &  0.3263398874&708.881659  & 0.3258353918 \\
		
		& CS &  &174.232600  & 0.3263398971& 380.852014  & 0.3258353920 \\
		& SS &   & 176.845275  &  0.3263398982 &  338.122579   &   0.3258353920\\

		& RS& \multirow{3}{*}{1000} &  299.840227  & 0.3258550019   & 687.638964  & 0.3258354037 \\
		
		& CS     &  & 171.384496   & 0.3258550023   &  209.564000 & 0.3258354039 \\
		
		&  SS    &   &  172.268358  &   0.3258550036& 213.877001   & 0.3258354041  \\
		\hline
		
		\multirow{6}{*}{SVRG} & RS  & \multirow{3}{*}{200} & 297.620959   & 0.3258923266 & 406.229956 &0.3258354055  \\
		
		& CS &  &172.405902  & 0.3258923398 & 213.151577  & 0.3258354059 \\
		& SS &   & 172.612984  &  0.3258923141 &   185.538374  &  0.3258354064 \\

		& RS& \multirow{3}{*}{1000} &  297.259155  & 0.3261069018   & 497.501496  &  0.3258357363\\
		
		& CS     &  & 172.227776   &   0.3261069217 &159.518159   & 0.3258357365 \\
		
		&  SS    &   &  172.601618  &  0.3261068804 & 151.230547   &   0.3258357352\\
		\hline
		
		%
		%
		%
		%
		%
		
		\multirow{6}{*}{MBSGD} & RS  & \multirow{3}{*}{200} &  267.252470  &  0.3258635308& 312.865696 & 0.3258353862 \\
		
		& CS &  & 144.769059 & 0.3258635315& 121.004686  & 0.3258353865 \\
		& SS &   &  140.241334 &  0.3258635313 & 122.396247    & 0.3258353867  \\

		& RS& \multirow{3}{*}{1000} & 268.102817   &  0.3259704996  & 306.236327  & 0.3258354906 \\
		
		& CS     &  & 139.586141   & 0.3259704998   &82.340378   & 0.3258354912 \\
		
		&  SS    &   &  135.646766  &  0.3259704994 & 81.486252   & 0.3258354909  \\
		\hline
	\end{tabular}
\end{table*}

\begin{table*}[h]
	\centering
	\caption{Comparison of Training Time (in seconds) and objective function values after 30 epochs using dataset SUSY}
	\label{tab_SUSY}      
	\begin{tabular}{|l|c|r|r|r|r|r|}
		\hline
		\multirow{2}{*}{Method} & \multirow{2}{*}{Sampling} & \multirow{2}{*}{Batch} & \multicolumn{2}{c|}{Constant Step} & \multicolumn{2}{c|}{Line Search}\\
		& &  & Time & Objective & Time & Objective\\
		\hline
		
		\multirow{6}{*}{SAG} & RS  & \multirow{3}{*}{200} & 84.331058    & 0.3759925588 & 267.199348 & 0.3759914171 \\
		
		& CS &  & 20.571099 & 0.3759958448& 96.172599  & 0.3759912712 \\
		& SS &   &20.380268   & 0.3759925419  &  136.218536  &  0.3759914133 \\

		& RS& \multirow{3}{*}{1000} & 86.462106   & 0.3761630312   &261.746213   &0.3759920053  \\
		
		& CS     &  & 19.652540   &  0.3764117775  & 45.862451  & 0.3759919137 \\
		
		&  SS    &   &  23.803318  &  0.3761654289 & 56.056274   &  0.3759920025 \\
		\hline
		
		\multirow{6}{*}{SAGA} & RS  & \multirow{3}{*}{200} & 119.419700   & 0.3761615578 &244.424858  & 0.3759911729 \\
		
		& CS &  & 49.054361 &0.3761668056 &88.012460   &0.3759911528  \\
		& SS &   &  50.032764 &  0.3761616154 &  125.125252   &  0.3759911723 \\

		& RS& \multirow{3}{*}{1000} &  115.240795  &   0.3767228974 & 218.120526  &0.3759920084  \\
		
		& CS     &  & 48.484855   &   0.3767380851 &41.586102   &0.3759920131  \\
		
		&  SS    &   &  53.076179  & 0.3767231692  &  47.792283  &0.3759919544   \\
		\hline
		
		%
		%
		%
		%
		%
		
		\multirow{6}{*}{SAAG-II} & RS  & \multirow{3}{*}{200} & 122.687745   & 0.3761359459 &319.227862  & 0.3759910754 \\
		
		& CS &  &57.748874  &0.3761359525 &  150.975243 & 0.3759910741 \\
		& SS &   & 58.840208  & 0.3761359506  &   156.970161  &  0.3759910768 \\

		& RS& \multirow{3}{*}{1000} & 114.239646   & 0.3759934574   & 336.178564  & 0.3759912580 \\
		
		& CS     &  &  58.377580  &  0.3759934575  & 88.889686  & 0.3759912583 \\
		
		&  SS    &   &  67.017390  &  0.3759934576 & 88.760368   & 0.3759912574  \\
		\hline
		
		\multirow{6}{*}{SVRG} & RS  & \multirow{3}{*}{200} &  128.216331  &  0.3763077569&215.183518  & 0.3759913085 \\
		
		& CS &  & 57.819864 &0.3763078069 & 88.178545  & 0.3759913110 \\
		& SS &   & 60.425710  &  0.3763079011 &  98.404560   & 0.3759913135  \\

		& RS& \multirow{3}{*}{1000} & 112.850682   &  0.3773260313  & 192.681710  & 0.3759936315 \\
		
		& CS     &  &  58.064456  & 0.3773259828   & 57.283939  &0.3759936439  \\
		
		&  SS    &   &  66.207054  &0.3773261006   &  58.508361  &  0.3759936213 \\
		\hline
		
		%
		%
		%
		%
		%
		
		\multirow{6}{*}{MBSGD} & RS  & \multirow{3}{*}{200} & 101.673967   & 0.3761560102 &174.388236  & 0.3759910145 \\
		
		& CS &  & 47.587231 & 0.3761559893 & 58.634676  &  0.3759910107\\
		& SS &   & 48.042649  & 0.3761559954  &  63.347389   &   0.3759910169\\

		& RS& \multirow{3}{*}{1000} &  103.513668  &   0.3766979888 &  128.650218 & 0.3759918962 \\
		
		& CS     &  & 47.346182   &  0.3766979329  & 31.396945  & 0.3759918950 \\
		
		&  SS    &   & 55.683464   &  0.3766979490 & 41.701046   & 0.3759918931  \\
		\hline
	\end{tabular}
\end{table*}

\begin{table}[h]
	\caption{Comparison of Training Time (in seconds) and objective function values after 30 epochs using dataset covtype.binary}
	\label{tab_covtype}      
	\begin{tabular}{|l|c|r|r|r|r|r|}
		\hline
		\multirow{2}{*}{Method} & \multirow{2}{*}{Sampling} & \multirow{2}{*}{Batch} & \multicolumn{2}{c|}{Constant Step} & \multicolumn{2}{c|}{Line Search}\\
		& &  & Time & Objective & Time & Objective\\
		\hline
		
		\multirow{6}{*}{SAG} & RS  & \multirow{3}{*}{200} &  7.688000   &  0.0000000003&16.485357  &0.0000009291  \\
		
		& CS &  & 2.158810 &0.0000000002 & 5.874758  &0.0000024168  \\
		& SS &   & 1.796637 & 0.0000000004   &   3.327142  & 0.0000005212  \\

		& RS& \multirow{3}{*}{1000} &  7.624630  &  0.0000000015  & 20.674648  & 0.0000003902 \\
		
		& CS     &  & 2.110367   &  0.0000014178  & 3.436548  & 0.0000006984 \\
		
		&  SS    &   & 1.608393   &   0.0000000003& 4.217670  &  0.0000012383 \\
		\hline
		
		\multirow{6}{*}{SAGA} & RS  & \multirow{3}{*}{200} &  9.512428  &  0.0000398812& 33.541463 & 0.0000000001 \\
		
		& CS &  &4.200649  &0.0000425219 & 36.688386  &0.0000000001  \\
		& SS &   & 2.909938  & 0.0000370651  & 23.002344    & 0.0000000001  \\

		& RS& \multirow{3}{*}{1000} & 9.378465   &  0.0001809634  & 34.770427  & 0.0000000001 \\
		
		& CS     &  &  4.049684  &   0.0001901968 &7.649198   & 0.0000000002 \\
		
		&  SS    &   & 2.414478   & 0.0001763171  &  19.890575  & 0.0000000001  \\
%
%

%
%
		\hline
		
		\multirow{6}{*}{SAAG-II} & RS  & \multirow{3}{*}{200} & 8.510917   &0.0000001565  &44.963308  & 0.0000000001 \\
		
		& CS &  & 4.568239 & 0.0000001565& 37.523382  & 0.0000000001 \\
		& SS &   & 2.729112  &  0.0000001566 &  17.364469   & 0.0000000001  \\

		& RS& \multirow{3}{*}{1000} &  13.558037  & 0.0000000063   & 56.251227  &  0.0000000001\\
		
		& CS     &  & 6.355990   &  0.0000000063  & 37.831273  &  0.0000000001\\
		
		&  SS    &   & 4.325896   &  0.0000000063 &  19.511468  & 0.0000000001  \\
		\hline
		
		\multirow{6}{*}{SVRG} & RS  & \multirow{3}{*}{200} &  9.370933  & 0.0000771729 &32.835999  & 0.0000000001 \\
		
		& CS &  &5.234851  & 0.0000000005& 39.737523  &  0.0000000001\\
		& SS &   & 3.645048  & 0.0000759334  & 21.836150    & 0.0000000001  \\

		& RS& \multirow{3}{*}{1000} &  10.200179  &  0.0003404686  & 23.870644  &0.0000000001  \\
		
		& CS     &  &  5.031777  &   0.0000009870 &34.195127   & 0.0000000001 \\
		
		&  SS    &   & 3.609489   & 0.0003181269  & 7.203759   &  0.0000000005 \\
%
%

%
%
		\hline
		
		\multirow{6}{*}{MBSGD} & RS  & \multirow{3}{*}{200} &  8.314825  & 0.0000390866 & 27.593407 &0.0000000001  \\
		
		& CS &  & 4.142554 & 0.0000393200& 17.189169  & 0.0000000001 \\
		& SS &   &  3.050657 & 0.0000391765  &  11.707410   &  0.0000000001 \\

		& RS& \multirow{3}{*}{1000} &  8.966491  & 0.0001756815   & 11.959532  & 0.0000000001 \\
		
		& CS     &  &3.979792    &  0.0001747116  & 18.732325  &0.0000000001  \\
		
		&  SS    &   &  2.456082  & 0.0001707026  &  9.960456  &  0.0000000001 \\
		\hline
	\end{tabular}
\end{table}

\section{Conclusion}
\label{sec_conclusion}
In this paper, novel systematic sampling and cyclic sampling techniques have been proposed, for solving large-scale learning problems, for improving the training time by reducing the data access time. Methods have similar convergence in expectation for systematic and cyclic sampling, as for widely used random sampling, but with cyclic and systematic sampling the training is up to six times faster than widely used random sampling technique, at the expense of fractionally small difference in the minimized objective function value, for a given number of epochs. Thus, systematic sampling technique is suitable for solving large-scale problem with low accuracy solution. Random shuffling of data can be used before the data is fed to the learning algorithms with systematic and cyclic sampling to improve their results for the cases where similar data points are grouped together. These sampling techniques can be extended to parallel and distributed learning algorithms.

\section*{Acknowledgements}
	First author is thankful to Ministry of Human Resource Development, Government of INDIA, to provide fellowship (University Grants Commission - Senior Research Fellowship) to pursue his PhD.\\
	We acknowledge Manish Goyal, Senior Research Fellow, Department of Statistics, Panjab University for helping with the statistical insights around sampling and expectation.

%


\clearpage

\begin{thebibliography}{33}
	\providecommand{\natexlab}[1]{#1}
	\providecommand{\url}[1]{{#1}}
	\providecommand{\urlprefix}{URL }
	\expandafter\ifx\csname urlstyle\endcsname\relax
	\providecommand{\doi}[1]{DOI~\discretionary{}{}{}#1}\else
	\providecommand{\doi}{DOI~\discretionary{}{}{}\begingroup
		\urlstyle{rm}\Url}\fi
	\providecommand{\eprint}[2][]{\url{#2}}
	
	\bibitem[{Byrd et~al(2016)Byrd, Hansen, Nocedal, and Singer}]{Byrd2016}
	Byrd RH, Hansen SL, Nocedal J, Singer Y (2016) A stochastic quasi-newton method
	for large-scale optimization. {SIAM} Journal on Optimization 26(2):1008--1031
	
	\bibitem[{Chang et~al(2008)Chang, Hsieh, and Lin}]{Chang2008}
	Chang KW, Hsieh CJ, Lin CJ (2008) Coordinate descent method for large-scale
	l2-loss linear support vector machines. J Mach Learn Res 9:1369--1398
	
	\bibitem[{Chauhan et~al(2017)Chauhan, Dahiya, and Sharma}]{Chauhan2017Saag}
	Chauhan VK, Dahiya K, Sharma A (2017) Mini-batch block-coordinate based
	stochastic average adjusted gradient methods to solve big data problems. In:
	Proceedings of the Ninth Asian Conference on Machine Learning, PMLR, vol~77,
	pp 49--64, \urlprefix\url{http://proceedings.mlr.press/v77/chauhan17a.html}
	
	\bibitem[{Chauhan et~al(2018{\natexlab{a}})Chauhan, Dahiya, and
		Sharma}]{Chauhan2018SS}
	Chauhan VK, Dahiya K, Sharma A (2018{\natexlab{a}}) Faster algorithms for
	large-scale machine learning using simple sampling techniques. arXiv
	\urlprefix\url{https://arxiv.org/abs/1801.05931v2}
	
	\bibitem[{Chauhan et~al(2018{\natexlab{b}})Chauhan, Dahiya, and
		Sharma}]{Chauhan2018Review}
	Chauhan VK, Dahiya K, Sharma A (2018{\natexlab{b}}) Problem formulations and
	solvers in linear svm: a review. Artificial Intelligence Review
	\doi{10.1007/s10462-018-9614-6},
	\urlprefix\url{https://doi.org/10.1007/s10462-018-9614-6}
	
	\bibitem[{Chauhan et~al(2018{\natexlab{c}})Chauhan, Sharma, and
		Dahiya}]{Chauhan2018SS_AI}
	Chauhan VK, Sharma A, Dahiya K (2018{\natexlab{c}}) Faster learning by
	reduction of data access time. Applied Intelligence
	\doi{10.1007/s10489-018-1235-x},
	\urlprefix\url{https://doi.org/10.1007/s10489-018-1235-x}
	
	\bibitem[{Chauhan et~al(2018{\natexlab{d}})Chauhan, Sharma, and
		Dahiya}]{Chauhan2018SAAGs34}
	Chauhan VK, Sharma A, Dahiya K (2018{\natexlab{d}}) {SAAGs: Biased Stochastic
		Variance Reduction Methods}. arXiv
	\urlprefix\url{http://arxiv.org/abs/1807.08934}, \eprint{1807.08934}
	
	\bibitem[{Cotter et~al(2011)Cotter, Shamir, Srebro, and Sridharan}]{Cotter2011}
	Cotter A, Shamir O, Srebro N, Sridharan K (2011) {Better Mini-Batch Algorithms
		via Accelerated Gradient Methods}. Nips pp 1--9,
	\urlprefix\url{http://arxiv.org/abs/1106.4574}, \eprint{1106.4574}
	
	\bibitem[{Csiba and Richt(2016)}]{Csiba2016}
	Csiba D, Richt P (2016) {Importance Sampling for Minibatches} pp 1--19,
	\eprint{arXiv:1602.02283v1}
	
	\bibitem[{Csiba and Richt{\'{a}}rik(2016)}]{Csiba2016b}
	Csiba D, Richt{\'{a}}rik P (2016) {Coordinate Descent Face-Off: Primal or
		Dual?} pp 1--17, \eprint{1605.08982}
	
	\bibitem[{Defazio et~al(2014)Defazio, Bach, and Lacoste-Julien}]{Defazio2014}
	Defazio A, Bach F, Lacoste-Julien S (2014) Saga: A fast incremental gradient
	method with support for non-strongly convex composite objectives. In:
	Proceedings of the 27th International Conference on Neural Information
	Processing Systems, MIT Press, Cambridge, MA, USA, NIPS'14, pp 1646--1654
	
	\bibitem[{Gopal(2016)}]{Gopal2016}
	Gopal S (2016) {Adaptive Sampling for SGD by Exploiting Side Information}. Icml
	
	\bibitem[{Johnson and Zhang(2013)}]{Johnson2013}
	Johnson R, Zhang T (2013) Accelerating stochastic gradient descent using
	predictive variance reduction. In: Burges CJC, Bottou L, Welling M,
	Ghahramani Z, Weinberger KQ (eds) Advances in Neural Information Processing
	Systems 26, Curran Associates, Inc., pp 315--323
	
	\bibitem[{Kone{\v{c}}n{\'{y}} and Richt{\'{a}}rik(2013)}]{Konecny2013}
	Kone{\v{c}}n{\'{y}} J, Richt{\'{a}}rik P (2013) {Semi-Stochastic Gradient
		Descent Methods} 1:19, \urlprefix\url{http://arxiv.org/abs/1312.1666},
	\eprint{1312.1666}
	
	\bibitem[{Li et~al(2014)Li, Zhang, Chen, and Smola}]{Li2014}
	Li M, Zhang T, Chen Y, Smola AJ (2014) Efficient mini-batch training for
	stochastic optimization. In: Proceedings of the 20th ACM SIGKDD International
	Conference on Knowledge Discovery and Data Mining, ACM, New York, NY, USA,
	KDD '14, pp 661--670
	
	\bibitem[{Madow(1949)}]{William1949}
	Madow WG (1949) On the theory of systematic sampling, ii. The Annals of
	Mathematical Statistics 20:333--354,
	\urlprefix\url{http://www.jstor.org/stable/2236532}
	
	\bibitem[{Madow and Madow(1944)}]{William1944}
	Madow WG, Madow LH (1944) On the theory of systematic sampling, i. The Annals
	of Mathematical Statistics 15:1--24,
	\urlprefix\url{http://www.jstor.org/stable/2236209}
	
	\bibitem[{Needell et~al(2014)Needell, Ward, and Srebro}]{Deanna2014}
	Needell D, Ward R, Srebro N (2014) Stochastic gradient descent, weighted
	sampling, and the randomized kaczmarz algorithm. In: Ghahramani Z, Welling M,
	Cortes C, Lawrence ND, Weinberger KQ (eds) Advances in Neural Information
	Processing Systems 27, Curran Associates, Inc., pp 1017--1025
	
	\bibitem[{Nemirovski et~al(2009)Nemirovski, Juditsky, Lan, and
		Shapiro}]{Nemirovski2009}
	Nemirovski A, Juditsky A, Lan G, Shapiro A (2009) Robust stochastic
	approximation approach to stochastic programming. SIAM Journal on
	Optimization 19(4):1574--1609
	
	\bibitem[{Qu and Richt(2015)}]{Qu2015}
	Qu Z, Richt P (2015) {Randomized Dual Coordinate Ascent with Arbitrary
		Sampling}. Neural Information Processing Systems (1):1--34,
	\eprint{arXiv:1411.5873v1}
	
	\bibitem[{Reddi(2016)}]{Reddi2016}
	Reddi SJ (2016) {New Optimization Methods for Modern Machine Learning}. PhD
	thesis
	
	\bibitem[{Schmidt et~al(2016)Schmidt, Le~Roux, and Bach}]{Schmidt2016}
	Schmidt M, Le~Roux N, Bach F (2016) Minimizing finite sums with the stochastic
	average gradient. Math Program pp 1--30
	
	\bibitem[{Shalev-Shwartz et~al(2011)Shalev-Shwartz, Singer, Srebro, and
		Cotter}]{Shalev-Shwartz2011}
	Shalev-Shwartz S, Singer Y, Srebro N, Cotter A (2011) Pegasos: primal estimated
	sub-gradient solver for svm. Math Program 127(1):3--30
	
	\bibitem[{Takáč et~al(2013)Takáč, Bijral, Richtárik, and
		Srebro}]{Martin2013}
	Takáč M, Bijral A, Richtárik P, Srebro N (2013) Mini-batch primal and dual
	methods for svms. In: In 30th International Conference on Machine Learning,
	Springer, pp 537--552
	
	\bibitem[{Wang et~al(2017)Wang, Wang, and Zhang}]{Wang2017}
	Wang X, Wang S, Zhang H (2017) Inexact proximal stochastic gradient method for
	convex composite optimization. Computational Optimization and Applications
	\doi{10.1007/s10589-017-9932-7},
	\urlprefix\url{https://doi.org/10.1007/s10589-017-9932-7}
	
	\bibitem[{Wright(2015)}]{Wright2015}
	Wright SJ (2015) Coordinate descent algorithms. Math Program 151(1):3--34
	
	\bibitem[{Xu and Yin(2015)}]{Xu2015}
	Xu Y, Yin W (2015) Block stochastic gradient iteration for convex and nonconvex
	optimization. SIAM Journal on Optimization 25(3):1686--1716
	
	\bibitem[{Yu et~al(2010)Yu, Hsieh, Chang, and Lin}]{Yu2010}
	Yu HF, Hsieh CJ, Chang KW, Lin CJ (2010) Large linear classification when data
	cannot fit in memory. In: Proceedings of the 16th ACM SIGKDD International
	Conference on Knowledge Discovery and Data Mining, ACM, New York, NY, USA,
	KDD '10, pp 833--842
	
	\bibitem[{Zhang and Gu(2016)}]{Zhang2016}
	Zhang A, Gu Q (2016) Accelerated stochastic block coordinate descent with
	optimal sampling. In: Proceedings of the 22Nd ACM SIGKDD International
	Conference on Knowledge Discovery and Data Mining, ACM, New York, NY, USA,
	KDD '16, pp 2035--2044
	
	\bibitem[{Zhang(2004)}]{Zhang2004}
	Zhang T (2004) Solving large scale linear prediction problems using stochastic
	gradient descent algorithms. In: Proceedings of the Twenty-first
	International Conference on Machine Learning, ACM, New York, NY, USA, ICML
	'04, pp 116--
	
	\bibitem[{Zhao and Zhang(2014)}]{Zhao2014b}
	Zhao P, Zhang T (2014) {Accelerating Minibatch Stochastic Gradient Descent
		using Stratified Sampling}. arXiv Prepr arXiv14053080 pp 1--13,
	\urlprefix\url{http://arxiv.org/abs/1405.3080}
	
	\bibitem[{Zhao and Zhang(2015)}]{Zhao2015}
	Zhao P, Zhang T (2015) {Stochastic Optimization with Importance Sampling for
		Regularized Loss Minimization}. ICML 37
	
	\bibitem[{Zhou et~al(2017)Zhou, Pan, Wang, and Vasilakos}]{Zhou2017}
	Zhou L, Pan S, Wang J, Vasilakos AV (2017) Machine learning on big data:
	Opportunities and challenges. Neurocomputing 237:350 -- 361,
	\doi{https://doi.org/10.1016/j.neucom.2017.01.026},
	\urlprefix\url{https://www.sciencedirect.com/science/article/pii/S0925231217300577}
	
\end{thebibliography}


\end{document}